\documentclass[accepted]{uai2023}
\usepackage[american]{babel}
\usepackage{natbib}
\usepackage{mathtools} %
\usepackage{booktabs} %
\usepackage{tikz}
\usepackage{amsmath, amsthm,amssymb, amsfonts, xcolor, tikz, float, url, algorithm,algpseudocode, bm, bbm, mathtools}
\usepackage{thmtools}
\usepackage{algorithm}
\usepackage{algpseudocode}
\usepackage{caption}
\usepackage{subcaption}
\usepackage{lipsum}
\usepackage{wrapfig}
\usepackage{xpatch}
\usepackage{xspace}
\usepackage{bookmark}
\usepackage{hyperref}
\usepackage[capitalise]{cleveref}

\makeatletter

\definecolor{hanblue}{rgb}{0.27, 0.42, 0.81}
\hypersetup{
hidelinks,
    colorlinks=true,
    linkcolor=hanblue,
    urlcolor=hanblue,
    citecolor=hanblue,
    anchorcolor=black}
\usetikzlibrary{decorations, calligraphy, positioning}
\usepackage{multirow}

\newcommand\norm[1]{\|#1 \|}

\newtheorem{theorem}{Theorem}[section]
\newtheorem{remark}[theorem]{Remark}
\newtheorem{assumption}[theorem]{Assumption}
\newtheorem{proposition}[theorem]{Proposition}
\newtheorem{definition}[theorem]{Definition}
\newtheorem{corollary}[theorem]{Corollary}
\newtheorem{lemma}[theorem]{Lemma}

\DeclareMathOperator{\diag}{diag}
\DeclareMathOperator{\trace}{tr}

\definecolor{parnian}{rgb}{0.36, 0.54, 0.66}
\definecolor{oracle}{HTML}{4f7992}
\definecolor{naive}{HTML}{d33f49}
\definecolor{meta}{HTML}{419d78}
\definecolor{federated}{HTML}{f9a620}
\definecolor{grey}{rgb}{0.7, 0.75, 0.71}
\definecolor{optimalrate}{HTML}{69C498}%
\definecolor{notopt}{HTML}{d33f49}%

\def\eqref#1{equation~\ref{#1}}
\def\Eqref#1{Equation~\ref{#1}}

\def\floor#1{\lfloor #1 \rfloor}
\def\1{\bm{1}}

\def\btheta{{\bm{\theta}}}

\def\bb{{\bm{b}}}

\def\bw{{\bm{w}}}
\def\bx{{\bm{x}}}
\def\by{{\bm{y}}}

\def\mB{{\bm{B}}}

\def\mS{{\bm{S}}}

\def\mPhi{{\bm{\Phi}}}

\def\calA{{\mathcal{A}}}

\def\calC{{\mathcal{C}}}
\def\calD{{\mathcal{D}}}

\def\calH{{\mathcal{H}}}

\def\calN{{\mathcal{N}}}
\def\calO{{\mathcal{O}}}

\def\calU{{\mathcal{U}}}

\def\calX{{\mathcal{X}}}

\def\sN{{\mathbb{N}}}

\def\sP{{\mathbb{P}}}

\def\sR{{\mathbb{R}}}

\newcommand{\ubar}[1]{\text{\b{$#1$}}}

\newcommand{\Ls}{\mathcal{L}}
\newcommand{\R}{\mathbb{R}}

\DeclareMathOperator*{\argmax}{arg\,max}
\DeclareMathOperator*{\argmin}{arg\,min}

\def\algoff{{\textsc{Meta-KGL}}\xspace}
\def\algon{{\textsc{LiBO}}\xspace}
\def\falgoff{{\textsc{F-Meta-KGL}}\xspace}
\def\falgon{{\textsc{F-LiBO}}\xspace}
\def\gpucb{{\textsc{GP-UCB}}\xspace}
\def\bba{{\textsc{BaseBO}}\xspace}

\def\tH{{\calH_{k^\star}}}

\def\tk{{k^\star}}

\def\tvbeta{\bm{\beta}^\star}
\def\khat{{\hat{k}}}

\def\vbetahat{\hat{\bm{\beta}}}

\def\vbeta{{\bm{\beta}}}
\def\vphi{{\bm{\phi}}}
\def\gj{{^{(j)}}}
\def\tJ{{J^\star}}
\def\Jhat{{\hat{J}}}

\def\kfull{{k^\mathrm{full}}}
\def\Roracle{{R^\star}} %
\def\Dexp{\calD^{\mathrm{exp}}}

\makeatletter
\def\munderbar#1{\underline{\sbox\tw@{$#1$}\dp\tw@\z@\box\tw@}}
\makeatother

\usepackage{multicol}
\newcommand*{\myalign}[2]{\multicolumn{1}{#1}{#2}}

\usepackage{colortbl}

\usepackage{pifont}
\newcommand{\cmark}{\ding{51}}%
\newcommand{\xmark}{\ding{55}}%
\usepackage{makecell}

\makeatletter
\newcommand{\printfnsymbol}[1]{%
  \textsuperscript{\@fnsymbol{#1}}%
}
\makeatother

\title{Lifelong Bandit Optimization: No Prior and No Regret}

\author[1]{Felix~Schur\thanks{ Equal contribution.}}
\author[1]{Parnian~Kassraie\printfnsymbol{1}}
\author[1]{Jonas~Rothfuss}
\author[1]{Andreas~Krause}
\affil[1]{%
    ETH Zurich\\
    Switzerland
}
  
\begin{document}
\maketitle

\begin{abstract}
\looseness -1 Machine learning algorithms are often repeatedly applied to problems with similar structure over and over again. We focus on solving a sequence of bandit optimization tasks and develop \algon, an algorithm which {\em adapts} to the environment by learning from past experience and becomes more sample-efficient in the process. 
We assume a kernelized structure where the kernel is {\em unknown} but {\em shared} across all tasks.
\algon sequentially meta-learns a kernel that approximates the true kernel and solves the incoming tasks with the latest kernel estimate. Our algorithm can be paired with {\em any} kernelized or linear bandit algorithm and guarantees {\em oracle optimal} performance, meaning that as more tasks are solved, the regret of \algon on each task converges to the regret of the bandit algorithm with oracle knowledge of the true kernel. Naturally, if paired with a sublinear bandit algorithm, \algon yields a sublinear lifelong regret.  We also show that direct access to the data from each task is not necessary for attaining sublinear regret. We propose \falgon, which solves the lifelong problem in a federated manner. \looseness-1
\end{abstract}
\section{Introduction} \label{sec:intro}
A key aspect of human intelligence is our ability to harness previous experience and quickly improve when repeatedly solving similar problems. 
In this paper, we study how to solve a {\em sequence of learning problems, on related instances}, and become more efficient in the process.  
In particular we focus on problems which are solved through Bayesian Optimization, a.k.a.~kernelized bandit algorithms (BO), where the kernel captures regularity structure of the tasks. 
A motivating application are AutoML systems, which perform hyper-parameter tuning for the same model on different datasets, or different models on the same dataset. We expect that the more tasks our machine learning system solves, the better the system becomes at solving the next one.  

\looseness -1 We model this as {\em lifelong learning},
where an agent sequentially faces kernelized bandit problems with different unknown reward functions. 
While prior work assumes the kernel to be known (e.g., hand-designed), we consider the kernel $k^*$ to be {\em unknown}, but {\em shared} between the problem instances. 
After each bandit task, we use the previously collected data to {\em meta-learn a kernel function} $\hat{k}$ as a proxy for the unknown $\tk$. 
We transfer knowledge across tasks by sequentially updating the meta-learned kernel and using it to solve the next task. This way, we adapt to the environment and gradually improve the bandit performance.
Ideally, we would like to reach the oracle-optimal performance, i.e. the performance of a bandit algorithm with complete knowledge of the environment.

Lifelong bandit optimization is a delicate problem for two reasons. First, the success of each round of BO depends on the validity of the meta-learned kernel: We only have guaranteed convergence and sublinear regret if the reproducing kernel Hilbert space (RKHS), induced by the estimated kernel $\hat{k}$, contains the reward functions. Second, the data that is used for meta-learning is collected at the previous BO tasks. Thus, during each BO round, we not only have to quickly find reward maximizing actions, but also have to collect exploratory data that is sufficiently informative for successful meta-learning of the kernel.

We address these challenges when the true kernel is a sparse convex combination of a large number of candidate kernels.
We propose an approach for meta-learning a provably consistent estimator of the true kernel, given data from previous tasks (\cref{thm:offline_main_consitency}).
To ensure that this data is sufficiently informative, we interlace the queries of the BO agent with purely exploratory queries.
Combining these two key ideas, we design our main algorithm, the {\em Lifelong Bandit Optimizer (\algon)}.
This algorithm is versatile since it is {\em agnostic to the bandit policy}, i.e. it can be wrapped around {\em any} kernelized or linear base bandit algorithm to influence its policy and satisfy lifelong guarantees. 
We prove that it is {\em oracle-optimal}, i.e. that by using \algon, we can eventually achieve the {\em same} worst-case performance as the base bandit algorithm which has oracle knowledge of the true kernel (\cref{thm:lifelong}). 
We {\em do not make assumptions} about the base bandit algorithm, and our convergence guarantees hold for many bandit solvers such as OFUL \citep{abbasi2011improved}, GP-UCB \citep{srinivas2009gaussian} or GP-TS \citep{chowdhury2017kernelized}. 
Additionally, we consider a federated setting where each BO task is performed by a client node in a network and the data ought not to be exchanged with the server node due to privacy concerns.
We propose the {\em Federated Lifelong Bandit Optimizer (\falgon)}, 
and show that it satisfies a guarantee similar to \algon (\cref{thm:lifelong_federated}).
If we take \gpucb as base bandit solver,
\algon and \falgon have the {\em same worst-case regret bound rates} as the \gpucb solver when given {\em oracle knowledge} of the true kernel (\cref{cor:ll_ucb_decreasing_main_text} and \ref{cor:ucb_federated}).
In \cref{sec:experiments} we support our theoretical findings by experiments on synthetic and real-world data in the AutoML context. 
Lastly, we discuss related works in \cref{sec:related_work}. \looseness-1
\section{Problem Statement} 
\label{sec:problem}
\looseness -1 We consider a lifelong optimization setting, where an agent interacts with a sequence of black-box optimization problems, arriving one after another. Throughout the sequence of optimization tasks, the agent can adapt to the environment based on the previously collected data and improve its performance on the succeeding tasks.
Formally, the agent iteratively faces bandit problems with unknown reward functions $f_1, ..., f_m$ residing in a RKHS $\tH$ that corresponds to an {\em unknown} kernel function $\tk$. 
To impose regularity, we assume that the reward function has a bounded kernel norm $\norm{f}_\tk \leq B$ and that the domain $\calX \subset \sR^{d_0}$ is compact.
The agent interacts with each task $f_s$ for $n$ time steps. For each task $s =1,\dots,m$, at time step $i =1, \dots, n$, the agent selects an action $\bx_{s,i} \in \calX$ and receives a stochastic reward via
$y_{s,i} = f_{s}(\bx_{s,i}) + \varepsilon_{s,i}$. Here, $\varepsilon_{s,i}$ are i.i.d.~samples from a zero-mean sub-Gaussian noise with variance proxy $\sigma^2$.
The goal of the agent is to maximize its rewards across all tasks. This can be formalized as minimizing the {\em lifelong regret} over $m$ tasks of size $n$, defined as
\[
R(m,n) \coloneqq \sum_{s=1}^m \sum_{i=1}^n f_s(\bx_s^\star) - f_s(\bx_{s,i}),
\]
where $\bx_s^\star$ is a global maximum of $f_s$.
If \smash{$\frac{R(m,n)}{mn} \rightarrow 0$} as $m,n \rightarrow \infty$ then the agent eventually converges to the global optimum of each upcoming optimization task. This property is commonly referred to as \emph{sublinearity} of the regret.
\looseness -1 To attain a small regret, the agent maintains an estimate of the unknown reward function $f_s$ based on its history.
Typically, a kernelized regression oracle (e.g. kernel ridge regression or Gaussian Processes) is employed for this task. The choice of the kernel function plays a key role in the success and data-efficiency of the bandit optimization. If the hypothesis space $\calH_k$ induced by the kernel $k$ is too restrictive and does not contain the true reward functions $f_s$, the agent will likely never find reward maximizing actions. To prevent this, most practitioners pick a kernel with a conservatively complex kernel with a large hypothesis space that is very likely to contain $\calH_{k^*}$. However, the larger $\calH_k$, the more observations it takes to form a good reward estimate, making the finding an optimal solution less efficient.

\looseness -1 We take a data-driven approach to select the kernel. In particular, we aim to sequentially meta-learn a kernel $\hat{k}$ which approximates the true kernel $\tk$, using the data from previous bandit tasks.
Let $\calD_s \coloneqq \{ (\bx_{s,i}, y_{s,i})_{i \leq n} \}$ be the data corresponding to task $s$, and $\calD_{1:s} \coloneqq \calD_1 \cup \dots \cup \calD_s$ be the collection of datasets from the first $s$ tasks.
Then once the agent solves task $s$, we pass $\calD_{1:s}$ to the {\em meta-agent}, who meta-learns a kernel $\khat_s$. This kernel is then provided to the agent who uses it for solving the next BO task. 
Our meta-learning algorithm can be paired with any kernelized bandit algorithm and achieves sublinear lifelong regret, if the bandit algorithm achieves sublinear single-task regret given oracle knowledge of the kernel. 
\section{Meta-Learning Kernels}
\label{sec:metalearn}

We first present {\em Meta Kernelized Group Lasso (\algoff)}, our approach to estimate the kernel $\tk$, given data from previous tasks.
We consider a large set of eligible known base kernels $\{k_1, \dots, k_p\}$ where $k_j: \calX \times \calX \to \sR$ for all $j = 1, \dots, p$ and $k_j(\bx,\bx^\prime) \leq 1$ for all $\bx,\bx^\prime \in \calX$ without loss of generality.
We assume that while $\tk$ is unknown, it is a sparse
linear combination of kernels selected from this set, i.e., 
there exists $J^\star \subset \{1,\dots,p\}$ and $\alpha_1, \dots \alpha_p \in \mathbb{R}$ such that
\[
    \tk(\bx,\bx') = \sum_{j \in J^\star} \alpha_j k_j(\bx,\bx^\prime),
\]
where $|\tJ| \ll p$. The set $\{k_1, \ldots, k_p\}$ can be very large, since we prove that the cost of finding $\tJ$ depends only logarithmically on $p$.
We further assume that each $k_j$ corresponds to a $d_j$-dimensional feature map, i.e., $k_j(\bx,\bx^\prime) = \vphi_j(\bx)^T\vphi_j(\bx^\prime)$, where $\vphi_j \in \sR^{d_j}$ and $d_j < \infty$.
This setting generalizes the common linear bandit assumption, to also account for higher-order terms and interaction between coordinates of the input.
Let $\vphi(\bx)$ denote the concatenated $d$-dimensional feature map, where $d \coloneqq  \sum_{j=1}^pd_j$ and
$
\vphi(\bx) \coloneqq [\vphi_j^T(\bx)]_{j \leq p}.
$
Then for $s = 1, \dots, m$ the reward functions can be written as 
$f_s(\cdot) = \sum_{j=1}^p \vphi_j^\top(\cdot)\tvbeta_s\gj$ such that
$\tvbeta_s\gj = 0$ for all $j \notin \tJ$. Moreover, the RKHS norm of $f_s$ will be equal to $\norm{f_s}^2_\tk = \sum_{j=1}^p \norm{\tvbeta_s\gj}_2^2$. 
This kernel model is inspired by \citet{kassraie2022metalearning}, who assume $\tk$ lies in the convex cone of the base kernels.  

\looseness -1 In the lifelong setting, we sequentially form kernel estimates $\khat_s$ based on $\calD_{1:s}$ for $s=1, \dots, m$. In this section, we consider one snapshot of this process for $s=m$, where we have fixed meta-training data $\calD_{1:m}$ and meta-learn $\khat \coloneqq \khat_m$.
We assume without loss of generality (c.f. \cref{app:wlog}),
\begin{align*}
    \tk(\bx,\bx') = \frac{1}{|\tJ|} \sum_{j \in J^\star} k_j(\bx,\bx^\prime),
\end{align*}
and minimize a sparsity inducing loss which allows us to discard kernels that do not appear in the above formulation.
\algoff first minimizes  $\Ls(\vbeta; \calD_{1:m})$ over $\vbeta \in \sR^{md}$. 
\begin{align}
\label{eq:meta_loss}
\Ls\left(\vbeta; \calD_{1:m} \right) \coloneqq & \frac{1}{\vert \calD_{1:m}\vert} \norm{ \by - \mPhi \vbeta}_2^2  + \lambda \sum_{j=1}^{p} \|\vbeta\gj\|_2 \\
 =& \frac{1}{mn}  \sum_{s=1}^m\sum_{i=1}^n \big(y_{s,i} - \sum_{j=1}^{p} \vphi_j^T(\bx_{s,i})\vbeta_s\gj\big)^2 \nonumber\\
& \,\,+ \lambda \sum_{j=1}^{p} \sqrt{\sum_{s=1}^m\|\vbeta_s\gj\|^2_2}\nonumber
\end{align}
The vectorized formulation uses the following notation
\begin{align*}
    \by & \coloneqq \left[\left[y_{1, i}\right]_{i\leq n}, \dots,  \left[y_{m, i}\right]_{i\leq n}\right] \in \sR^{mn},\\
    \vbeta & \coloneqq \left[ [\vbeta_1\gj]_{j \leq p}, \dots, [\vbeta_m\gj]_{j \leq p} \right] \in \sR^{md},\\
    \vbeta\gj & \coloneqq \left[ \vbeta_1\gj, \dots, \vbeta_m\gj \right] \in \sR^{md_j},\\
    \mPhi & \coloneqq \diag(\mPhi_1, \dots \mPhi_m) \in \sR^{mn \times dm}.
\end{align*}
Here $\mPhi_s \coloneqq (\vphi^\top(\bx_{s,1}), \dots, \vphi^\top(\bx_{s, n}))^\top$ is the $n \times d$ feature matrix of a task $s$, and therefore $\mPhi$ denotes a block diagonal matrix which gathers the features across all tasks. 
This meta-loss function is convex, and is equivalent to the well-known Group Lasso objective \citep{lounici2011oracle}. Therefore, it can be efficiently optimized using Group Lasso solvers \citep[e.g.,][]{celer2018} and enjoys the statistical properties of the Group Lasso, e.g., consistency and variable selection.
Let $\vbetahat \coloneqq \argmin \Ls(\vbeta; \calD_{1:m})$. 
The first term in \cref{eq:meta_loss} represents the squared prediction error of $\vbetahat$, while the second is a regularization term that induces group sparsity in $\vbetahat$.
Mainly, the solutions $\vbetahat = (\vbetahat^{(1)}, \dots, \vbetahat^{(p)})$ to this problem are group sparse, i.e. $\vbetahat\gj=0$ for many of the indices $j\in \{1, \dots, p\}$.
\algoff then constructs the set of plausible kernels $\Jhat$, by thresholding $\norm{\vbetahat\gj}_2$ and discarding the kernels that do no appear to be influencing the data, i.e., 
\begin{align*}
    \Jhat \coloneqq \left\{ j ~|~ j \in \{1,\dots,p\}\ \text{s.t.}\ \norm{\vbetahat\gj}_2 > \omega \sqrt{m} \right\}.
\end{align*}
where $\omega > 0$ is a hyperparamter of the algorithm. We then construct the estimated kernel as
\[
\khat(\bx,\bx^\prime) \coloneqq \frac{1}{|\Jhat|} \sum_{j \in \Jhat} k(\bx,\bx^\prime).
\]
\algoff is summarized in Algorithm \ref{alg:meta_learning}. 
Under mild assumptions on the dataset, we can show that $\khat$ converges to the true kernel $\tk$ in probability. 
Our first assumption ensures that if $j \in \tJ$, i.e., $k_j$ is active in the true kernel, then the contribution of $k_j$ to the data is large enough to be statistically detectable under noise.
\begin{assumption}[Beta-min]
\label{ass:betamin}
    There exists $c_1 > 0$ such that for all $j \in J^\star$,
    \begin{equation*}
        \norm{\vbeta^{*(j)}}_2 \geq c_1 \sqrt{m}.
    \end{equation*}
\end{assumption}
This assumption is commonly used in the high-dimensional statistics literature \citep{buhlmann2011statistics,bunea2013group,zhao2006model}.
Our second assumption requires that the meta-training data is sufficiently diverse. In \cref{prop:forced_exp}, we propose a policy which provably satisfies this assumption.
\begin{assumption}[Sufficiently Informative Data]
\label{ass:compatibility}
The feature matrix  $\mPhi\in \sR^{mn \times d}$ is sufficiently informative if
there exists a constant $c_\kappa>0$ such that $\kappa(\mPhi) \geq c_\kappa$ where
\begin{align*}
\kappa(\mPhi) \coloneqq & \inf_{(J,\bb)} \frac{1}{\sqrt{n}} \frac{\|\mPhi \bb\|_2}{ \sum_{j \in J} \|\bb\gj\|_2}\\
 \text{s.t. }&  \ \bb \in \sR^d\backslash \{0\},\ \sum_{j \notin J} \|\bb^{(j)}\|_2 \leq 3 \sum_{j \in J} \|\bb^{(j)}\|_2,\\
 \,\, & J \subset \{1, \dots, p\}, \ \vert J \vert \leq \vert \tJ \vert.
\end{align*}
\end{assumption}
\looseness -1 Intuitively, $\kappa(\Phi)$ measures the quality of the data: data points that are almost identical decrease $\kappa(\Phi)$, and $\kappa(\Phi)$ is large when data points are diverse.
If the minimum eigenvalue of $\mPhi$ is positive, Assumption \ref{ass:compatibility} is automatically fulfilled.
This type of assumption is common in the literature on representation/meta-learning for sequential decision-making \citep{yang2021impact, cella2021multi,kassraie2022metalearning} and sparse linear bandits \citep{bastani2020online,hao2020high, kim2019doubly}. It is also known in the Lasso literature as the compatibility condition \citep{buhlmann2011statistics}. 
Given these assumptions, we show that \algoff recovers the true kernel with high probability.
\begin{theorem}[Consistency of \algoff]
\label{thm:offline_main_consitency}
    Suppose $D_{1:m}$ satisfies \cref{ass:betamin}~and~\ref{ass:compatibility} with constants $c_1$ and $c_\kappa$ respectively.
    Set $\omega \in (0, c_1)$ and define $\bar \omega = \min \{\omega, c_1-\omega\}$.
    Choose $\lambda = \bar{\omega} c_{\kappa}^2/(8 \sqrt{m})$. 
    Then for \smash{$ \sqrt{n}> 32 \sigma /(\bar \omega c_{\kappa}^2)$}, \algoff satisfies
    \begin{align*}
        \mathbb{P} \left[\Jhat = J^\star \right] \geq 1 - p \exp\left( -m \left(\tfrac{\bar{\omega} c_{\kappa}^2 \sqrt{n}}{32 \sigma} - 1 \right)^2 \right).  
    \end{align*}
    In particular, $\Jhat$ is a consistent estimator both in $n$ and $m$,
    \begin{align*}
        \lim_{n \to \infty} \mathbb{P}\left[ \Jhat = J^\star \right] = 1, \ \text{and} \ \lim_{m \to \infty} \mathbb{P}\left[ \Jhat = J^\star \right] = 1.
    \end{align*}
\end{theorem}
\cref{app:offline} presents the proof to Theorem \ref{thm:offline_main_consitency}.
This theorem shows that our meta-learned kernel converges to $\tk$ as the number of meta-training tasks increases. First, this implies that the meta-learned hypothesis space includes the unknown reward functions
allowing downstream bandit algorithms to provably converge to the optimum.
Second, all candidate kernels $k_j$ that are not active in $k^\star$ are eventually excluded from $\khat$. By excluding all $k_j$ with $j \notin J^\star$ which are not necessary for estimating $f_s \in \calH_{k^\star}$, we effectively shrink the size of the hypothesis space, thereby reducing the uncertainty of the reward function estimates during bandit optimization.
Compared to \smash{$\kfull \coloneqq \tfrac{1}{p}\sum_{j=1}^p k_j$}, which naively uses all kernels, this leads to significant improvements in the query efficiency and performance of the bandit optimization.

\textbf{Comparison with Prior Work.}
\citet{kassraie2022metalearning} propose \textsc{Meta-KeL}, a Lasso-equivalent loss for meta-learning a sparse kernel, given i.i.d. offline data from i.i.d. tasks. 
We emphasize that is not possible to achieve lifelong guarantees by sequentially applying this algorithm. 
\algoff differs from \textsc{Meta-KeL} in key points, and satisfies stronger consistency guarantees: 1) It converges to $k^\star$ as either $n$ the number of samples per task, or $m$ number of tasks grow. In contrast, \textsc{Meta-KeL} converges in $m$ only. 
2) \algoff satisfies the exact recovery guarantee for $k^\star$ since $J^* = \hat{J}$ with high probability. 
While \textsc{Meta-KeL} only guarantees that $J^\star \subset \hat{J}$. This is not sufficient to show that meta-learning improves upon the trivial kernel choice $k_{\mathrm{full}}$.
Both of these properties are required in the lifelong analysis.  \looseness -1
\section{Lifelong Bandit Optimization}
\label{sec:lifelong}
\looseness -1 We now use \algoff as a building block to develop the {\em Lifelong Bandit Optimizer} (\algon), an algorithm for lifelong bandit or Bayesian optimization. 
\algon is paired with a \bba agent which can be instantiated by any kernelized bandit algorithm, e.g., \gpucb \citep{srinivas2009gaussian} or \textsc{GP-TS} \citep{chowdhury2017kernelized}. 
For each task $f_s$, the \bba agent is given the kernel $\khat_{s-1}$ meta-learned on the $s-1$ first tasks. Equipped with the kernel, \bba interacts with the current bandit environment, aiming to optimize its payoff by balancing exploration and exploitation. 

In the lifelong setting, we not only have to explore for the sake of optimizing the current reward function $f_s$, but also we need to make sure to that the sequence of action-reward pairs will be sufficiently informative (in the sense of Assumption \ref{ass:compatibility}) for meta-learning $\hat{k}_s$ in the next stage.
To this end, \algon forces the base agent to select purely exploratory actions for the first $n_s$ steps of the task, by i.i.d. sampling from uniform distribution on $\calX$. 
Following \citet{basu2021no}, we refer to this as {\em forced exploration}
and use $\Dexp_s \coloneqq \{(\bx_{s,i}, y_{s,i}), i\leq n_s\}$ to refer to the collected exploratory data of task $f_s$. We use a decreasing sequence $(n_1, \dots, n_m)$ as detailed below, since less exploration by \bba will be required once more multi-task data is collected. For steps $i > n_s$, \bba selects actions according to its normal bandit policy. 
After the agent has interacted with the current task for $n$ steps, we pass the exploratory data $\Dexp_{1:s}$ to \algoff to meta-learn $\khat_s$. We then announce this new kernel estimate to the \bba agent for solving the next task $s+1$. Figure~\ref{fig:lifelong_algo} visualizes this process and \cref{alg:lifelong} summarizes \algon.

\begin{figure}
    \centering
    \resizebox{\columnwidth}{!}{
        \begin{tikzpicture}[node distance=25mm, roundnode/.style={circle, draw=blue!80, fill=blue!5, very thick, minimum size=7mm},
    squaoraclenode/.style={rectangle, draw=oracle!80, fill=oracle!5, very thick, minimum size=7mm},
    arrow/.style = {thick,-stealth}
    ]
    \draw[ultra thick, rounded corners, dashed, draw=black!80, fill=black!5] (4.5cm, -2.2cm) rectangle (10.5cm, 2.5cm) {};
    \node[rectangle, draw=naive!80, fill=naive!5, very thick, minimum size=7mm, rounded corners=.03cm, align=center] (5) at (6.5cm, -0.6cm) {\bba\\($i > n_s)$};
    \node[squaoraclenode, rounded corners=.03cm] (6) [right of=5] {Environment};
    \node[rectangle, draw=meta!80, fill=meta!5, very thick, minimum size=7mm, rounded corners=.03cm, align=center] (52) at (6.5cm, 1.7cm) {Forced Exploration\\
    $(i \leq n_s)$};
    
    \draw[rectangle, draw=meta!80, fill=meta!5, very thick, minimum size=7mm, rounded corners=.03cm, align=center] (-0.5cm, 0.5cm) rectangle (2cm, 1.5cm) {};    
   \node at (0.75cm,1cm) (ml) {\large{\algoff}};

    \draw[ thick] (52) -- (5);
    \node (h) at (2.25cm, 2.6cm) {\large$\khat_{s-1}$};
    \draw[ultra thick] (0.75cm, 1.5cm) -- (0.75cm, 2.5cm) -- (1.7cm, 2.5cm);
    \draw[arrow, ultra thick] (2.8cm, 2.5cm) -- ++(1.1cm, 0) -- ++(0, -3.1cm) -- (5);
    \draw[arrow, ultra thick] (0.75cm, -0.25) -- (0.75cm, 0.5cm);
    \draw[arrow] (5.north)-- ++(0,0.5cm) -- ++(25mm,0) node[midway, above] {$\bx_{s,i}$} -- (6.north);
    \draw[arrow] (6.south)-- ++(0,-0.5cm) -- ++(-25mm,0) node[midway, below] {$f_s(\bx_{s,i}) + \epsilon_{s,i}$} -- (5.south);
    
    \draw[arrow, ultra thick] (7.5cm, -2.2cm)-- ++(0,-0.8cm) -- ++(-4.2cm,0);
    \draw[rounded corners, draw=oracle!80, fill=oracle!5, very thick] (2.3cm, -3.5cm) rectangle (3.3cm, -2.5cm) {};
    \node at (2.8cm, -3cm) {$\Dexp_{s}$};
    \draw[arrow, ultra thick] (2.3cm, -3cm) -- ++(-1.55cm, 0) -- ++(0, 1cm);
    
    \draw[rounded corners, draw=oracle!80, fill=oracle!5, very thick] (-1.6cm, -1.8cm) rectangle (-0.6cm, -0.5cm) {};
    \node at (-1.1cm, -1.15cm) {$\Dexp_1$};
    \node at (-0.45, -1.15cm) {\large ,};
    
    \draw[rounded corners, draw=oracle!80, fill=oracle!5, very thick] (-0.3 cm, -1.55cm) rectangle (0.7cm, -0.65cm) {};
    \node at (0.2cm, -1.15cm) {$\Dexp_2$};
    
    \node at (0.95, -1.15cm) {\large ,};
    \node at (1.45cm, -1.15cm)  {\large$\dots$};
   \node at (1.9, -1.15cm) {\large ,};
   
    \draw[rounded corners, draw=oracle!80, fill=oracle!5, very thick] (2.1cm, -1.45cm) rectangle (3.1cm, -0.75cm) {};
    \node at (2.6cm, -1.15cm) {$\Dexp_{s-1}$};
    
    \draw [ultra thick, decorate, decoration={calligraphic brace,amplitude=7pt}] (-1.7cm, -2cm) --(-1.7cm, -0.3cm) ; %
    \draw [ultra thick, decorate, decoration={calligraphic brace,amplitude=7pt}] (3.2cm, -0.3cm) --(3.2cm, -2cm) ;

    \end{tikzpicture}
    }
    \caption{Overview of \algon.\looseness-1}
    \label{fig:lifelong_algo}
\end{figure}

\subsection{Regret Bounds}
\looseness-1 Let $\Roracle(n)$ be the worst-case regret of \bba with oracle knowledge of true kernel $\tk$ on single tasks when the reward resides in $\tH$. When employed sequentially on $m$ bandit tasks, the worst-case lifelong regret $R(m,n)$ will be of the order $m \Roracle(n)$ with high probability. We refer to this as oracle regret, since the $\bba$ has access to the true kernel $k^*$ which does not hold in practice. Since our meta-learned kernels $\hat{k}_s$ are an approximations of $k^*$, the oracle regret is a natural lower bound on the regret of \algon.

\looseness -1 In the following, we show that if $\Roracle(n)$ the single-task oracle regret of the base bandit algorithm is sublinear (e.g., as for \gpucb or \textsc{GP-TS}), then so is the lifelong regret $R(m,n)$ of \algon. 
Importantly, $R(m,n)$ is not only sublinear in $n$, but also converges with high probability to $\Roracle(m,n)$.
\cref{thm:lifelong} presents this guarantee, assuming that the forced exploration datasets $\Dexp_s$ satisfy assumption \cref{ass:compatibility} which \algoff requires to yield a provably consistent estimator of $\tk$. Later in \cref{prop:forced_exp}, we show that exploration by i.i.d. sampling from a uniform distribution over $\calX$ will guarantee this assumption.
\begin{theorem}
\label{thm:lifelong}
     For all tasks $s=1, \dots, m$, assume that the reward function $f_s \in \calH_{\tk}$ has bounded RKHS norm $\norm{f_s}_{\tk} \leq B$. Set the number of forced exploration actions as $n_s = \frac{\sqrt{n}}{s^{1/4}}$, and assume that \cref{ass:betamin} and \ref{ass:compatibility} hold for the data $\Dexp_{1:s}$ for all $s = 1, \dots, m$. Suppose, with probability greater than $1-\delta/2$, \bba has worst-case oracle regret $\Roracle(m,n)$.
    Then, the lifelong regret of \algon satisfies
    \begin{equation*}
    \resizebox{\columnwidth}{!}{
    $
        R(m,n) - \Roracle(m,n) = \calO \Big(  \underbrace{Bm^{3/4}\sqrt{n}}_{\text{forced exp.}} + \underbrace{ B (nm)^{1/3} \log^{3/4}(mp/\delta) }_{\text{kernel mismatch} }\Big) 
    $
    }        
    \end{equation*}
    with probability greater than $1-\delta$.
\end{theorem}
The explicit inequality without the $\calO$-notation can be found in Appendix \ref{app:lifelong}, together with the proof.
In the following, we give a sketch of the proof, aiming to explain the source of each term in the bound.
For every forced exploration step, in the worst-case, we suffer regret of $2B$. When accumulated over a total of $\sum_{s=1}^m n_s$ such steps, this gives the first term in the bound.
If $\khat_s \neq \tk$, it is possible to suffer from linear regret in the worse-case. 
To account for this, we calculate the smallest integer $m_0$, for which, with high probability, $\khat_{s} = \tk$ for all $m_0 < s \leq m$. Based on \cref{thm:offline_main_consitency}, we show that $m_0 = \calO((m/n^2)^{1/3}\log^{3/4}(mp/\delta))$.
For every task $s \leq m_0$ we suffer a linear regret of $2Bnm_0$ in the worst-case. This is upper bounded by the second term in \cref{thm:lifelong}, which can be regarded as the cost of learning $\tJ$. Notably, it grows only logarithmically with $p$ the number of considered features/kernels, offering a significant improvement about the polynomial rates given by prior works \citep[e.g.,][]{yang2021impact, hong2022hierarchical}. \cref{tab:litreview1} and \cref{tab:litreview2} present a comprehensive list of the related regret bounds. \looseness-1

\looseness -1 We highlight that the excess regret of \algon in \cref{thm:lifelong} is sublinear in both $m$ and $n$. This implies that the algorithm is {\em oracle optimal}, meaning that as $m\rightarrow \infty$, the single-task regret {\em without} knowledge of $\tk$, eventually approaches the oracle single-task regret. Recall that $\Roracle(m,n) = m \Roracle(n)$ and therefore, $R(m,n)/m \rightarrow \Roracle(n)$.
This guarantee is stronger than that of \citep{basu2021no, pmlr-v151-peleg22a}, where the excess regret depends linearly on $m$ due to excessive forced exploration. By decreasing $n_s \propto s^{-1/4}$ the number of exploratory steps vanishes throughout the sequence of tasks.

As an example, we analyze the performance if \gpucb\footnote{\cref{app:gp_ucb}  provides a background on \gpucb.} \citep{srinivas2009gaussian} is used as the \bba algorithm. In this case, we demonstrate that the worst-case lifelong regret of \algon is of the same rate as the corresponding oracle regret.
To highlight the benefit of this oracle optimality we compare to a naive baseline which uses $\khat_s = \kfull=  \smash{\sum_{j=1}^p \frac{1}{p} k_j} $ for all tasks instead of meta-learning $\khat_s$ sequentially. 
In particular, we consider solving a sequence of $m$ tasks in three scenarios: 1) running \algon paired with \gpucb 2) repeatedly running \gpucb with oracle access to $k^*$, and 3) repeatedly running \gpucb with $\kfull$.
The following corollary shows that the worst-case upper bound for the first two scenarios match in $\calO$-notation. 

\begin{corollary}[Lifelong \gpucb]
\label{cor:ll_ucb_decreasing_main_text}
    Consider the setting of \cref{thm:lifelong} with \gpucb as \bba agent. 
    Then, with probability at least $1-\delta$, the lifelong regret of \algon paired with \gpucb satisfies
    \begin{align*}
        R(m,n) & = \calO\big(\Roracle(m,n)\big)\\
        & = \calO \left( Bm d^\star \sqrt{n}\log\tfrac{n}{d^\star} + m\sqrt{nd^\star \log \tfrac{n}{d^\star}\log\tfrac{ 1}{\delta}}\right)
    \end{align*}
    where $\smash{d^\star \coloneqq \sum_{j \in \tJ} d_j}$. 
\end{corollary}
In the third scenario, we conservatively set $\khat_s = \kfull$ for all $s=1, \dots, m$. While this is sufficient for attaining a lifelong regret that is sublinear in $n$, the performance will \emph{not} be oracle optimal. In particular, this algorithm suffers from a regret of \looseness-1
    \begin{equation*}
    \resizebox{\columnwidth}{!}{
    $
        R(m,n) = \calO\left( B \tfrac{m dp}{\vert \tJ\vert} \sqrt{n}\log\tfrac{n}{d} + \sqrt{nd \log \tfrac{n}{d}\log \tfrac{1}{\delta}}  \right)
        $
        }
    \end{equation*}
where $d = \sum_{j=1}^p d_j \gg d^\star$ and $\smash{p/\vert \tJ\vert}$ can be very large. Our experiments confirm that the performance of the naive approach is significantly worse than the other variants. This is due to the fact that confidence bounds constructed using $\kfull$ tend to contract slower than the ones constructed with the sparse meta-learned $\khat_s$.

\subsection{Forced Exploration} \label{sec:forced_exp}

Our forced exploration scheme ensures that the collected data is sufficiently informative to guarantee successful meta-learning. From a technical perspective, it ensures that \cref{ass:compatibility} is met and allows for a consistent estimator of $\tk$.
The cost of this exploration in the regret of each task is smaller in rate than the minimax regret bound \citep{lattimore2020bandit}. Therefore it has only a negligible effect on the overall performance guarantees of \algon (see \cref{cor:ll_ucb_decreasing_main_text}).
We show that by uniformly drawing actions from the domain, the collected data satisfies this assumption:
\begin{proposition}
\label{prop:forced_exp}
    Assume that $\phi_j \in L^2(\calX)$, $j \in \{1,\dots,p\}$ are orthonormal and let $d_j=1$.
    Draw $\bx_{1}, \dots, \bx_{n_1}$ independently and uniformly from $\calX$, and repeatedly use them to construct $\Dexp_{1:s}$.
    Then with probability at least $1-\delta$, 
     $\Dexp_{1:s}$ satisfies \cref{ass:compatibility},
    for $s = 1,\dots, m$.
\end{proposition}
\looseness -1 The proof can be found in Appendix \ref{app:kappa_bounds}.
 The $d_j=1$ condition is met without loss of generality, by splitting the higher dimensional feature maps and introducing more base features, which will increase $p$. Moreover, the orthonormality condition is met by orthogonalizing and re-scaling the feature maps. Basis functions such as Legendre polynomials and Fourier features \citep{rahimi2007random} satisfy these conditions.

\looseness -1 Generally, it is natural to require \bba to explore more in lifelong setting compared to when it is used in isolation and with a known kernel. We observe in our experiments that \algon has a better empirical performance with forced exploration (i.e., $n_s>0$) than without. 
This additional exploration is also required in the Representation Learning \citep{yang2021impact, yang2022nearly, cella2021multi, cella2022meta} and hierarchical Bayesian bandit literature \citep{basu2021no, pmlr-v151-peleg22a, hong2022hierarchical}, where it is assumed that either the context distribution or the chosen actions are diverse enough. In the case of contextual bandits, if there is sufficient randomness, the \bba can be greedy and yet sample diverse enough actions \citep{bastani2021mostly}. \cref{tab:litreview3} gives a detailed overview of how the related works rely on uniform exploration.
\vspace{-0.5em}
\section{Federated \algon}
\label{sec:federated}

We consider a federated extension of the lifelong learning problem. Here, each BO task is performed by a peer in a network and the corresponding data is not exchanged due to privacy concerns, limited bandwidth, etc. 
Operations are mainly done at the client level, and the central server only performs light computations.
This setting formalizes problems such as optimizing the user experience of a software product on each user's device, e.g., for making better recommendations. 
Limiting the client-server communication 
reduces the transmit overhead time and motivates faster federated computation.
Moreover, sending detailed data on user preferences and interaction patterns to the central server may jeopardize the user's privacy. 
However, we want to harness the statistical patterns across the user pool to improve the automated tailoring of the software product to new users. 
We interpret such a federated learning problem \citep{kairouz2021federated} as a client-server adaptation of our lifelong setting as described in Section \ref{sec:problem}. 
The meta-agent represents the server and BO tasks arise sequentially at a client node $s=1, ..., m$ with a client specific reward function $f_s$.  

We propose the {\em Federated Lifelong Bandit Optimizer} (\falgon) to solve this problem without directly sharing $\calD_s$ the data corresponding to each client with the server. 
\falgon, pairs the clients and the server as follows.  
First, the client node $s$ receives $\khat_{s-1}$, the most recent estimate of the true kernel, and the required number of forced exploration queries $n_s$ from the server.
After taking some exploratory steps, the client performs actions according to its \bba policy.
In contrast to \algon, once the task is over after $n$ steps, the client keeps $\Dexp_s$ to itself, instead of passing it back the server. The client node optimizes for a local loss
\begin{align*}
\vbetahat_s^{(\mathrm{client})} & \coloneqq \argmin_{\vbeta_s \in \sR^{d}} \Ls\left(\vbeta_s; \Dexp_{s} \right) \\
& = \argmin_{\vbeta_s \in \sR^{d}} \frac{1}{n_s} \norm{ \by_s - \mPhi_s \vbeta_s}_2^2  + \lambda \sum_{j=1}^{p} \|\vbeta_s\gj\|_2,
\end{align*}
and calculates a local estimate of $\tJ$ by thresholding $\vbetahat_s^{(\mathrm{client})} $ with the hyperparamter $\omega >0$
\[
\hat J_s^{(\mathrm{client})} \coloneqq \left\{ j \in \{1,\dots,p\}\ \text{s.t.}\ \norm{\vbetahat_s^{(\mathrm{client})}\gj}_2 > \omega \right\}.
\]
It then sends {\em only} the indices $\Jhat_s^{(\mathrm{client})}$ back to the server. 
This leaves the server with the simple task of taking a $\alpha$-majority vote among the $s$ first clients, to decide which base kernels to include in $\hat{k}_s$. Formally, the server chooses 
\[
\khat_{s}(\bx, \bx^\prime)\coloneqq \frac{1}{\vert \hat J_s\vert } \sum_{j \in \hat J_{s}} k_j(\bx, \bx^\prime)
\]
where for $\alpha \in [0,1]$,
\[
\hat J_{s} \coloneqq \left\{  j \in \{1,\dots,p\}\ \text{s.t.}\  \sum_{r =1}^s \mathbbm{1}(j \in \hat J_r^{(\mathrm{client})}) \geq s\alpha\right\}.
\]
In other words, after client $s$ finishes its job, the server includes the $j$-th kernel into its updated estimate $\hat{J}_s$, if and only if more than $s\alpha$ of the clients so far believe that it should be included.
Figure~\ref{fig:lifelong_algo_fed} in the appendix visualizes this process and \cref{alg:lifelong_fed} presents the pseudo-code to \falgon.
Similar to \algon, we show that if $\Roracle(n)$ the worst-case oracle regret of the base bandit algorithm is sublinear in $n$, then so is the lifelong regret $R(n,m)$ of \falgon:
\begin{theorem}
\label{thm:lifelong_federated}
For all tasks $s=1, \dots, m$, assume that the reward function $f_s \in \calH_{\tk}$ has bounded RKHS norm $\norm{f_s}_{\tk} \leq B$. Set the number of forced exploration actions as $n_s = \sqrt{n}$, and assume that \cref{ass:betamin} and \ref{ass:compatibility} hold for the data $\Dexp_s$. 
    Suppose, with probability $>$ $1-\delta/2$, that \bba has worst-case oracle regret $\Roracle(m,n)$. Then the lifelong regret of \falgon satisfies
    \begin{equation*}
    \resizebox{\columnwidth}{!}{
    $
        R(m,n) - \Roracle(m,n) = \calO \Big(Bm\sqrt{n} + B\sqrt{n} \log(mp/\delta)  \Big) 
    $
    }        
    \end{equation*}
    with probability greater than $1-\delta$.
\end{theorem}
\looseness -1 See \cref{app:federated_lifelong} for the proof.  
\cref{thm:lifelong_federated} demonstrates that even without direct access to the data, the lifelong regret of $\falgon$ will be sublinear in $n$. 
This theorem does not imply oracle optimality, since  $R(m,n)/m-\Roracle(n) \not\to 0$ for $m \rightarrow \infty$. 
This is due to the linear dependency of the first term on $m$, which arises from forced exploration. In the federated setting, we require all clients to take a fixed number of exploratory action \smash{$n_s =\sqrt{n}$}, so that they have equal resources for estimating $\tJ$ and the server's majority vote is fair.
We conjecture that with simple modifications, \algon can become provably differentially private. Replacing the majority voting step with GNMax Aggregator \citep{papernot2018scalable} or PRIME \citep{liu2021robust} yields a differential private voting mechanism to select $\hat{J}_s$, while preserving the lifelong regret guarantee.

\begin{figure*}
    \centering
    \begin{subfigure}{0.33\textwidth}
    \centering
         \includegraphics[width=\textwidth]{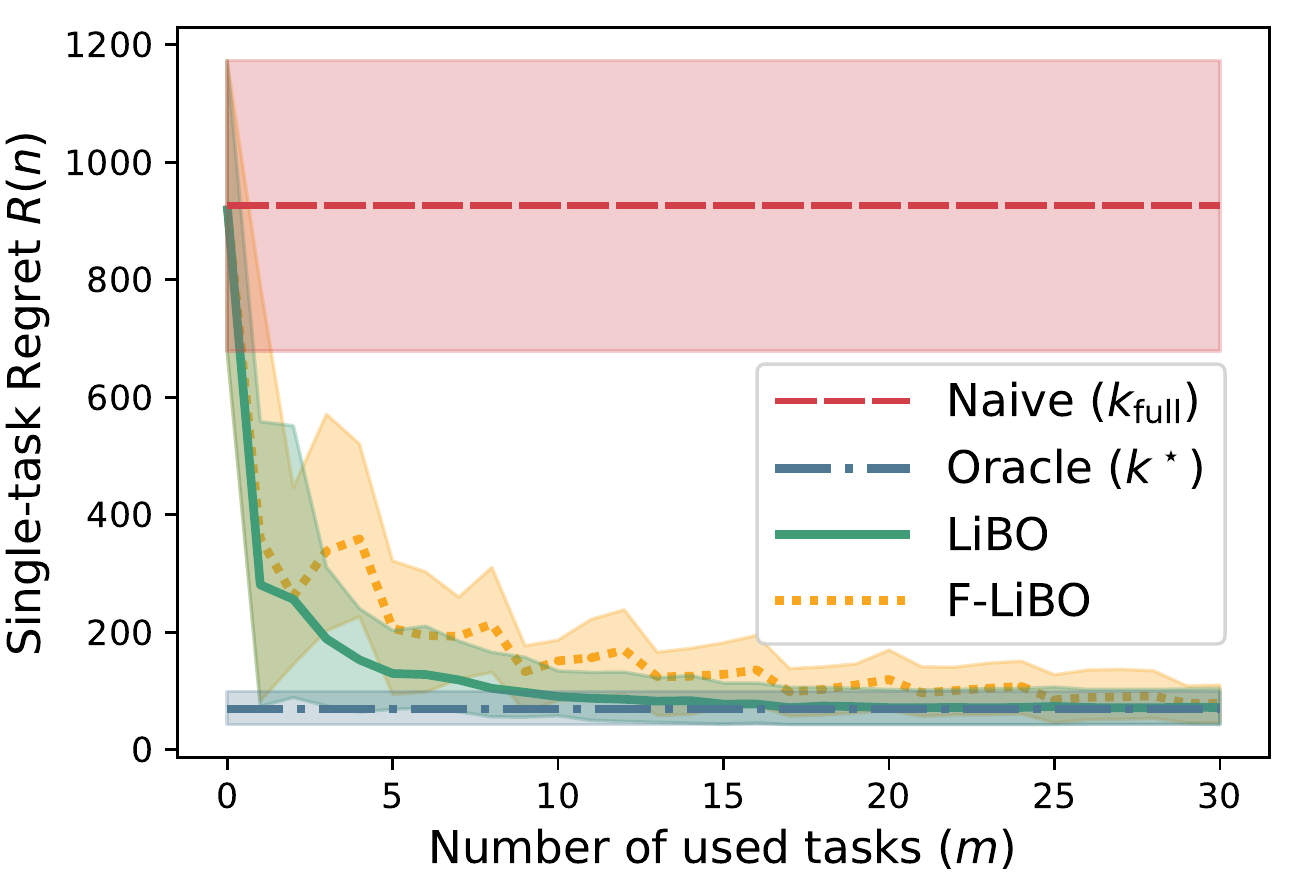}
    \end{subfigure}
    \begin{subfigure}{0.32\textwidth}
        \centering
        \includegraphics[width=\textwidth]{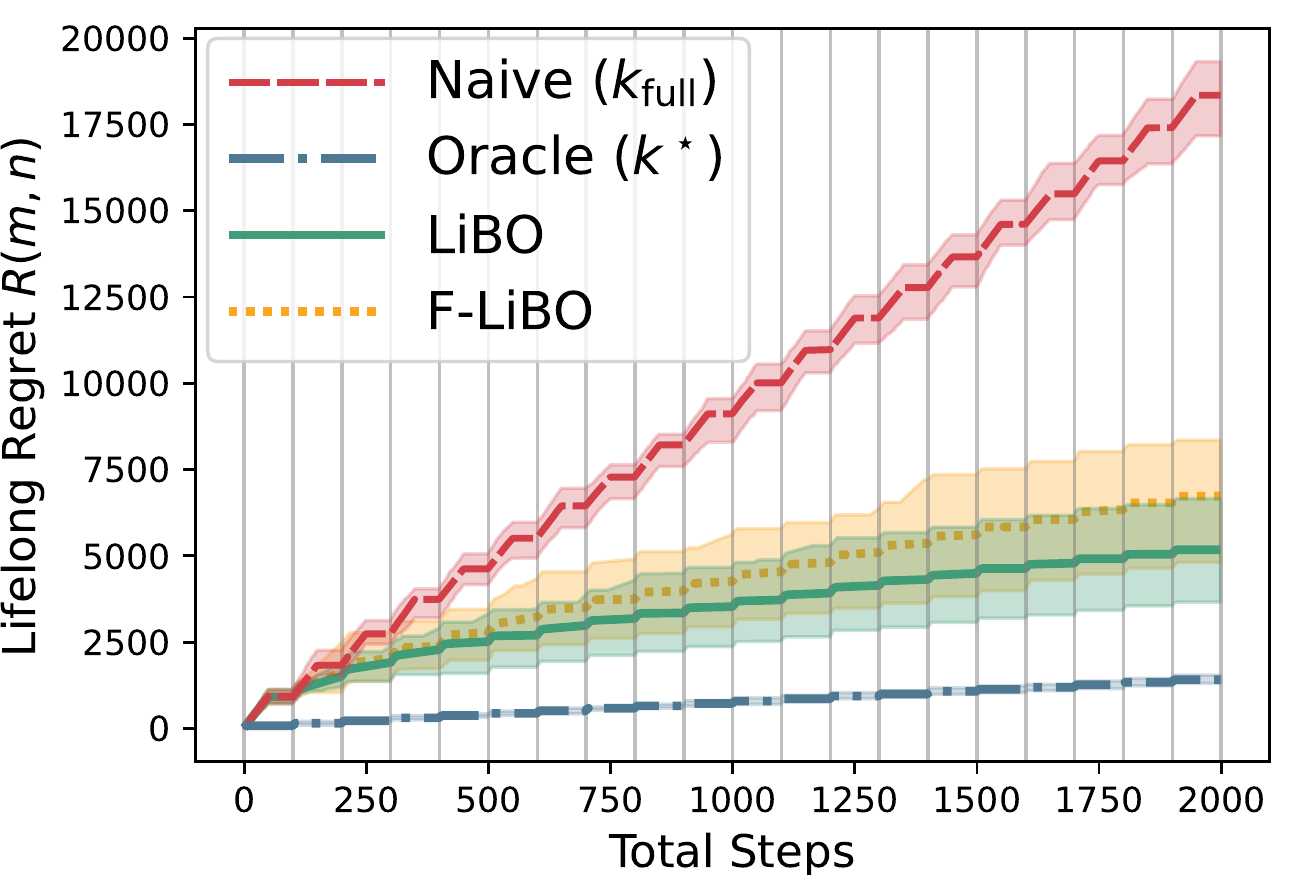}
    \end{subfigure}
    \begin{subfigure}{0.32\textwidth}
        \centering
        \includegraphics[width=\textwidth]{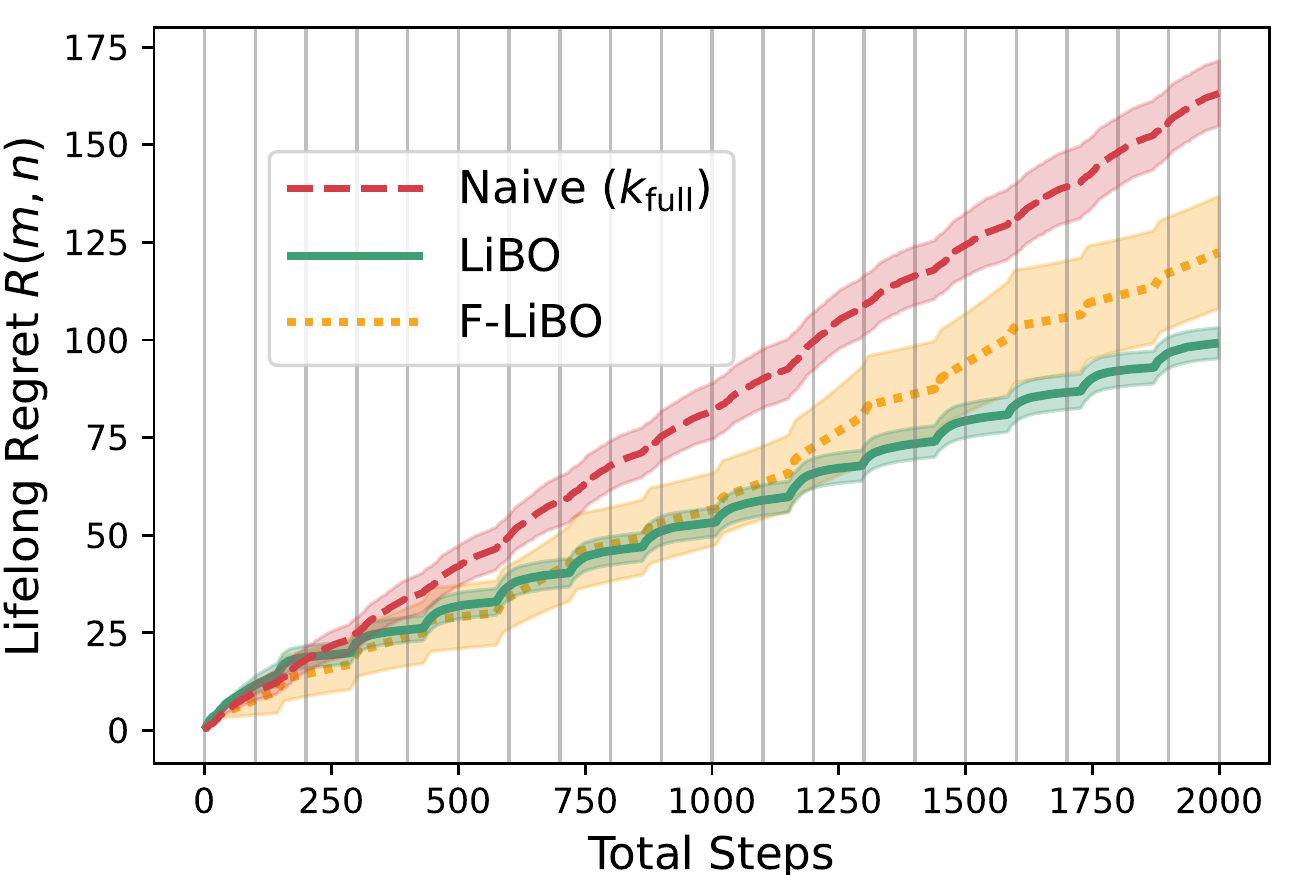}
    \end{subfigure}
    \begin{minipage}[t]{0.37\textwidth}
               \caption{Singe-task cumulative regret of \gpucb with meta-learned kernel $\khat_m$ on an increasing number of meta-training tasks $m$. 
    \label{fig:offline}}
       \end{minipage}
       \hfill
    \begin{minipage}[t]{0.6\textwidth}
               \caption{Lifelong cumulative regret of \gpucb for synthetic tasks (left) and \textsc{GLMNET} hyperparameter tuning (right). Vertical lines indicate the beginning of a new task. \label{fig:online}}    
    \end{minipage}
\end{figure*}

\looseness -1 Consider an example where we instantiate \falgon with \gpucb as \bba. 
The worst-case regret bound of \falgon, which neither has knowledge of $\tk$ nor direct access to $\Dexp_s$, matches the worst-case regret of the oracle \gpucb in $\calO$-notation.
\cref{cor:ucb_federated} formalizes this claim.
Here, $\Roracle(m,n)$ is the same as in \cref{cor:ll_ucb_decreasing_main_text}.
\begin{corollary}[Federated Lifelong \gpucb]
\label{cor:ucb_federated}
    Consider the setting of \cref{thm:lifelong_federated} with \gpucb as \bba. 
    Then, with probability at least $1-\delta$, \falgon paired with \gpucb satisfies 
    \begin{align*}
        R(m,n) & = \calO\big(\Roracle(m,n)\big).
    \end{align*}
\end{corollary}
\section{Experiments} 
\label{sec:experiments}

In all experiments, we use \gpucb as the \bba.
We repeat all experiments with 20 random seeds and report the corresponding mean outcome with standard error.
To evaluate the proposed algorithms, we use a synthetic as well as a hyper-parameter tuning environment.

\textbf{Synthetic environment.} The synthetic environment is based on our data model from Section \ref{sec:problem}.
We choose $\mathcal{X} = [0, 1]$ as the domain and use the first $p=50$ cosine basis functions $\phi_j(x) = \cos(j \pi x)$ as feature maps which form the kernels $k_j(x, x') = \phi_j(x)^\top\phi_j(x^\prime)$ for $j \in \{1, ..., 50\}$. 
The active kernel indices $\tJ$ are sampled uniformly from the set of 5-element subsets of $\{1, ..., 50\}$, i.e., $ |J^\star| = 5$. We sample the reward functions $f_s$ independently and uniformly from $\tH$ such that $||f||_{\tk} \leq 10$ and beta-min condition of $c_1 \geq 0.5$ holds. To the function evaluations we add i.i.d.~Gaussian noise with a standard deviation of $\sigma=0.1$.

\textbf{AutoML data.}
\looseness -1 A common application of Bayesian Optimization is AutoML, i.e., optimizing the hyper-parameters of machine learning algorithms. In this setting, $\calX$ is the learning algorithm's hyper-parameter space, and $f_s$ represents the test performance of the machine learning system. 
In our experiments, we consider a realistic lifelong AutoML setting where we face a sequence of hyper-parameter optimization problems. Here, each task corresponds to tuning the hyper-parameters of the \textsc{GLMNET} learning algorithm \citep{friedman2010regularization} for a different dataset.
Following previous work \citep[e.g.][]{perrone2018scalable, rothfuss2021meta}, we replace the evaluation step by a table lookup based on a large number of hyper-parameter evaluations \citep{kuhn2018automatic} on 38 classification datasets from the OpenML platform \citep{Bischl2017OpenMLBS}.

\subsection{Experiment with Offline Data}
\label{sec:offline_exps}
\looseness -1 We investigate how meta-leaning kernels with \algoff and its federated variant (\falgoff) affects the performance of test tasks. In particular, we use meta-training data that was generated offline, based on the synthetic environment. 
We create data for $m=30$ synthetic tasks, each of size $n=10$ to be used as offline meta-data. The tasks are generated according to our synthetic environment and the details can be found in \cref{app:exp_offline_detail}. Note that $n \ll p$ i.e., we are in the overparameterized setting.
We meta-learn a kernel with \algoff and \falgoff using the meta-training data $\calD_{1:s}$ for $s=1, ..., m$,
and evaluate the estimated kernel $\khat$ by running \gpucb, equipped with $\khat$, for $n=70$ iterations. Figure \ref{fig:offline} illustrates the corresponding single-task regrets in response to increasing the number of meta-training tasks in the offline data.
We report the performance of the agent that uses $\kfull$ (red) as a naive baseline, and the performance of an oracle agent that uses the true kernel (blue) as a natural lower bound for the achievable regret. 
Figure \ref{fig:offline} shows that the regret of both meta-learned agents quickly converges to the regret of the oracle agent as the number of meta-training tasks increases. 
\algoff performs slightly better than \falgoff, since it has direct access to all the data while the federated algorithm loses information during the voting mechanism. 
In \cref{app:experiment_results}, we evaluate \algoff and \falgoff with other choices of base kernels and higher dimensional action domains. Similar to Figure \ref{fig:offline}, we observe fast convergence to the oracle regret. Further details about the experiments are provided in \cref{app:exp_offline_detail}.

\subsection{Lifelong Experiments}
\label{exp:lifelong}
\looseness -1 We return to the lifelong setting where the tasks arrive sequentially and evaluate our algorithms.
We consider both the synthetic and AutoML environments. The horizon of each task is set to $n=100$ time steps. To solve the synthetic problem, we consider the $p=50$ first 1-dimensional cosine bases as the candidate feature maps. Since \textsc{GLMNET} has two hyper-parameters to tune, i.e., $\calX \subset \R^2$, here we use the $p=100$ first 2-dimensional cosine bases, i.e. $\phi_{i,j}(\bx) = \cos(i \pi x_1) \cos(j \pi x_2)$ for $i,j=1, ..., 10$. 

Figure \ref{fig:online} illustrates the cumulative lifelong regrets achieved by \algon, \falgon, the baseline \gpucb with $\kfull$, and oracle \gpucb with access to $\tk$. Note that in the AutoML environment, we do not know the true kernel $\tk$ and thus, cannot report the oracle performance. 
As we would expect, \algon and \falgon initially suffer the same regret as the the naive actor with $\kfull$ since no meta-learning data is available yet. 
However, as more tasks are attempted, the estimated kernel is improved and in turn, the base algorithm becomes more sample efficient on future tasks.
In case of \algon (green), over time, the forced exploration decreases and the estimated kernel converges to the true kernel. As a result, the behavior of the actor paired with the \algon becomes indistinguishable from the actor using the oracle kernel, reflected by the same slope of the regret curves.
Compared to the naive actor (red), our lifelong BO methods significantly improve the efficiency of the base agent as they accumulate more experience. In the AutoML setting, this means that we can find good hyper-parameters with fewer costly function evaluations.
This showcases how incorporating knowledge transfer into deployed machine learning systems
can yield significant performance gains and cost savings.
\begin{table}[t]
\centering
\setlength\tabcolsep{2.5pt}
\begin{tabular}{l|c  |c | c | c | c} 
&   \makecell{oracle\\ optimal} &  \makecell{policy\\agnostic} & \makecell{learns \\ sparsity}  &  \makecell{meta \\cost} & \makecell{tasks}  \\[0.6ex]
 \hline
  \citeauthor{hu2021near} & \xmark & \xmark & \xmark & $\mathrm{poly}\, d$ &   conc\\[0.6ex] 
 \hline
 \citeauthor{yang2021impact} &  \cmark & \xmark &\xmark & $\mathrm{poly}\, d$ & conc\\[0.6ex] 
 \hline
 \hline
\citeauthor{pmlr-v151-peleg22a} & \xmark & \xmark & \xmark & $\mathrm{poly}\, d$ &  seq \\[0.6ex] 
\hline
 \multirow{ 2}{*}{\citeauthor{hong2022hierarchical}} & \cmark & \xmark &  \xmark & $\mathrm{poly}\, d$  & seq  \\[0.6ex] 
 \cline{2-6}
& \xmark &\xmark &  \xmark & $\mathrm{poly}\, d$  & conc \\[0.6ex] 
\hline
\hline
\algon  &   \cmark & \cmark& \cmark & $\log\, d$  & seq 
\end{tabular}
\caption{Related work (\cref{tab:litreview3} gives a comprehensive list.)\looseness-1\label{tab:litreview_main}}
\end{table} 

\vspace{-0.5em}
\section{Related Work}
\label{sec:related_work}

 \looseness -1 The lifelong bandit optimization problem addresses key shortcomings of classic kernelized bandits and Bayesian optimization. 
Early approaches assume that the agent knows the true kernel
\citep{srinivas2009gaussian,valko2013finite,chowdhury2017kernelized}, which is often not the case in practice. 
Recent work addresses this problem, either by studying the implications of misspecified kernels \citep{dylan2020miss,simchowitz2021bayesian, bogunovic2021misspecified,camilleri2021high} or proposing methods for adapting kernel parameters during the optimization \citep{wang2014theoretical, berkenkamp2019no}. 
Alternatively, the appropriate kernel can be learned from related data.
To this end, a number of algorithms are developed for meta-learning a kernelized Gaussian process (GP) prior \citep{harrison2018meta, perrone2018scalable, rothfuss2021pacoh, rothfuss2021meta, rothfuss2022meta}.
However, they come without theoretical guarantees.

Theory of knowledge transfer between concurrent or sequential linear bandits has received recent attention from multiple perspectives.
Representation Learning literature \citep{yang2021impact,hu2021near,yang2022nearly,cella2022meta} assumes existence of a shared low-dimensional linear representation for the reward function, i.e. $f_s(\bx) = \langle \btheta_s, \mB^T\bx\rangle$ where $\mB \in \sR^{d\times d^\star}$ is shared by the tasks. 
This matrix is unknown, however $d^\star$ is known and $d^\star \ll d$.
Feature selection \citep{cella2021multi} takes a similar approach by assuming that $f_s(\bx) = \langle \btheta_s, \mS^T\bx \rangle$, where the unknown matrix $\mS$ screens the relevant features \smash{$\{\bx_j,\,j\in \tJ\}$}. The elements of this matrix are $0$ or $1$, but contrary to representation learning, $d^\star = \vert \tJ \vert$ is unknown. 
Alternatively, works on Bayesian Prior learning  assume existence of a shared Gaussian prior over the parameter vector, i.e. $f_s(\bx) = \langle \btheta_s, \bx\rangle$, where $\btheta_s \sim \calN(\mu, \Sigma)$.  This formulation does not aim for a low-dimensional solution.
Following this model, \citet{basu2021no} and \citet{hong2022hierarchical} assume that $\Sigma$ is known and learn distribution of $\mu$. \citet{pmlr-v151-peleg22a} estimate both the mean and the covariance. We consider a more relaxed setup where the mean is not shared, and meta-learn a shared covariance function. \cref{app:litreview} goes into more depth to formally compare the mentioned work. \looseness-1
 
\looseness-1 We compare \algon with prior algorithms based on the following properties.
In the context of meta-learning for BO, a desirable method is 1) oracle optimal, i.e., attains the regret guarantee of the oracle solver as $m$ grows, 2) able to utilize any BO algorithm, 3) sample efficient, i.e., pays a small cost for meta-learning the prior/relevant features and 4) recovers low-dimensional solutions, since the effective dimension influences the sample efficiency of the base algorithm. \cref{tab:litreview_main} compares \algon with previous work applicable to infinite action domains. Works limited to finite action sets are considered in \cref{tab:litreview3}.
\algon is the only oracle optimal algorithm that learns the effective dimension $d^\star$, while paying a cost that scales only logarithmically with the Euclidean dimension $d$. This is an exponential improvement compared to the polynomial dependency of prior work; moreover, it also applies to reward functions that are a linear combinations of non-linear features $f_s(\bx) = \langle \btheta_s, \bm{\phi}(\bx)\rangle$. Further, it can be wrapped around any linear or kernelized bandit algorithm, while earlier work require a specific bandit policy.

\falgon contributes to recent literature on federated learning which studies how agents can cooperate to solve a single bandit task \citep{dubey2020differentially, shi2021federated, huang2021federated, dai2022federated}. In federated lifelong learning, each agent interacts with a different environment, but collaborates with others to learn relevant features. 

 \looseness -1 Our work builds on ideas from Multiple-Kernel Learning \citep{cristianini2001kernel, bach2004multiple, ong2005learning, xu2010simple, gonen2011multiple} and Multi-Task Lasso \citep{obozinski2006multi, argyriou2006multi, lounici2011oracle} which address consistency of model selection for offline supervised learning. 
Our contribution is lifelong uncertainty quantification, using a meta-learned kernel. \looseness-1
\section{Conclusion}
\looseness -1 We introduce \algon, an algorithm which allows for lifelong knowledge transfer across BO tasks trough meta-learned kernels. We show theoretically and empirically that, if paired with \algon, the performance of a base bandit algorithm improves as more experience is gained on previous tasks.
In particular, we prove that \algon is oracle optimal in the limit.
With \falgon, the federated variant of our main algorithm, we establish that sublinear knowledge transfer is possible even without direct access to the bandit data. 

\looseness -1 This work opens up directions of future research such as quantifying the cost of privacy in Lifelong Learning, understanding the necessity of exploration in lifelong setting, or using large neural networks to extract relevant features from prior tasks instead of working with pre-determined features. 
\begin{acknowledgements}
 This research was supported by the European Research Council (ERC) under the European Union’s Horizon 2020 research and innovation program grant agreement no. 815943. Jonas Rothfuss was supported by the Apple Scholars in AI/ML fellowship. 
 \end{acknowledgements}

\bibliography{refs}

\begin{thebibliography}{58}
\providecommand{\natexlab}[1]{#1}
\providecommand{\url}[1]{\texttt{#1}}
\expandafter\ifx\csname urlstyle\endcsname\relax
  \providecommand{\doi}[1]{doi: #1}\else
  \providecommand{\doi}{doi: \begingroup \urlstyle{rm}\Url}\fi

\bibitem[Abbasi-Yadkori et~al.(2011)Abbasi-Yadkori, P{\'a}l, and
  Szepesv{\'a}ri]{abbasi2011improved}
Yasin Abbasi-Yadkori, D{\'a}vid P{\'a}l, and Csaba Szepesv{\'a}ri.
\newblock Improved algorithms for linear stochastic bandits.
\newblock \emph{NeurIPS}, 24, 2011.

\bibitem[Argyriou et~al.(2006)Argyriou, Evgeniou, and
  Pontil]{argyriou2006multi}
Andreas Argyriou, Theodoros Evgeniou, and Massimiliano Pontil.
\newblock Multi-task feature learning.
\newblock In \emph{NeurIPS}, 2006.

\bibitem[Bach et~al.(2004)Bach, Lanckriet, and Jordan]{bach2004multiple}
Francis~R Bach, Gert~RG Lanckriet, and Michael~I Jordan.
\newblock Multiple kernel learning, conic duality, and the smo algorithm.
\newblock In \emph{ICML}, 2004.

\bibitem[Bastani and Bayati(2020)]{bastani2020online}
Hamsa Bastani and Mohsen Bayati.
\newblock Online decision making with high-dimensional covariates.
\newblock \emph{Operations Research}, 2020.

\bibitem[Bastani et~al.(2021)Bastani, Bayati, and Khosravi]{bastani2021mostly}
Hamsa Bastani, Mohsen Bayati, and Khashayar Khosravi.
\newblock Mostly exploration-free algorithms for contextual bandits.
\newblock \emph{Management Science}, 2021.

\bibitem[Basu et~al.(2021)Basu, Kveton, Zaheer, and Szepesv{\'a}ri]{basu2021no}
Soumya Basu, Branislav Kveton, Manzil Zaheer, and Csaba Szepesv{\'a}ri.
\newblock No regrets for learning the prior in bandits.
\newblock \emph{NeurIPS}, 2021.

\bibitem[Berkenkamp et~al.(2019)Berkenkamp, Schoellig, and
  Krause]{berkenkamp2019no}
Felix Berkenkamp, Angela~P Schoellig, and Andreas Krause.
\newblock No-regret bayesian optimization with unknown hyperparameters.
\newblock \emph{JMLR}, 2019.

\bibitem[Bischl et~al.(2017)Bischl, Casalicchio, Feurer, Hutter, Lang,
  Mantovani, Rijn, and Vanschoren]{Bischl2017OpenMLBS}
B.~Bischl, Giuseppe Casalicchio, Matthias Feurer, F.~Hutter, Michel Lang,
  R.~Mantovani, J.~N. Rijn, and J.~Vanschoren.
\newblock Openml benchmarking suites and the openml100.
\newblock \emph{arXiv preprint}, 2017.

\bibitem[Bogunovic and Krause(2021)]{bogunovic2021misspecified}
Ilija Bogunovic and Andreas Krause.
\newblock Misspecified gaussian process bandit optimization.
\newblock \emph{NeurIPS}, 2021.

\bibitem[B{\"u}hlmann and Van De~Geer(2011)]{buhlmann2011statistics}
Peter B{\"u}hlmann and Sara Van De~Geer.
\newblock \emph{Statistics for high-dimensional data: methods, theory and
  applications}.
\newblock Springer Science \& Business Media, 2011.

\bibitem[Bunea et~al.(2013)Bunea, Lederer, and She]{bunea2013group}
Florentina Bunea, Johannes Lederer, and Yiyuan She.
\newblock The group square-root lasso: Theoretical properties and fast
  algorithms.
\newblock \emph{IEEE Transactions on Information Theory}, 2013.

\bibitem[Camilleri et~al.(2021)Camilleri, Jamieson, and
  Katz-Samuels]{camilleri2021high}
Romain Camilleri, Kevin Jamieson, and Julian Katz-Samuels.
\newblock High-dimensional experimental design and kernel bandits.
\newblock In \emph{ICML}, 2021.

\bibitem[Cella and Pontil(2021)]{cella2021multi}
Leonardo Cella and Massimiliano Pontil.
\newblock Multi-task and meta-learning with sparse linear bandits.
\newblock In \emph{UAI}, 2021.

\bibitem[Cella et~al.(2022)Cella, Lounici, and Pontil]{cella2022meta}
Leonardo Cella, Karim Lounici, and Massimiliano Pontil.
\newblock Meta representation learning with contextual linear bandits.
\newblock \emph{arXiv preprint}, 2022.

\bibitem[Chowdhury and Gopalan(2017)]{chowdhury2017kernelized}
Sayak~Ray Chowdhury and Aditya Gopalan.
\newblock On kernelized multi-armed bandits.
\newblock In \emph{ICML}, 2017.

\bibitem[Cristianini et~al.(2001)Cristianini, Shawe-Taylor, Elisseeff, and
  Kandola]{cristianini2001kernel}
Nello Cristianini, John Shawe-Taylor, Andre Elisseeff, and Jaz Kandola.
\newblock On kernel-target alignment.
\newblock \emph{NeurIPS}, 2001.

\bibitem[Dai et~al.(2022)Dai, Shu, Verma, Fan, Low, and
  Jaillet]{dai2022federated}
Zhongxiang Dai, Yao Shu, Arun Verma, Flint~Xiaofeng Fan, Bryan Kian~Hsiang Low,
  and Patrick Jaillet.
\newblock Federated neural bandit.
\newblock \emph{arXiv preprint}, 2022.

\bibitem[Dubey and Pentland(2020)]{dubey2020differentially}
Abhimanyu Dubey and AlexSandy Pentland.
\newblock Differentially-private federated linear bandits.
\newblock \emph{NeurIPS}, 2020.

\bibitem[Foster et~al.(2020)Foster, Gentile, Mohri, and Zimmert]{dylan2020miss}
Dylan~J Foster, Claudio Gentile, Mehryar Mohri, and Julian Zimmert.
\newblock Adapting to misspecification in contextual bandits.
\newblock In \emph{NeurIPS}, 2020.

\bibitem[Friedman et~al.(2010)Friedman, Hastie, and
  Tibshirani]{friedman2010regularization}
Jerome Friedman, Trevor Hastie, and Rob Tibshirani.
\newblock Regularization paths for generalized linear models via coordinate
  descent.
\newblock \emph{Journal of statistical software}, 2010.

\bibitem[G{\"o}nen and Alpayd{\i}n(2011)]{gonen2011multiple}
Mehmet G{\"o}nen and Ethem Alpayd{\i}n.
\newblock Multiple kernel learning algorithms.
\newblock \emph{JMLR}, 2011.

\bibitem[Hao et~al.(2020)Hao, Lattimore, and Wang]{hao2020high}
Botao Hao, Tor Lattimore, and Mengdi Wang.
\newblock High-dimensional sparse linear bandits.
\newblock \emph{NeurIPS}, 2020.

\bibitem[Harrison et~al.(2018)Harrison, Sharma, and Pavone]{harrison2018meta}
James Harrison, Apoorva Sharma, and Marco Pavone.
\newblock Meta-learning priors for efficient online bayesian regression.
\newblock In \emph{International Workshop on the Algorithmic Foundations of
  Robotics}, 2018.

\bibitem[Hong et~al.(2022)Hong, Kveton, Zaheer, and
  Ghavamzadeh]{hong2022hierarchical}
Joey Hong, Branislav Kveton, Manzil Zaheer, and Mohammad Ghavamzadeh.
\newblock Hierarchical bayesian bandits.
\newblock In \emph{AISTATS}, 2022.

\bibitem[Hu et~al.(2021)Hu, Chen, Jin, Li, and Wang]{hu2021near}
Jiachen Hu, Xiaoyu Chen, Chi Jin, Lihong Li, and Liwei Wang.
\newblock Near-optimal representation learning for linear bandits and linear
  rl.
\newblock In \emph{ICML}, 2021.

\bibitem[Huang et~al.(2021)Huang, Wu, Yang, and Shen]{huang2021federated}
Ruiquan Huang, Weiqiang Wu, Jing Yang, and Cong Shen.
\newblock Federated linear contextual bandits.
\newblock \emph{NeurIPS}, 2021.

\bibitem[Kairouz et~al.(2021)Kairouz, McMahan, Avent, Bellet, Bennis, Bhagoji,
  Bonawitz, Charles, Cormode, Cummings, et~al.]{kairouz2021federated}
Peter Kairouz, H~Brendan McMahan, Brendan Avent, Aur{\'e}lien Bellet, Mehdi
  Bennis, Arjun~Nitin Bhagoji, Kallista Bonawitz, Zachary Charles, Graham
  Cormode, Rachel Cummings, et~al.
\newblock Advances and open problems in federated learning.
\newblock \emph{Foundations and Trends in Machine Learning}, 2021.

\bibitem[Kassraie et~al.(2022)Kassraie, Rothfuss, and
  Krause]{kassraie2022metalearning}
Parnian Kassraie, Jonas Rothfuss, and Andreas Krause.
\newblock {Meta-Learning Hypothesis Spaces for Sequential Decision-making}.
\newblock In \emph{ICML}, 2022.

\bibitem[Kim and Paik(2019)]{kim2019doubly}
Gi-Soo Kim and Myunghee~Cho Paik.
\newblock Doubly-robust lasso bandit.
\newblock \emph{NeurIPS}, 2019.

\bibitem[K{\"u}hn et~al.(2018)K{\"u}hn, Probst, Thomas, and
  Bischl]{kuhn2018automatic}
Daniel K{\"u}hn, Philipp Probst, Janek Thomas, and Bernd Bischl.
\newblock Automatic exploration of machine learning experiments on openml.
\newblock \emph{arXiv preprint}, 2018.

\bibitem[Lattimore and Szepesv{\'a}ri(2020)]{lattimore2020bandit}
Tor Lattimore and Csaba Szepesv{\'a}ri.
\newblock \emph{Bandit algorithms}.
\newblock Cambridge University Press, 2020.

\bibitem[Li et~al.(2019)Li, Wang, and Zhou]{li2019nearly}
Yingkai Li, Yining Wang, and Yuan Zhou.
\newblock Nearly minimax-optimal regret for linearly parameterized bandits.
\newblock In \emph{Conference on Learning Theory}, 2019.

\bibitem[Li et~al.(2021)Li, Wang, Chen, and Zhou]{li2021tight}
Yingkai Li, Yining Wang, Xi~Chen, and Yuan Zhou.
\newblock Tight regret bounds for infinite-armed linear contextual bandits.
\newblock In \emph{AISTATS}, 2021.

\bibitem[Liu et~al.(2021)Liu, Kong, Kakade, and Oh]{liu2021robust}
Xiyang Liu, Weihao Kong, Sham Kakade, and Sewoong Oh.
\newblock Robust and differentially private mean estimation.
\newblock \emph{NeurIPS}, 2021.

\bibitem[Lounici et~al.(2011)Lounici, Pontil, Van De~Geer, and
  Tsybakov]{lounici2011oracle}
Karim Lounici, Massimiliano Pontil, Sara Van De~Geer, and Alexandre~B Tsybakov.
\newblock Oracle inequalities and optimal inference under group sparsity.
\newblock \emph{The annals of statistics}, 2011.

\bibitem[Massias et~al.(2018)Massias, Gramfort, and Salmon]{celer2018}
Mathurin Massias, Alexandre Gramfort, and Joseph Salmon.
\newblock Celer: a fast solver for the lasso with dual extrapolation.
\newblock In \emph{ICML}, 2018.

\bibitem[Obozinski et~al.(2006)Obozinski, Taskar, and
  Jordan]{obozinski2006multi}
Guillaume Obozinski, Ben Taskar, and Michael Jordan.
\newblock Multi-task feature selection.
\newblock \emph{Statistics Department, UC Berkeley, Tech. Rep}, 2006.

\bibitem[Ong et~al.(2005)Ong, Smola, and Williamson]{ong2005learning}
Cheng~Soon Ong, Alexander~J. Smola, and Robert~C. Williamson.
\newblock Learning the kernel with hyperkernels.
\newblock \emph{JMLR}, 2005.

\bibitem[Papernot et~al.(2018)Papernot, Song, Mironov, Raghunathan, Talwar, and
  Erlingsson]{papernot2018scalable}
Nicolas Papernot, Shuang Song, Ilya Mironov, Ananth Raghunathan, Kunal Talwar,
  and Ulfar Erlingsson.
\newblock Scalable private learning with {PATE}.
\newblock In \emph{ICLR}, 2018.

\bibitem[Peleg et~al.(2022)Peleg, Pearl, and Meir]{pmlr-v151-peleg22a}
Amit Peleg, Naama Pearl, and Ron Meir.
\newblock Metalearning linear bandits by prior update.
\newblock In \emph{AISTATS}, 2022.

\bibitem[Perrone et~al.(2018)Perrone, Jenatton, Seeger, and
  Archambeau]{perrone2018scalable}
Valerio Perrone, Rodolphe Jenatton, Matthias~W Seeger, and C{\'e}dric
  Archambeau.
\newblock Scalable hyperparameter transfer learning.
\newblock In \emph{NeurIPS}, 2018.

\bibitem[Rahimi et~al.(2007)Rahimi, Recht, et~al.]{rahimi2007random}
Ali Rahimi, Benjamin Recht, et~al.
\newblock Random features for large-scale kernel machines.
\newblock In \emph{NeurIPS}, 2007.

\bibitem[Rothfuss et~al.(2021{\natexlab{a}})Rothfuss, Fortuin, Josifoski, and
  Krause]{rothfuss2021pacoh}
Jonas Rothfuss, Vincent Fortuin, Martin Josifoski, and Andreas Krause.
\newblock {PACOH: Bayes-optimal meta-learning with PAC-guarantees}.
\newblock In \emph{ICML}, 2021{\natexlab{a}}.

\bibitem[Rothfuss et~al.(2021{\natexlab{b}})Rothfuss, Heyn, Krause,
  et~al.]{rothfuss2021meta}
Jonas Rothfuss, Dominique Heyn, Andreas Krause, et~al.
\newblock {Meta-learning Reliable Priors in the Function Space}.
\newblock In \emph{NeurIPS}, 2021{\natexlab{b}}.

\bibitem[Rothfuss et~al.(2022)Rothfuss, Koenig, Rupenyan, and
  Krause]{rothfuss2022meta}
Jonas Rothfuss, Christopher Koenig, Alisa Rupenyan, and Andreas Krause.
\newblock Meta-learning priors for safe bayesian optimization.
\newblock In \emph{Conference on Robot Learning (CoRL)}, 2022.

\bibitem[Russo and Van~Roy(2014)]{russo2014learning}
Daniel Russo and Benjamin Van~Roy.
\newblock Learning to optimize via information-directed sampling.
\newblock \emph{NeurIPS}, 2014.

\bibitem[Shi et~al.(2021)Shi, Shen, and Yang]{shi2021federated}
Chengshuai Shi, Cong Shen, and Jing Yang.
\newblock Federated multi-armed bandits with personalization.
\newblock In \emph{AISTATS}, 2021.

\bibitem[Simchowitz et~al.(2021)Simchowitz, Tosh, Krishnamurthy, Hsu, Lykouris,
  Dudik, and Schapire]{simchowitz2021bayesian}
Max Simchowitz, Christopher Tosh, Akshay Krishnamurthy, Daniel~J Hsu, Thodoris
  Lykouris, Miro Dudik, and Robert~E Schapire.
\newblock Bayesian decision-making under misspecified priors with applications
  to meta-learning.
\newblock \emph{NeurIPS}, 2021.

\bibitem[Srinivas et~al.(2010)Srinivas, Krause, Kakade, and
  Seeger]{srinivas2009gaussian}
Niranjan Srinivas, Andreas Krause, Sham Kakade, and Matthias Seeger.
\newblock Gaussian process optimization in the bandit setting: No regret and
  experimental design.
\newblock In \emph{ICML}, 2010.

\bibitem[Thompson(1933)]{thompson1933likelihood}
William~R Thompson.
\newblock On the likelihood that one unknown probability exceeds another in
  view of the evidence of two samples.
\newblock \emph{Biometrika}, 1933.

\bibitem[Vakili et~al.(2021)Vakili, Khezeli, and
  Picheny]{vakili2021information}
Sattar Vakili, Kia Khezeli, and Victor Picheny.
\newblock On information gain and regret bounds in gaussian process bandits.
\newblock In \emph{AISTATS}. PMLR, 2021.

\bibitem[Valko et~al.(2013)Valko, Korda, Munos, Flaounas, and
  Cristianini]{valko2013finite}
Michal Valko, Nathaniel Korda, R{\'e}mi Munos, Ilias Flaounas, and Nelo
  Cristianini.
\newblock Finite-time analysis of kernelised contextual bandits.
\newblock \emph{arXiv preprint}, 2013.

\bibitem[Vershynin(2018)]{vershynin2018high}
Roman Vershynin.
\newblock \emph{High-dimensional probability: An introduction with applications
  in data science}, volume~47.
\newblock Cambridge university press, 2018.

\bibitem[Wang and de~Freitas(2014)]{wang2014theoretical}
Ziyu Wang and Nando de~Freitas.
\newblock Theoretical analysis of bayesian optimisation with unknown gaussian
  process hyper-parameters.
\newblock \emph{arXiv preprint}, 2014.

\bibitem[Xu et~al.(2010)Xu, Jin, Yang, King, and Lyu]{xu2010simple}
Zenglin Xu, Rong Jin, Haiqin Yang, Irwin King, and Michael~R Lyu.
\newblock Simple and efficient multiple kernel learning by group lasso.
\newblock In \emph{ICML}, 2010.

\bibitem[Yang et~al.(2021)Yang, Hu, Lee, and Du]{yang2021impact}
Jiaqi Yang, Wei Hu, Jason~D. Lee, and Simon~Shaolei Du.
\newblock Impact of representation learning in linear bandits.
\newblock In \emph{ICLR}, 2021.

\bibitem[Yang et~al.(2022)Yang, Lei, Lee, and Du]{yang2022nearly}
Jiaqi Yang, Qi~Lei, Jason~D Lee, and Simon~S Du.
\newblock Nearly minimax algorithms for linear bandits with shared
  representation.
\newblock \emph{arXiv preprint}, 2022.

\bibitem[Zhao and Yu(2006)]{zhao2006model}
Peng Zhao and Bin Yu.
\newblock On model selection consistency of lasso.
\newblock \emph{JMLR}, 2006.

\end{thebibliography}

\onecolumn

\newpage

\appendix
\numberwithin{equation}{section}
\section{Pseudo-codes to Algorithms}
\begin{algorithm}[ht]
\caption{\algoff}
\label{alg:meta_learning}
\begin{algorithmic}
\Require Data from previous tasks $\Dexp_{1:s}$, threshold parameter $\omega > 0$
\State $\hat{\vbeta} \gets \min_{\vbeta \in \sR^{sd}} \Ls\left(\vbeta ; \Dexp_{1:s}\right)$ \Comment{solves \cref{eq:meta_loss}}
\State $\hat{J} \gets \{ j \leq p \ \vert \ \norm{\hat{\vbeta}^{(j)}}_2 \geq \omega \sqrt{s}\}$
\State $\hat{k}_s \gets \frac{1}{|\hat{J}|} \sum_{j \in \hat{J}} k_j$
\end{algorithmic}
\end{algorithm}

\begin{algorithm}
\caption{\algon}
\label{alg:lifelong}
\begin{algorithmic}
\Require $n,m \in \mathbb{N}$, $0 < \omega < c_1$, \bba %
\State $\hat{k}_0 \gets \sum_{j=1}^p \frac{1}{p} k_j$

\For{$s \in \{1, \dots,m\}$}
    \State $\Dexp_s \gets \emptyset$ \Comment{Dataset for kernel prediction}
    \State $\calD_s \gets \emptyset$ \Comment{Dataset for the base bandit algorithm}
    \For{$i \in \{1, \dots, n\}$}
        \If{$i \leq \frac{\sqrt{n}}{s^{1/4}}$} \Comment{Forced exploration with rate $\sqrt{n}/s^{1/4}$}
            \State Sample $\bx_{s,i}$ uniformly from $\calX$
            \State Play action $\bx_{s,i}$ and observe $y_{s,i}$
            \State $\Dexp_s \gets \Dexp_s \cup \{(\bx_{s,i}, y_{s,i})\}$ \Comment{Add to kernel prediction dataset}
        \Else 
            \State $\bx_{s,i} \gets \text{\bba}(\hat{k}_{s-1})$ \Comment{Select action using base bandit algorithm}
            \State Play action $\bx_{s,i}$ and observe $y_{s,i}$
        \EndIf
        \State $\calD_s \gets \calD_s \cup \{(\bx_{s,i}, y_{s,i})\}$
        \State Update \bba using $\calD_s$ \Comment{Update base bandit algorithm}
    \EndFor
    \State $\khat_{s} \gets \text{\algoff}(\Dexp_{1:s}, \omega)$\Comment{Update $\hat{k}$ using \algoff and $\Dexp_s$}
\EndFor

\end{algorithmic}
\end{algorithm}

\begin{algorithm}[ht]
\caption{\falgon}
\label{alg:lifelong_fed}
\begin{algorithmic}
\Require $n,m \in \mathbb{N}$, $0 < \omega < c_1$, \bba
\State $\hat{k}_0 \gets \sum_{j=1}^p \frac{1}{p} k_j$

\For{$s \in \{1, \dots,m\}$}
    \State $\Dexp_s \gets \emptyset$ \Comment{Dataset for kernel prediction}
    \State $\calD_s \gets \emptyset$ \Comment{Dataset for the base bandit algorithm}
    \For{$i \in \{1, \dots, n\}$}
        \If{$i \leq \sqrt{n}$} \Comment{Forced exploration with rate $\sqrt{n}$}
            \State Sample $\bx_{s,i}$ uniformely from $\calX$
            \State Play action $\bx_{s,i}$ and observe $y_{s,i}$
            \State $\Dexp_s \gets \Dexp_s \cup \{(\bx_{s,i}, y_{s,i})\}$ \Comment{Add to kernel prediction dataset}
        \Else 
            \State $\bx_{s,i} \gets \text{\bba}(\hat{k}_{s})$ \Comment{Select action using base bandit algorithm}
            \State Play action $\bx_{s,i}$ and observe $y_{s,i}$
        \EndIf
        \State $\calD_s \gets \calD_s \cup \{(\bx_{s,i}, y_{s,i})\}$
        \State Update \bba using $\calD_s$ \Comment{Update base bandit algorithm}
    \EndFor
    \State $\khat_{s} \gets \text{\falgoff}(\Dexp_{1:s}, \omega)$\Comment{Update $\hat{k}$ using \falgoff and $\Dexp_s$}
\EndFor

\end{algorithmic}
\end{algorithm}

\begin{algorithm}[H]
\caption{\falgoff}
\label{alg:offline_federated}
\begin{algorithmic}
\Require $n,m \in \mathbb{N}$, data for each task $\calD_s$, $\omega > 0$, $\alpha \in [0,1]$
\State $\mathrm{count}_1, \dots, \mathrm{count}_p \gets 0$
\For{$s \in \{1, \dots,m\}$}
    \State $\hat{\vbeta}_s \gets \min_{\vbeta \in \sR^{d}} \Ls\left(\vbeta ; \Dexp_{s}\right)$ 
    \For{$j \in \{1, \dots,p\}$}
        \State $\mathrm{count}_j \gets \mathrm{count}_j + \mathbbm{1}{\{ \norm{\hat{\vbeta}^{(j)}_s}_2 \geq \omega \}}$
    \EndFor
\EndFor
\State $\hat{J}\gets \{ j \leq p \ \vert \ \mathrm{count}_j \geq  m\alpha\}$
\State $\hat{k} \gets \frac{1}{|\hat{J}|} \sum_{j \in \hat{J}} k_j$
\end{algorithmic}
\end{algorithm}

\section{Extended Literature Review}\label{app:litreview}
In this section, we present an overview of works that consider learning a potentially low-dimensional reward function by leveraging data of similar bandit tasks.

\paragraph{Linear Contextual Bandits with Shared Representation.} The common assumption here is that the reward function for all $s \in [m]$, is linear $f_s(\bx) = \langle \btheta_s, \bx\rangle$ where $\btheta_s = \mB\bw_s$. The matrix $\mB \in \sR^{d\times r}$ is a shared representation matrix and $r\ll d$ is an intrinsic dimension.  This assumption becomes more intuitive if we re-write the reward as  $f_s(\bx) = \langle \bw_s, \mB^T\bx\rangle$, which implies that there exists a mapping $\mB: \sR^d \rightarrow \sR^r$ that produces a low-dimensional representation of the actions.
Our reward assumption implies that there exists a sparse matrix $\mS \in \sR^{d\times d^\star}$ which satisfies $f_s(\bx) = \langle \btheta_s, \mS^T\vphi(\bx)\rangle$ and screens the relevant features $\vphi_j$ with $j \in \tJ$. 
The intrinsic dimension $r$ then corresponds to $\vert \tJ \vert$. 

Recent work on shared representation learning, often consider the contextual setting, where at every step of the bandit problem, actions may only be chosen from a set $\calA_{s, t}$. Once the action is chosen, a noisy reward is observed. Regarding the occurrence of the tasks, two scenarios are often studied. The multi-task setting where all the tasks are solved concurrently, and the lifelong setting where the tasks arrive consecutively. \cref{tab:litreview1} summarizes these efforts in terms of the obtained regret bounds. Here, $\tilde \calO$ hides polylogarithmic factors.
With the exception of \citep{hu2021near}, these works either 1) require forced exploration to fulfill sufficient exploration assumptions (SE) similar to \cref{{ass:compatibility}}, or 2) design a greedy algorithm assuming that the actions in set $\calA_{s,t}$ are sampled from a diverse context distribution (DC) which gives free exploration \citep[c.f.][]{bastani2021mostly}. This suggests that for minimax optimality, either the algorithm has to explore, or the presented context should induce exploration for free.
To better understand the tightness of the results in \cref{tab:litreview1}, we recall that the oracle solver which has knowledge of the representation matrix $\mB$, has a lower-bound of $R^\star(n)= \Omega(\sqrt{rn\log n \log k})$, when $|\calA_{s,t}|=k$ \citep{li2019nearly}. If $\calA_{s,t}$ is an ellipsoid, the lower-bound achievable by the oracle is $\Omega(r\sqrt{n})$ \citep{li2021tight}.  Clearly, for $r \ll d$, the algorithms of \cite{hu2021near}, \cite{cella2021multi}, and \cite{cella2022meta} do not converge to the oracle solver as $m\rightarrow \infty$, since $R(m,n)/m \not\to R^\star(n)$. 

\begin{table}[ht]
\centering
\begin{tabular}{l| l| l |l |l |c | c} 
  & \myalign{c|}{$\calA_{s,t}$} &  \myalign{c|}{$r$} & \myalign{c|}{Tasks} & \myalign{c|}{Expected Lifelong Regret} & Base Policy & Assumptions \\ [0.6ex] 
 \hline
 \multirow{ 2}{*}{\cite{yang2021impact} }
& finite & known & conc.& $\tilde \calO \left(m\sqrt{rn} + \sqrt{rdnm}\right)$ & Greedy & DC\\ [0.6ex] 
& ellipsoid & known & conc.& $\tilde \calO \Big(mr\sqrt{n} + d^{3/2}r\sqrt{nm}\Big)$ & ETC & SE\\ [0.6ex] 
\hline
\cite{hu2021near} & compact & known & conc.& \cellcolor{notopt!70} $\tilde \calO \Big(m\sqrt{drn} + d\sqrt{rnm}\Big)$ & OFUL & - \\[0.6ex] 
 \hline
\cite{cella2022meta} & finite &  unknown & seq. & \cellcolor{notopt!70}$\tilde \calO \left(mr\sqrt{n} + r\sqrt{dnm}\right)$ & Greedy & DC \\[0.6ex] 
 \hline
\cite{cella2021multi} & finite & unknown & seq./conc. & \cellcolor{notopt!70} $\tilde \calO \Big(mr\sqrt{n}\Big)$ & Greedy & DC \& SE \\[0.6ex]
 \hline 
\cite{yang2022nearly} & compact &  known & seq. & $\tilde \calO \Big(mr\sqrt{n} + d\sqrt{rnm}\Big)$ & ETC & SE\\[0.6ex] 
 \hline
\algon (Ours)  & compact &  unknown & seq. & $\tilde \calO \Big(mr\sqrt{n} + m^{3/4}\sqrt{n} \Big)$ & any & SE\\[0.6ex] 
\end{tabular}
\caption{Overview of recent work on representation learning for contextual linear bandits. Oracle lower-bound is $R^\star(m,n)=\Omega\left(mr\sqrt{n}\right)$ for infinite action set, and $R^\star(m,n)=\Omega\left(m\sqrt{rn\log n\log k}\right)$ for finite set. Polylog terms are not included.\label{tab:litreview1}}
\end{table}

\paragraph{Bayesian Bandits with Shared Prior Distributions.} Alternatively, some works consider a Bayesian reward model, but without any assumption on sparsity, or low-dimensional representations. Let $f_s(\bx) = \langle \btheta_s, \bx\rangle$ where $\btheta_s$ are i.i.d.~from $\calN(\bm{\mu}, \Sigma)$ and the parameters $(\bm{\mu}, \Sigma)$ are shared across all tasks.
\cite{pmlr-v151-peleg22a} assume that $(\bm{\mu}, \Sigma)$ are unknown, and estimate it using the exploratory action-reward pairs collected during the first $m_0$ tasks. The suggested meta-algorithm can be wrapped around any Quasi-Bayesian base policy, such as Thomspon Sampling \citep{thompson1933likelihood} or Information Directed Sampling \citep{russo2014learning}, however, the resulting algorithm over-explores as indicated by the $\tilde \calO(md)$ term in the regret bound (See \cref{tab:litreview2}).\looseness -1

Taking a hierarchically Bayesian approach, \cite{basu2021no} and \cite{hong2022hierarchical} further assume that  $\bm{\mu} \sim \calN(\bm{\mu}_0, \Sigma_0)$ where $\bm{\mu}_0$ is unknown, but both covariance matrices $\Sigma$ and $\Sigma_0$ are known. Prior distribution of $\bm{\mu}$ is updated after each task, according to the evidence collected during the task. Both papers design a meta-algorithm with Thomspon Sampling as the base solver. While \cite{basu2021no} suffers from over-exploration, \cite{hong2022hierarchical} does not require any exploration. Indeed, if $\Sigma$ the covariance matrix between the actions is known, it helps with inferring rewards of other actions, and reduces the need for uniform exploration.

An overview is given in \cref{tab:litreview2}, here $R^\star(m,n)$ indicates the Bayesian lifelong regret of the oracle agent who has knowledge of $(\bm{\mu}, \Sigma)$. Note that \cref{thm:lifelong} gives slightly stronger result, which is a high-probability bound over the regret. Here, we have taken the average to make it comparable with the Bayes regret reported in other works. As $m$ grows, the average single-task regret is upper bounded by $R(m,n)/m$, implying that only \cite{hong2022hierarchical} and \algon can converge to the oracle solver.

\begin{table}[ht]
\centering
\begin{tabular}{l| c| c |l |l |c | c} 
  & $\Sigma \slash\Sigma_0$ &  sparse & \myalign{c|}{Tasks} & \myalign{c|}{Bayesian Regret} & Policy & Exp \\[0.6ex] 
 \hline
\cite{basu2021no} & known & no & seq. & $ \cellcolor{notopt!70}\tilde \calO \left(R^\star(m,n) + \sqrt{dnm} + md \right)$ & TS & yes \\[0.6ex] 
 \hline
\cite{pmlr-v151-peleg22a} & unknown &  no & seq. & $ \cellcolor{notopt!70}\tilde \calO \Big((1+d^3/\sqrt{m}) R^\star(m,n)+ md\Big)$ & any QB & yes\\[0.6ex] 
\hline
 \multirow{ 2}{*}{\cite{hong2022hierarchical}} & known & no & seq. &  $\tilde \calO \left(R^\star(m,n) + \sqrt{dmn} + d^{3/2} \right)$ & TS & no\\[0.6ex] 
& known & no & conc.& \cellcolor{notopt!70} $\tilde \calO \left(R^\star(m,n) + \sqrt{dmn} + md^{3/2}\right)$ & TS & yes   \\
\hline
\algon (Ours) & - & yes & seq. & $\tilde \calO \Big(R^\star(m,n) + m^{3/4}\sqrt{n} + (mn)^{1/3}\log (md)\Big)$ & any & yes \\[0.6ex] 
\end{tabular}
\caption{Overview of recent work on meta-learning Bayesian priors for linear bandits \label{tab:litreview2}. All works consider compact action set, except for \cite{basu2021no} which requires a finite set of actions selected from $\sR^d$. The regret of the oracle solver is denoted by $R^\star(m,n)$.}
\end{table}

\paragraph{Overall Landscape of Research.} \label{par:overview}We merge the two lines of work in \cref{tab:litreview3}, to give an overview of ongoing efforts on meta-learning for linear bandits and the important properties of each method.
Column $\vert \calA_t \vert$ shows if the model holds for compact action sets, or only finite ones.
Column ``sparse'' and ``learns $r$'' denote whether the model aims for sparse solution, and if it requires knowledge of the true sparsity $r$ or preferably, it learns it.
Column `` learns $\Sigma/\Sigma_0$'' only applies to Bayesian method, where some assume the covariance matrix of $\btheta_s$ is known, and some estimate it from data.
Column ``Tasks'' shows if the method considers simultaneous or sequentially appearing bandit tasks.
Column ``O-opt'' refers to oracle optimality, and has a checkmark only if $R(m,n)/m \rightarrow R^\star(n)$.
Column ``MS Cost'' shows the cost of model selection/meta-learning. In particular, shows whether the additional regret, incurred due to not knowing the true representation/features, is logarithmic in dimension $d$ or is it polynomial. 
Column ``Policy'' shows the base BO solvers that can be paired with the meta-learning algorithm, ``any'' indicates that the method can work with any (linear) policy, and ``any QB'' refers to Quasi-Bayesian methods.
Column ''Ass.'' shows the assumptions of the method on diversity of data, SE indicates Sufficient Exploration type assumptions, and DC refers to Diverse Context assumptions. 
Column ''Has no uniform draws'' shows if the algorithm requires forces exploration or not.

\begin{table}[ht]
\centering
\begin{tabular}{l| c| c |c |c |c | c | c | c|c|c} 
 & $\vert\calA_t\vert$ & Sparse &\makecell{Learns \\ $r$} & \makecell{Learns\\ $\Sigma \slash\Sigma_0$} &  Tasks & O-opt & MS cost & Policy & Ass. & \makecell{ Has no \\unif. \\ draws} \\[0.6ex] 
 \hline
 \multirow{ 2}{*}{\cite{yang2021impact} } & $k$ & \cmark & \xmark & - & conc & \cmark & $\mathrm{poly}(d)$ & greedy & DC & \xmark \\[0.6ex] 
 \cline{2-11}
 & $\infty$ & \cmark & \xmark & - & conc &\cmark & $\mathrm{poly}(d)$ & ETC & SE & \xmark \\[0.6ex] 
 \hline
 \cite{hu2021near} & $\infty$ & \cmark & \xmark & - & conc & \xmark & $\mathrm{poly}(d)$ &  OFUL & - & \cmark \\[0.6ex] 
 \hline
 \cite{cella2021multi} & $k$ & \cmark &  \cmark &- & both & \xmark & $\log(d)$ & greedy & \makecell{DC\\SE} & \xmark \\[0.6ex] 
 \hline
 \cite{cella2022meta} & $k$ & \cmark &  \cmark &- & seq & \xmark & $\mathrm{poly}(d)$ & greedy & DC &  \cmark\\[0.6ex] 
 \hline
 \cite{yang2022nearly} & $\infty$ & \cmark & \xmark & - & conc & \cmark & $\mathrm{poly}(d)$ & ETC & SE & \xmark \\[0.6ex] 
 \hline
 \hline
\cite{basu2021no} & $k$ & \xmark & - &  \xmark & seq &\xmark & $\mathrm{poly}(d)$ & TS & SE & \xmark \\[0.6ex] 
 \hline
\cite{pmlr-v151-peleg22a} & $\infty$ & \xmark & - & \cmark & seq & \xmark & $\mathrm{poly}(d)$ & any QB & SE & \xmark\\[0.6ex] 
\hline
 \multirow{ 2}{*}{\cite{hong2022hierarchical}} & $\infty$ & \xmark & - &  \xmark & seq & \cmark & $\mathrm{poly}(d)$ & TS & - & \cmark \\[0.6ex] 
 \cline{2-11}
& $\infty$ & \xmark &   - & \xmark & conc & \xmark & $\mathrm{poly}(d)$ & TS & 
 SE & \xmark \\
\hline
\hline
\algon (Ours) &  $\infty$ & \cmark &  \cmark & - & seq & \cmark & $\log(d)$ & any & SE & \xmark\\[0.6ex] \cline{2-11}
\falgon (Ours) & $\infty$ & \cmark &  \cmark & - & seq & \cmark & $\log(d)$ & any & SE & \xmark
\end{tabular}
\caption{Collective pro and cons of related works: \algon gives an overall pareto-optimal solution. Refer to the corresponding paragraph in \cref{par:overview} for information on meaning of each column. \cref{tab:litreview_main} presents a concise version. \label{tab:litreview3}}
\end{table}
\section{Generality of the Kernel Assumption}\label{app:wlog}
In \cref{sec:problem}, we claim that, the average of kernels formulation, i.e.
\[
\tk(\cdot, \cdot) =\frac{1}{\vert \tJ\vert} \sum_{j \in \tJ} k_j(\cdot, \cdot)
\]
is without loss of generality equivalent to assuming a linear combination,
\[
\tk(\cdot, \cdot) = \sum_{j \in \tJ}\alpha_j k_j(\cdot, \cdot).
\]
Here, we formally show this claim.
Assume there exist $\alpha_1, \dots, \alpha_p \in \mathbb{R}$ and kernels $k_1,\dots, k_p$ such that
\begin{align*}
    \tk(\bx,\bx^\prime) = \sum_{j \in J^\star} \alpha_j k_j(\bx,\bx^\prime), \quad \forall \bx, \bx^\prime \in \calX.
\end{align*}

Let $f \in \mathcal{H}_{\tk}$, then there exists $\beta_1 \in \mathbb{R}^{d_1}, \dots, \beta_p \in \mathbb{R}^{d_p}$ such that for all $x \in \mathcal{X}$
\begin{align*}
    f(\bx) = \sum_{j \in J^\star} \sqrt{\alpha_j} \beta_j^{\top} \phi_j(\bx).
\end{align*}

Define $m_j \coloneqq \max\{k(\bx, \bx) \, \mid \, \bx \in \mathcal{X} \}$, $\Tilde{\beta}_j \coloneqq p m_j \beta_j \sqrt{\alpha_j}$ and $\Tilde{k}_j \coloneqq k_j / m_j$ for all $j \in \{1,\dots,p\}$, then
\begin{align*}
    f(\bx) = \frac{1}{p} \sum_{j \in J^\star} \Tilde{\beta}_j^{\top} \phi_j(\bx)
\end{align*}
and therefore $f \in \mathcal{H}_{\Tilde{k}}$ for
\begin{align*}
    \Tilde{k}^\star(\bx,\bx') = \frac{1}{|J^\star|} \sum_{j \in J^\star} k_j(\bx,\bx^\prime).
\end{align*}

This shows that the corresponding Reproducing Kernel Hilbert Spaces are equivalent, i.e. the same functions reside in both, while the norm is scaled. Therefore, we can assume, without loss of generality, that the base kernels are normalized and that the true kernel is an average of base kernels.
\section{Consistency of \algoff (Proof of \texorpdfstring{Theorem~\ref{thm:offline_main_consitency}}{})} \label{app:offline}
We start by proving the necessary lemmas. During this section we assume a slightly more general setting. More precisely, we assume that we have $n_s \leq n$ samples in task $s$, which means that the total samples size of the meta-dataset is $N = N_m \coloneqq \vert \Dexp \vert = \sum_{s=1}^m n_s$.

\begin{definition}[sub-Gaussian random variables]
    Let $X$ be a random variable. We call $X$ a \emph{$\sigma$ sub-Gaussian random variable} if $\mathbb{E}[X] = 0$ and
    \begin{equation}
        \mathbb{P}[|X| \geq t] \leq 2\exp\left(-\frac{t^2}{2\sigma^2}\right).
    \end{equation}
\end{definition}

\begin{lemma}[Theorem 6.3.2 of \citet{vershynin2018high}]
\label{thm:concentration}
    Let $\epsilon_1, \dots, \epsilon_n$ be independent, zero mean, unit variance sub-Gaussian random variables. Define $\epsilon = (\epsilon_1, \dots, \epsilon_n)$. Let $A \in \sR^{m \times n}$ and $t \geq 0$. Then
    \begin{align*}
        \mathbb{P}\left( \Big \vert \norm{A\epsilon}_2 - \norm{A}_F \Big \vert \geq t \right) \leq \exp\left( -\frac{t^2}{2\norm{A}^2_2} \right).
    \end{align*}
\end{lemma}

\begin{corollary}
\label{cor:concentration}
    Let $\epsilon_1, \dots, \epsilon_n$ be $i.i.d.$~$\sigma$ sub-Gaussian random variables and define $\epsilon = (\epsilon_1, \dots, \epsilon_n)$. Let $A \in \sR^{m \times n}$ and $t \geq \sigma \sqrt{\trace(AA^T)}$. Then
    \begin{align*}
        \mathbb{P}\left( \norm{A\epsilon}_2 \geq t \right) \leq \exp\left( -\frac{\left(t/\sigma - \sqrt{\trace(AA^T)}\right)^2}{2\norm{AA^T}_2} \right).
    \end{align*}
\end{corollary}

\begin{proof}
    The standard deviation of an $\sigma$ sub-Gaussian random variable is smaller equal $\sigma$. Therefore
    \begin{align*}
        \mathbb{P}\left( \norm{A\epsilon}_2 \geq t \right)
        & = \mathbb{P}\left( \norm{A\epsilon}_2 / \sqrt{Var(\epsilon_1)}\geq t/ \sqrt{Var(\epsilon_1)} \right)\\
        & \leq \mathbb{P}\left( \norm{A\epsilon}_2 / \sqrt{Var(\epsilon_1)}\geq t/ \sigma \right).
    \end{align*}
    It holds that $\norm{A}_F = \sqrt{\trace(AA^T)}$. Define $\tilde{\epsilon} = \epsilon / \sqrt{Var(\epsilon_1)}$. We have
    \begin{align*}
        \mathbb{P}\left( \norm{A\epsilon}_2 \geq t \right)
        & \leq \mathbb{P}\left( \norm{A\tilde{\epsilon}}_2 - \norm{A}_F \geq t/\sigma - \sqrt{\trace(AA^T)}\right)\\
        & \leq \mathbb{P}\left( \vert \norm{A\tilde{\epsilon}}_2 - \norm{A}_F \vert \geq t/\sigma - \sqrt{\trace(AA^T)}\right).
    \end{align*}
    Using \cref{thm:concentration} and noting that $\norm{A}^2_2 = \norm{AA^T}_2$ yields the desired result.
\end{proof}

\begin{lemma}
\label{lem:3_1}
    Consider the model in \cref{eq:meta_loss} with $\sigma$ sub-Gaussian $i.i.d.$~noise.
    Then, for $\frac{\lambda N}{4\sigma} > \sqrt{\trace(\mPhi^{(j)} (\mPhi^{(j)})^T)}$ with probability at least
    \begin{align*}
        1 - \sum_{j=1}^p \exp\left( -\frac{\left(\frac{\lambda N}{4\sigma} - \sqrt{\trace(\mPhi^{(j)} (\mPhi^{(j)})^T)}\right)^2}{2\norm{\mPhi^{(j)}(\mPhi^{(j)})^T}_2} \right)    
    \end{align*}
    we have for any solution $\hat{\vbeta}$ of \cref{eq:meta_loss}
    \begin{align*}
        \frac{1}{N} \norm{\mPhi (\hat{\vbeta} - \vbeta^\star)}_2^2 + &\frac{\lambda}{2} \sum_{j=1}^p \norm{\hat{\vbeta}^{(j)} - \vbeta^{\star(j)}}_2 \leq \\
        &2\lambda \sum_{j \in J^\star} \min \left( \norm{\hat{\vbeta}^{(j)} - \vbeta^{\star(j)}}_2 ,\norm{\vbeta^{\star(j)}}_2 \right).
    \end{align*}
\end{lemma}

\begin{proof}
    This proof is inspired by the proof of Lemma 3.1 in \citet{lounici2011oracle}.\\
    For all solutions $\hat{\vbeta}$ of \cref{eq:meta_loss}
    \begin{align*}
        \hat{\vbeta} = \argmin_{\vbeta \in \sR^{dm}} \frac{1}{N} \norm{\mPhi \vbeta^\star + \epsilon - \mPhi \vbeta}_2^2 + \lambda \sum_{j=1}^p \norm{\vbeta^{(j)}}_2.
    \end{align*}
    Therefore for all $\vbeta \in \sR^{dm}$
    \begin{align*}
        \frac{1}{N} \norm{\mPhi \vbeta^\star + \epsilon - \mPhi \hat{\vbeta}}_2^2 + \lambda \sum_{j=1}^p \norm{\hat{\vbeta}^{(j)}}_2 \leq \frac{1}{N} \norm{\mPhi \vbeta^\star + \epsilon - \mPhi \vbeta}_2^2 + \lambda \sum_{j=1}^p \norm{\vbeta^{(j)}}_2.
    \end{align*}
    This yields 
    \begin{align*}
        \frac{1}{N} \norm{\mPhi (\hat{\vbeta} - \vbeta)}_2^2 \leq&  \frac{1}{N} \norm{\mPhi(\vbeta - \vbeta^\star)}_2^2 +\frac{2}{N} \epsilon^T \mPhi(\hat{\vbeta} - \vbeta) \\
        &+ \lambda \sum_{j=1}^p \left( \norm{\vbeta^{(j)}}_2 - \norm{\hat{\vbeta}^{(j)}}_2 \right).
    \end{align*}
    And in particular if $\vbeta = \vbeta^\star$, then,
    \begin{align*}
        \frac{1}{N} \norm{\mPhi (\hat{\vbeta} - \vbeta^\star)}_2^2 \leq& \frac{2}{N} \epsilon^T \mPhi(\hat{\vbeta} - \vbeta^\star) \\
        &+ \lambda \sum_{j=1}^p \left( \norm{\vbeta^{\star(j)}}_2 - \norm{\hat{\vbeta}^{(j)}}_2 \right).
    \end{align*}
    By \cref{cor:concentration} and union bound we have jointly for all $j \leq p$ with probability at least 
    \begin{align*}
        1 - \sum_{j=1}^p \exp\left( -\frac{\left(\frac{\lambda N}{4\sigma} - \sqrt{\trace(\mPhi^{(j)} (\mPhi^{(j)})^T)}\right)^2}{2\norm{\mPhi^{(j)}(\mPhi^{(j)})^T}_2} \right)
    \end{align*}
    that
    \begin{align*}
    \label{eq:conc}
        \norm{(\epsilon^T \mPhi)^{(j)}}_2 \leq \frac{\lambda N}{4}.
    \end{align*}
    Therefore, by Cauchy-Schwarz,
    \begin{align*}
        \epsilon^T \mPhi(\hat{\vbeta} - \vbeta^\star) &\leq \sum_{j=1}^p \norm{(\epsilon^T \mPhi)^{(j)}}_2 \norm{\hat{\vbeta}^{(j)} - \vbeta^{\star(j)}}_2\\
        & \leq \frac{\lambda N}{4} \sum_{j=1}^p \norm{\hat{\vbeta}^{(j)} - \vbeta^{\star(j)}}_2.
    \end{align*}
    This implies that
    \begin{align*}
        \frac{1}{N} \norm{\mPhi (\hat{\vbeta} - \vbeta^\star)}_2^2 \leq \frac{\lambda}{2} \sum_{j=1}^p \norm{\hat{\vbeta}^{(j)} - \vbeta^{\star(j)}}_2  + \lambda \sum_{j=1}^p \left( \norm{\vbeta^{\star(j)}}_2 - \norm{\hat{\vbeta}^{(j)}}_2 \right).
    \end{align*}
    Therefore
    \begin{align*}
        \frac{1}{N} \norm{\mPhi (\hat{\vbeta} - \vbeta^\star)}_2^2 + &\frac{\lambda}{2} \sum_{j=1}^p \norm{\hat{\vbeta}^{(j)} - \vbeta^{\star(j)}}_2 \leq \\
        &\lambda \sum_{j=1}^p \left( \norm{\hat{\vbeta}^{(j)} - \vbeta^{\star(j)}}_2 + \norm{\vbeta^{\star(j)}}_2 - \norm{\hat{\vbeta}^{(j)}}_2 \right)
    \end{align*}
    and since $\vbeta^{\star(j)} = 0$ for all $j \notin J^\star$
    \begin{align*}
        \frac{1}{N} \norm{\mPhi (\hat{\vbeta} - \vbeta^\star)}_2^2 + &\frac{\lambda}{2} \sum_{j=1}^p \norm{\hat{\vbeta}^{(j)} - \vbeta^{\star(j)}}_2 \leq \\
        &2\lambda \sum_{j \in J^\star} \min \left( \norm{\hat{\vbeta}^{(j)} - \vbeta^{\star(j)}}_2 ,\norm{\vbeta^{\star(j)}}_2 \right).
    \end{align*}
    This proves the statement.
\end{proof}

\begin{lemma}
\label{thm:meta_kel_general}
    Let Assumption \ref{ass:betamin} hold. If $\hat{\vbeta}$ is a solution of \cref{eq:meta_loss} then
    we have for $\frac{\lambda N}{4\sigma} > \sqrt{\trace(\mPhi^{(j)} (\mPhi^{(j)})^T)}$ with probability at least
    \begin{align*}
        1 - \sum_{j=1}^p \exp\left( -\frac{\left(\frac{\lambda N}{4\sigma} - \sqrt{\trace(\mPhi^{(j)} (\mPhi^{(j)})^T)}\right)^2}{2\norm{\mPhi^{(j)}(\mPhi^{(j)})^T}_2} \right)    
    \end{align*}
    that
    \begin{align*}
        \sum_{j \notin J^\star} \norm{\hat{\vbeta}^{(j)} - \vbeta^{\star(j)}}_2 \leq 3 \sum_{j \in J^\star} \norm{\hat{\vbeta}^{(j)} - \vbeta^{\star(j)}}_2,
    \end{align*}
    and
    \begin{equation}
        \sum_{j=1}^p \norm{\hat{\vbeta}^{(j)} - \vbeta^{\star(j)}}_2 \leq \frac{8 \lambda m}{\bar{\kappa}^2 },
    \end{equation}
    where
    \begin{equation}
        \bar{\kappa} \coloneqq \frac{\sqrt{m}}{\sqrt{N}} \frac{\norm{\mPhi (\hat{\vbeta} - \vbeta^\star)}_2}{\sum_{j \in J^\star} \norm{\hat{\vbeta}^{(j)} - \vbeta^{\star(j)}}_2}.
    \end{equation}
\end{lemma}

\begin{proof}
    Lemma \ref{lem:3_1} implies
    \begin{align}
    \label{eq_uufg}
        \sum_{j=1}^p \norm{\hat{\vbeta}^{(j)} - \vbeta^{\star(j)}}_2 \leq 4 \sum_{j \in J^\star} \norm{\hat{\vbeta}^{(j)} - \vbeta^{\star(j)}}_2
    \end{align}
    and therefore
    \begin{align*}
        \sum_{j \notin J^\star} \norm{\hat{\vbeta}^{(j)} - \vbeta^{\star(j)}}_2 \leq 3 \sum_{j \in J^\star} \norm{\hat{\vbeta}^{(j)} - \vbeta^{\star(j)}}_2,
    \end{align*}
    which yields the first statement of  \cref{thm:meta_kel_general}.\\
    Again, by the first statement of  \cref{lem:3_1}, we have with probability at least
    \begin{align*}
        1 - \sum_{j=1}^p \exp\left( -\frac{\left(\frac{\lambda N}{4\sigma} - \sqrt{\trace(\mPhi^{(j)} (\mPhi^{(j)})^T)}\right)^2}{2\norm{\mPhi^{(j)}(\mPhi^{(j)})^T}_2} \right)    
    \end{align*}
    that
    \begin{equation}
        \norm{\mPhi (\hat{\vbeta} - \vbeta^\star)}_2 \leq \sqrt{2\lambda N} \sqrt{\sum_{j \in J^\star} \norm{\hat{\vbeta}^{(j)} - \vbeta^{\star(j)}}_2}.
    \end{equation}
    Therefore
    \begin{align*}
        \sum_{j \in J^\star} \norm{\hat{\vbeta}^{(j)} - \vbeta^{\star(j)}}_2
        &\leq \frac{\sum_{j \in J^\star} \norm{\hat{\vbeta}^{(j)} - \vbeta^{\star(j)}}_2}{\norm{\mPhi (\hat{\vbeta} - \vbeta^\star)}_2} \norm{\mPhi (\hat{\vbeta} - \vbeta^\star)}_2\\
        & \leq \frac{\sum_{j \in J^\star} \norm{\hat{\vbeta}^{(j)} - \vbeta^{\star(j)}}_2}{\norm{\mPhi (\hat{\vbeta} - \vbeta^\star)}_2} \sqrt{2\lambda N} \sqrt{\sum_{j \in J^\star} \norm{\hat{\vbeta}^{(j)} - \vbeta^{\star(j)}}_2}.
    \end{align*}
    Solving this yields
    \begin{align*}
        \sum_{j \in J^\star} \norm{\hat{\vbeta}^{(j)} - \vbeta^{\star(j)}}_2
        \leq \frac{2\lambda m}{\bar{\kappa}^2},
    \end{align*}
    and by \cref{eq_uufg} we have
    \begin{align*}
        \sum_{j=1}^p \norm{\hat{\vbeta}^{(j)} - \vbeta^{\star(j)}}_2 &\leq 4 \sum_{j \in J^\star} \norm{\hat{\vbeta}^{(j)} - \vbeta^{\star(j)}}_2\\
        & \leq \frac{8 \lambda m}{\bar{\kappa}^2 }.
    \end{align*}
    
\end{proof}

\begin{definition}[compatibility variable]
\label{def:comvar_app}
    Let
    \begin{align*}
        S \coloneqq \Big\{&(J,b) \subset \mathcal{P}(\{1,\dots,p\}) \times (\sR^d \backslash \{0\}) \ \Big\vert \ |J| \leq s^\star, \sum_{j \notin J} \|b^{(j)}\|_2 \leq 3 \sum_{j \in J} \|b^{(j)}\|_2 \Big\}.
    \end{align*}
    For $n,d \in \mathbb{N}$ and $\mPhi \in \sR^{N\times d}$ we define $\kappa(\mPhi)$ by
    \begin{align*}
        \kappa(\mPhi) \coloneqq \inf_{(J,b) \in S} \frac{\sqrt{m}}{\sqrt{N}} \frac{\|\mPhi b\|_2}{ \sum_{j \in J} \|b^{(j)}\|_2}.
    \end{align*}
    and call it the compatibility variable of $\mPhi$.
\end{definition}

\begin{remark}
\label{rem:kappa}
    It holds that $\kappa \leq \bar{\kappa}$.
\end{remark}

\begin{corollary}
\label{cor:main}
    Let $0< \omega < c_1$ and let $\hat{\vbeta}$ is a solution of \cref{eq:meta_loss} with
    \begin{align*}
        \lambda \leq \frac{\bar{\omega} \kappa^2 }{8 \sqrt{m}},
    \end{align*}
    where $\bar{\omega} \coloneqq \min \{ \omega, c_1 - \omega\}$.
    Then we have for $\frac{\lambda N}{4\sigma} > \sqrt{\trace(\mPhi^{(j)} (\mPhi^{(j)})^T)}$ with probability at least
    \begin{align*}
        1 - \sum_{j=1}^p \exp\left( -\frac{\left(\frac{\lambda N}{4\sigma} - \sqrt{\trace(\mPhi^{(j)} (\mPhi^{(j)})^T)}\right)^2}{2\norm{\mPhi^{(j)}(\mPhi^{(j)})^T}_2} \right) 
    \end{align*}
    that
    \begin{equation}
        \frac{1}{\sqrt{m}} \sum_{j=1}^p \norm{\hat{\vbeta}^{(j)} - \vbeta^{\star(j)}}_2 \leq \bar{\omega}.
    \end{equation}
    If additionally Assumption \ref{ass:betamin} holds, then we have with the same probability for
    \begin{align*}
        \hat{J} \coloneqq \left\{ j \in \{1,\dots, p\} \ \Big\vert \ \norm{\hat{\vbeta}^{(j)}}_2 > \omega \sqrt{m} \right\}
    \end{align*}
    that
    \begin{align*}
        \hat{J} = J^\star.
    \end{align*}
\end{corollary}

\begin{proof}
    The first statement follows directly from \cref{thm:meta_kel_general}.\\
    Assume $j \in J^\star$. Then by Assumption \ref{ass:betamin}
    \begin{align*}
        \norm{\hat{\vbeta}^{(j)}}_2 > \sqrt{m} \Big( c_1 - \norm{\hat{\vbeta}^{(j)} - \vbeta^{\star(j)}}_2 \Big)\geq \sqrt{m} (c_1 - \bar{\omega}) \geq\omega \sqrt{m},
    \end{align*}
    which implies $J^\star \subset \hat{J}$.
    Assume $j \notin J^\star$, then
    \begin{align*}
        \norm{\hat{\vbeta}^{(j)}}_2 \leq \norm{\hat{\vbeta}^{(j)} - \vbeta^{\star(j)}}_2 +  \norm{\vbeta^{\star(j)}}_2 \leq \bar{\omega}\sqrt{m} \leq \omega \sqrt{m},
    \end{align*}
    which implies $\hat{J} \subset J^\star$
\end{proof}

\begin{remark}
    Choosing $\omega$ optimally yields $\omega = \bar{\omega} = c_1/2$.
\end{remark}

\begin{proof}[\textbf{Proof of \cref{thm:offline_main_consitency}}]
    Note that $\mPhi^{(j)} \in \sR^{N \times md_j}$ is block-diagonal. Since by assumption $k_j(x,x') \leq 1, \forall j \leq p$ we have
    \begin{align*}
        \trace\left((\mPhi^{(j)})^T \mPhi^{(j)}\right) = \trace\left(\mPhi^{(j)} (\mPhi^{(j)})^T\right) = \sum_{s=1}^m \sum_{i=1}^{n_s} k_j\left(x^{(s)}_i, x^{(s)}_i\right) \leq N,
    \end{align*}
    and
    \begin{align*}
        \norm{(\mPhi^{(j)})^T \mPhi^{(j)}}_2
        &= \norm{\mPhi^{(j)} (\mPhi^{(j)})^T}_2 \\
        &\leq \max_{s\leq m} \trace\left(\mPhi_s^{(j)} (\mPhi_s^{(j)})^T\right)\\
        &\leq \max_{s\leq m} \sum_{i=1}^{n_s}  k_j\left(x^{(s)}_i, x^{(s)}_i\right)\\
        &\leq n.
    \end{align*}
    Corollary \ref{cor:main} yields the result.
\end{proof}

\section{Lifelong Analysis (Proof of \texorpdfstring{Theorem~\ref{thm:lifelong}}{})} \label{app:lifelong}

We start by proving a generic variant of \cref{thm:lifelong}, from which we can obtain the theorem in the main text as a corollary.

\begin{theorem}
\label{thm:lifelong_main}
    Assume that the true reward functions $f_1, \dots, f_m$ satisfy $\norm{f_i}_{\tH} \leq B$ for some constant $B > 0$. Assume $\{n_s\}_{s\in\mathbb{N}}$ is a non-increasing sequence with $n_s \leq n, \forall s$. Define $N_m \coloneqq \sum_{s=1}^m n_s$. Let $\nu$ be a distribution on $\calX^{N}$ independent of $\bm{\epsilon}_1, \dots, \bm{\epsilon}_m$. Let $V \sim \nu$ be the random vector used for forced exploration. Let $\tilde{\mPhi}_s \in \sR^{N_s \times md}$ be the data matrix obtained by forced exploration. Assume the forced exploration distribution $\nu$ and $\{k_j\}_{j\leq p}$ are such that, with probability at least $1-\delta/4$, there exists $c_{\kappa} > 0$ such that $\kappa(\tilde{\mPhi}_s) \geq c_{\kappa}, \forall s \leq m$.
    Assume further that \bba using the true kernel function for $m$ tasks with $n$ interactions with independent noise achieves with probability at least $1-\delta/2$ cumulative regret lower than $\Roracle(m, n)$ in the worst-case. Then, for $m_0 \in \mathbb{N}$ and $0 < \delta < 1$, if
    \begin{align*}
        N_{m_0} \geq \frac{2n_1 \log(4mp/\delta)}{(\sqrt{N_m/m} \frac{c_1c_{\kappa}^2}{32\sigma} - 1)^2},
    \end{align*}
    with probability at least $1-\delta$ \algon achieves
    \begin{align*}
        R(m,n) \leq 2B m_0 n + 2B N_m + \Roracle(m, n).
    \end{align*}
\end{theorem}

\begin{proof}
    Denote by $C \coloneqq \{v \ \vert \ \kappa(\tilde{\mPhi}_s(v)) \geq c_{\kappa}, \forall s \leq m \}$ the set of data points such that $\kappa$ is lower bounded by $c_{\kappa}$. By assumption we have $\mathbb{P}[V \in C] \geq 1 -\delta/4$.
    Denote by $\hat{J}_{s} \subset \{1,\dots,p\}$ the sparsity structure predicted by \algon after the first $s$ tasks. 
    Note that $\tilde{\mPhi}^{(j)}_s \in \sR^{n_s m \times md_j}$ is block-diagonal. Since by assumption $k_j(x,x') \leq 1, \forall j \leq p$ we have
    \begin{align*}
        \trace((\tilde{\mPhi}^{(j)}_s)^T \tilde{\mPhi}^{(j)}_s) = \trace(\tilde{\mPhi}^{(j)}_s (\tilde{\mPhi}_s^{(j)})^T) = \sum_{s=1}^m \sum_{i=1}^{n_s} k_j(x^{(s)}_i, x^{(s)}_i) \leq N,
    \end{align*}
    and
    \begin{align*}
        \norm{(\tilde{\mPhi}^{(j)}_s)^T \tilde{\mPhi}^{(j)}_s}_2
        &= \norm{\tilde{\mPhi}^{(j)}_s (\tilde{\mPhi}^{(j)}_s)^T}_2 \\
        &\leq \max_{s\leq m} \trace(\mPhi_s^{(j)} (\mPhi_s^{(j)})^T)\\
        &\leq \max_{s\leq m} \sum_{i=1}^{n_s}  k_j(x, x)\\
        &\leq \max_{s\leq m} n_s\\
        &= n_1.
    \end{align*}
    Since $V$ is independent of $\bm{\epsilon}_1, \dots, \bm{\epsilon}_m$, we have by Corollary \ref{cor:main} for all $s$ and $v' \in C$
    \begin{align*}
        \mathbb{P}\left[  \hat{J}_{s} = J^\star \mid V = v' \right] \geq 1 - p\exp\left(- \frac{N_s}{2n_1} \left( \sqrt{\frac{N_s}{s}}\frac{c_1 c_{\kappa}^2}{32 \sigma} -1\right)^2 \right).
    \end{align*}
    By union bound and since $\frac{N_s}{s}$ is non-increasing by assumption we have for $m_0 \leq m$
    \begin{align*}
        \mathbb{P}\left[ \forall m \geq s \geq m_0, \hat{J}_s = J^\star \mid V = v'  \right] &\geq 1 - \sum_{s=m_0}^m p\exp\left(- \frac{N_s}{2n_1} \left( \sqrt{\frac{N_s}{s}}\frac{c_1 c_{\kappa}^2}{32 \sigma} -1\right)^2 \right)\\
        & \geq 1 - m p\exp\left(- \frac{N_{m_0}}{2n_1} \left( \sqrt{\frac{N_m}{m}}\frac{c_1 c_{\kappa}^2}{32 \sigma} -1\right)^2 \right),
    \end{align*}
    where we defined $N_s \coloneqq \sum_{i=1}^s n_i$.
    If $m_0 \in \mathbb{N}$ is large enough such that
    \begin{align*}
        N_{m_0} \geq \frac{2n_1 \log(4mp/\delta)}{(\sqrt{N_m/m} \frac{c_1c_{\kappa}^2}{32\sigma} - 1)^2},
    \end{align*}
    then for all $v' \in C$
    \begin{align*}
        \mathbb{P}\left[ \forall m \geq s \geq m_0, \hat{J}_s = J^\star \mid V = v'  \right] &\geq 1 - \delta/4.
    \end{align*}
    By assumption we have
    \begin{align*}
        \mathbb{P}\left[ V \in C \right] \geq 1 - \delta/4.
    \end{align*}
    Because $V$ is independent of the noise
    \begin{align*}
        \mathbb{P}\left[  \exists m \geq s \geq m_0 , \hat{J}_s \neq J^\star \right] &= \int \mathbb{P}\left[ \exists m \geq s \geq m_0 ,\hat{J}_s \neq J^\star \mid V=v\right] p_V(v)dv\\
        &= \int_C \mathbb{P}\left[ \exists m \geq s \geq m_0,\hat{J}_s \neq J^\star \mid V=v\right] p_V(v)dv +\\
        & \quad \int_{C^c} \mathbb{P}\left[ \exists m \geq s \geq m_0,\hat{J}_s \neq J^\star \mid V=v\right] p_V(v)dv\\
        & \leq \mathbb{P} [V \in C^c] + \mathbb{P} [V \in C] \delta/2\\ 
        & \leq \delta/4 + \delta/4 = \delta/2.
    \end{align*}
    For all tasks that happen after task $m_0$ we have jointly with probability at least $1-\delta/2$ that $\hat{J}_s = J^\star$. \\
    Denote by $r(k, s)$ the regret that the base bandit algorithm achieves after $n$ interactions with kernel $k$ in task $s$. By assumption $\mathbb{P}[\sum_{s=m_0}^m r(\tk, s) \leq \Roracle(n, m-m_0)] \geq 1 - \delta/2$. Denote by $\hat{k}_s$ the predicted kernel for task $s$. By union bound
    \begin{align*}
        \mathbb{P} \left[ \sum_{s=m_0}^m r(\hat{k}_s, s) \leq \calO(\Roracle(m,n)) \right] &\geq \mathbb{P} \Bigg[\forall m_0 \leq s \leq m, \hat{k}_s = \tk \text{ and } \\
        & \qquad \sum_{s=m_0}^m r(\hat{k}_s, s) \leq \calO(\Roracle(m-m_0, n)) \Bigg]\\
        &\geq 1 - \mathbb{P} \left[\sum_{s=m_0}^m r(\tk_s, s) > \calO(\Roracle(m-m_0, n))  \right] \\
        & \quad - \mathbb{P} \left[\exists m_0 \leq s \leq m, \tk_s \neq \tk \right]\\
        &\geq 1- \delta
    \end{align*}
    
    Therefore it holds with probability at least $1 - \delta$
    \begin{align*}
        R(m,n) \leq  m_0 n L +  L N_m + \Roracle(m-m_0, n) .
    \end{align*}
    Here, the first term is an upper bound of the regret in the first $m_0$ tasks. The other terms are an upper bound on the reward for the other $m-m_0$ tasks. They can be divided into the regret obtained by forced exploration and the regret obtained by the base bandit task. By Lemma \ref{lem_jjsj} we know that the maximum instantaneous regret $L$ is bounded by $2B$. Therefore
    \begin{align*}
        R(m,n) &\leq  m_0 n L +  L N_m + \Roracle(m-m_0, n) \\
        &\leq 2B m_0 n + 2B N_m + \Roracle(m, n) .
    \end{align*}
\end{proof}

\begin{lemma}
\label{lem_jjsj}
    Let $k$ be a kernel with $k(x,x') \leq 1, \forall x,x' \in \mathcal{X}$ and let $f \in \calH_k$ with $\norm{f}_{k} \leq B$, then for all $\bx \in \calX$
    \begin{align*}
        |f(\bx)| \leq B.
    \end{align*}
\end{lemma}

\begin{proof}
    By the reproducing property, we have
    \begin{align*}
        |f(\bx)| &= | \langle f(\cdot), k(x,\cdot)\rangle_k |\\
        &\leq \norm{f}_{\mathcal{H}_{k}} k(x,x)\\
        & \leq B.
    \end{align*}
\end{proof}

A clarification is due, regarding the exact number of exploratory steps taken. 
In the algorithm design and in the main text, we require that during every task $s$, purely exploratory actions are taken at every step $i$ where $i \leq n_s$. 
The number of exploratory steps has to be an integer, while the proposed rate of $n_s = \sqrt{n}/s^{1/4}$ may not be an integer. 
Therefore, the $i \leq n_s$ condition implies that only the first $\floor{n_s}$ steps will be exploratory.
In our proofs so far, we have assumed that at least a total of $N_{m_0} = \sum_{s=1}^{m_0} n_s$ exploratory action are chosen, which may be well larger than $\sum_{s=1}^{m_0} \floor{n_s}$.
To resolve this gap, we accumulate the non-integer remainder $n_s - \floor{n_s}$ in a variable $r$. Whenever $r$ becomes larger than $1$, we increase the number forced exploration queries by $1$ to $\tilde n_s = \floor{n_s} + \floor{r}$. At every task $s$, we force exactly $\tilde n_s \in \sN$ exploratory actions, where $(\tilde n_1, \dots, \tilde n_s)$ is calculated as described in \cref{alg:forced_exp}.
Then to ensure that $N_{m_0}$ exploratory datapoint are available, we calculate the smallest $\tilde m_0$ which satisfies:
\[
\sum_{s=1}^{\tilde m_0} \tilde n_s \geq N_{m_0}
\]
It is straightforward to show that by construction of \cref{alg:forced_exp}, $m_0 \leq \tilde m_0 \leq m_0 + 1$.
In other words, by taking exploratory actions according to $\tilde n_s$ (which is an integer) we reqiure at most 1 additional task to fulfill the lower bound on the total number of required exploratory actions. 
In the next two corollaries we give a lower bound on the $\tilde m_0$ which satisfies the required dataset size $N_{m_0}$. 

\begin{algorithm}[ht]
\caption{Forced Exploration Rate to Integer number of Exploratory Steps}
\label{alg:forced_exp}
\begin{algorithmic}
\Require The sequence of $(n_1, \dots, n_m)$
\State $r \gets 0 $ \Comment{r is the sum of fractional residue}
\For{$s \in \{1, \dots, m\}$}
\State $r \gets r +  n_s - \floor{n_s}$ \Comment{Add the fractional part of $n_s$ to the residue sum}
\State $\tilde n_s \gets \floor{n_s} + \floor{r}$ \Comment{If the residue sum is over $1$, then add 1 to $\floor{n_s}$}
\State $r \gets r - \floor{r}$
\EndFor\\
\textbf{Output:} $(\tilde n_1, \dots, \tilde n_m)$
\end{algorithmic}
\end{algorithm}

\begin{corollary}
\label{cor:ll_const}
    Assume the setting of \cref{thm:lifelong_main}. Set the rate
    \begin{align*}
        n_s = \sqrt{n}
    \end{align*}
    for all $s \in \mathbb{N}$, and choose the integer number of forced exploration steps according to \cref{alg:forced_exp}. Then, for all $0 < \delta < 1$, with probability at least $1-\delta$
    \begin{equation*}
        R(m,n) \leq \calO\left( B\log(mp/\delta) \sqrt{n} \right) + 2mB \sqrt{n} + \Roracle(m,n).
    \end{equation*}
\end{corollary}

\begin{proof}
    Taking actions at a $n_s = \sqrt{n}$ rate via \cref{alg:forced_exp}, we can ensure that after $\tilde m_0$ many tasks the condition of \cref{thm:lifelong_main} on  $N_{m_0}$ is met, where 
    \begin{align*}
       \frac{2\log(4mp/\delta)}{( n^{1/4} \frac{c_1c_{\kappa}^2}{32\sigma} - 1)^2} \leq \tilde m_0 \leq \frac{2\log(4mp/\delta)}{( n^{1/4} \frac{c_1c_{\kappa}^2}{32\sigma} - 1)^2} + 1.
    \end{align*}
    Then the proof directly from \cref{thm:lifelong_main} with $n_1 = \sqrt{n} $, $N_m = m\sqrt{n} $.
\end{proof}

\begin{corollary}
\label{cor:ll_decreasing}
    Assume the setting of \cref{thm:lifelong_main}. Set the rate
    \begin{align*}
        n_s =  \frac{\sqrt{n}}{s^{1/4}} 
    \end{align*}
     for all $s \in \mathbb{N}$, and choose the explicit integer number of forced exploration steps according to \cref{alg:forced_exp}.
    Then, for all $0 < \delta < 1$, with probability at least $1-\delta$
    \begin{align*}
        R(m,n) \leq \calO\left( B n^{1/3} \log^{3/4}(mp/\delta) m^{1/3} +  B\sqrt{n} m^{3/4}\right) +  \Roracle(m,n).
    \end{align*}
\end{corollary}

\begin{proof}
    We have
    \begin{align*}
        N_m = \sum_{s=1}^m \frac{\sqrt{n}}{s^{1/4}} = \Theta\left( \sqrt{n}m^{3/4} \right)
    \end{align*}
    Choose
    \begin{align*}
        \tilde m_0 = \Theta \left(\frac{ \log(4mp/\delta)}{\left(\sqrt{\frac{n}{m^{1/4}}}\frac{c_1c_{\kappa}}{32\sigma} -1 \right)^2}\right)^{4/3} 
    \end{align*}
    and take exploratory actions according to \cref{alg:forced_exp} then, 
    \begin{align*}
        N_{m_0} \geq \frac{2n_1 \log(4mp/\delta)}{(\sqrt{N_m/m} \frac{c_1c_{\kappa}^2}{32\sigma} - 1)^2}.
    \end{align*}
    By \cref{thm:lifelong_main}, since $n_1 = \sqrt{n}$, for all $0 < \delta < 1$, with probability at least $1-\delta$ \algon achieves
    \begin{align*}
        R(m,n) 
        &\leq 2B m_0 n + 2B N_m + \Roracle(m,n) \\
        &\leq \calO\left( B n^{1/3} \log^{3/4}(mp/\delta) m^{1/3} + B\sqrt{n} m^{3/4} \right) + \Roracle(m,n).
    \end{align*}
\end{proof}
\subsection{Background on \gpucb}\label{app:gp_ucb}

To solve task $s$, \gpucb first constructs {\em confidence sets} for $f_s(\bx)$ based on the history $\{ (\bx_{s,t}, y_{s,t})_{t \leq i}\}$ to balance exploration and exploitation at any step $i$.
 For any $\bx \in \calX$, the set $\calC_{i-1}(\bx)$ defines an interval to which $f(\bx)$ belongs with high probability such that,
 \[\sP\left(\forall\bx\in\calX:f(\bx) \in \calC_{i-1}(\bx)\right)\geq 1-\delta.\]
Given a kernel $k$, \gpucb builds sets of the form 
\begin{align*}
    \calC_{i-1}(k;\bx) = [&\mu_{i-1}(k;\bx)-\nu_i\sigma_{i-1}(k;\bx), \,\, \mu_{i-1}(k;\bx)+\nu_i\sigma_{i-1}(k;\bx)]
\end{align*}
where the exploration coefficient $\nu_i$ depends on the desired confidence level $1-\delta$, and is often treated as a hyper-parameter of the algorithm. The functions $\mu_{i-1}$ and $\sigma_{i-1}$ set the center and width of the confidence set as
\begin{align*}
    \mu_{i-1} (k;\bx)& = {\bm{k}}_{i-1}^T(\bx)({\bm K}_{i-1}+\lambda^2_{\mathrm{ucb}}\bm{I})^{-1}\by_{i-1}  \\
     \sigma^2_{i-1}(k; \bx) & =  k(\bx, \bx) - {\bm k}^T_{i-1}(\bx)({\bm K}_{i-1}+\lambda^2_{\mathrm{ucb}}\bm{I})^{-1}{\bm k}_{i-1}(\bx)
\end{align*}
where $\lambda_{\mathrm{ucb}}$ is a regularizer, $\by_{i-1} = [y_{s,t}]_{t < i}$ is the vector of observed values, $\bm{k}_{i-1}(\bx) = [ k(\bx, \bx_{s,t})]_{t < i}$, and ${\bm K}_{i-1} = [ k(\bx_{s,t}, \bx_{s,t^\prime})]_{t,t^\prime < i}$ is the kernel matrix. 
\gpucb then chooses an action that maximizes the upper confidence bound, i.e.
\begin{equation*}\label{eq:UCB_policy}
\bx_{s,i} = \argmax_{\bx \in \mathcal{X}} \mu_{i-1}(\bx) + \nu_i\sigma_{i-1}(\bx).
\end{equation*}
The acquisition function balances exploring uncertain actions and exploiting the gained information via parameter $\nu_i$.
\citet{chowdhury2017kernelized} show that following this policy, and using $\tk$ as the kernel function, yields a regret of 
\[
\Roracle(n) = \calO \left( B d^\star \sqrt{n}\log\tfrac{n}{d^\star} + \sqrt{nd^\star \log \tfrac{n}{d^\star}\log\tfrac{ 1}{\delta}}\right).
\]
\subsection{Lifelong Regret of \gpucb Paired with \algon (Proof of \texorpdfstring{Corollary~\ref{cor:ll_ucb_decreasing_main_text}}{})}\label{app:lifelong_ucb}

\begin{definition}[maximum information gain]
    The maximum information gain after $t$ observations of GP-UCB with kernel $k$ and parameter $\lambda_{\mathrm{ucb}}$ is defined by
    \begin{equation*}
        \gamma_t(k) = \max_{\bx_1, \dots, \bx_t \in \calX} \frac{1}{2} \log \det \left( I + \lambda_{\mathrm{ucb}}^{-2} K_t\right),
    \end{equation*}
    where
    \begin{equation*}
        (K_t)_{ij} = (\mPhi^T \mPhi)_{ij} = k(\bx_i, \bx_j).
    \end{equation*}
\end{definition}

\begin{theorem}[Theorem 3 of \citet{chowdhury2017kernelized}]
\label{thm_cvgf_old}
    Let $k$ be a kernel and $f \in \calH_k$, where $\calH_k$ is the RKHS corresponding to kernel $k$. Let $\delta \in (0,1)$, $\norm{f}_{k} \leq B$ and assume the errors $\epsilon_t$ are conditionally $\sigma$-sub-Gaussian. Running GP-UCB with $\lambda_{\mathrm{ucb}} = 1 + 2/n$ for $n$ steps we have with probability at least $1-\delta$ that
    \begin{align*}
        R(n) &= \calO(B \sqrt{n \gamma_n(k)} + \sqrt{n \gamma_n(k)(\gamma_n(k) + \log(1/\delta))})
    \end{align*}
\end{theorem}

\begin{corollary}
\label{thm_cvgf}
    Let $k_s$ be kernels and $f_s \in \calH_{k_s}$, where $\calH_{k_s}$ is the RKHS corresponding to kernel $k_s$. Let $\delta \in (0,1)$, $\norm{f_s}_{k_s} \leq B$ and assume the errors $\epsilon_{s,t}$ are $i.i.d.$~$\sigma$-sub-Gaussian. Assume further that $k_s$ are $\sigma(\bm{\epsilon}_1,\dots,\bm{\epsilon}_{s-1})$-measurable. Running GP-UCB with $\lambda_{\mathrm{ucb}} = 1 + 2/n$ for $m$ tasks, each with $n$ steps, we have with probability at least $1-\delta$ that jointly for all $s \in \{1,\dots,m\}$
    \begin{align*}
        R_s(n) &= \calO(B \sqrt{n \gamma_n(k)} + \sqrt{n \gamma_n(k)(\gamma_n(k) + \log(1/\delta))})
    \end{align*}
    where $R_s(n)$ denotes the reward in task $s$ after $n$ interactions.
    In particular
    \begin{align*}
        \Roracle(m,n) \leq \calO \left( m\sqrt{n \gamma_n(k)}(B + \sqrt{\gamma_n(k)} + \log(1/\delta) ) \right).
    \end{align*}
\end{corollary}

\begin{proof}
    We will adapt the proof of Theorem 1 in \citet{chowdhury2017kernelized}.\\
    Let $\epsilon_1^s, \dots, \epsilon_n^s$ be the noise of task $s$. Define a function
    \begin{align*}
        s(t) = \sum_{j=1}^m j \mathbbm{1}_{\{ (j-1)n+1 \leq t \leq jn\}}
    \end{align*}
    and a filtration on $\{1,\dots,mn\}$
    \begin{align*}
        \mathcal{F}_t = \sigma(\epsilon_{1,1}, \dots, \epsilon_{1,n}, \epsilon_{2,1}, \dots, \epsilon_{2,n}, \dots, \epsilon_{s(t),1}, \dots, \epsilon_{s(t),t-(s(t)-1)n}).
    \end{align*}
    Further define for task $s$ a filtration on $\{1,\dots,n\}$
    \begin{align*}
        \mathcal{F}^s_t = \sigma(\epsilon_{s,1}, \dots, \epsilon_{s,t}).
    \end{align*}
    Similar to the proof of Theorem 1 in \citet{chowdhury2017kernelized} define for $t \in \{1,\dots,n\}$, $g:\calX \to \sR$ and $l_1,\dots,l_n \in \mathbb{N}$
    \begin{align*}
        M_t^{g,n}(s) = \exp \left( (\epsilon_{s,1:t})^T g_{1:t,l} - \frac{\sigma^2}{2} \norm{g_{1:t,l}}_2^2\right)
    \end{align*}
    where
    \begin{align*}
        g_{1:t,l} := [g(\bx_1) + l_1, \dots,g(\bx_t) + l_t]^T.
    \end{align*}
    Further let $N_1,\dots,N_n$ i.i.d. with distribution $\mathcal{N}(0, \kappa)$ and independent of $\mathcal{F}_n^s$ and let $h_s$ be a random function distributed according to the Gaussian Process measure $GP_{\calX}(0,k_s)$ and independent of $F^s_n$ and $N_1,\dots,N_n$. Define
    \begin{align*}
        M_t(s) = \mathbb{E}[ M_t^{h_s,N}(s) \mid \mathcal{F}^s_n].
    \end{align*}
    Now by the proof of Theorem 1 of \citet{chowdhury2017kernelized} we have that for all $s \in \{1,\dots,m\}$, $t\in\{1,\dots,n\}$ and all stooping times $\tau_s$ with respect to the filtration $\mathcal{F}^s_t$
    \begin{align}
    \label{eq_kjdfkjdfks}
        \mathbb{E}[M_{\tau_s}(s)] \leq 1.
    \end{align}
    Given stopping times $\tau_1,\dots,\tau_m$ on $\mathcal{F}^1_t,\dots,\mathcal{F}^m_t$ we construct a stopping time $\tau$ on $\mathcal{F}_t$
    \begin{align}
    \label{eq_kjdgjkgdf}
        \tau(\omega) = \min\{mn \geq t \geq 1 \mid \tau^{s(t)}(\omega) = t-(s(t)-1)n  \}.
    \end{align}
    We need to show that $\tau$ is a stopping time with respect to the filtration $\mathcal{F}_t$. We have
    \begin{align*}
        \{\omega\mid \tau(\omega) = t\} = \left(\bigcap_{s < s(t)} \{\omega \mid \tau_s(\omega) > n\} \right) \cap \{\omega \mid \tau^{s(t)}(\omega) = t-(s(t)-1)n \}.
    \end{align*}
    It holds that $\{\omega \mid \tau_s(\omega) > n\} = \{\omega \mid \tau_s(\omega) \leq n\}^c \in \mathcal{F}^s_n \subset \mathcal{F}_{sn}$ and $\{\omega \mid \tau^{s(t)}(\omega) = (s(t)-1)n-t \} \in \mathcal{F}^{s(t)}_{t-(s(t)-1)n} \subset \mathcal{F}_{t}$. This impies that $\{\omega\mid \tau(\omega) = t\} \in \mathcal{F}_t$ and therefore $\tau$ is a stopping time with respect to $\mathcal{F}_t$.
    Define
    \begin{equation*}
        M_t = M_{(s(t)-1)n - t}(s(t)).
    \end{equation*}
    We have that $M_t = M_{(s(t)-1)n - t}(s(t))$ is measurable with respect to $\mathcal{F}^{s(t)}_{(s(t)-1)n-t} \subset \mathcal{F}_t$, which means $M_t$ is $\mathcal{F}_t$-adapted. Let $\tau$ be a stopping time constructed as in Equation \ref{eq_kjdgjkgdf}. Then by Equation \ref{eq_kjdfkjdfks}
    \begin{align*}
        \mathbb{E}[M_{\tau}] \leq 1.
    \end{align*}
    Define for $t \in \{1,\dots,n\}$ and $s\in\{1,\dots,m\}$
    \begin{align*}
        B_t^s(\delta) = \left\{ \omega \mid \norm{\epsilon_{s,1:t}}^2_{((K^s_t +\kappa I)^{-1} + I)^{-1}} > 2 \log\left(\sqrt{\det((1+\kappa)I + K^s_t)}/\delta\right) \right\},
    \end{align*}
    where $K^s_t$ the design matrix for task $s$.
    Further define
    \begin{align*}
        \tau^s(\omega) = \min\{t \in \{1,\dots,n\} \mid \omega \in B^s_t(\delta)\}
    \end{align*}
    and let $\tau$ be the corresponding stopping time on $\mathcal{F}_t$. It holds by the proof of Theorem 1 of \citet{chowdhury2017kernelized} that
    \begin{align*}
        M_t(s) = \frac{\exp \left(\frac{1}{2} \norm{\epsilon_{s,1:t}}^2_{((K^s_{t} +\kappa I)^{-1} + I)^{-1}}\right)}{\sqrt{\det((1+\kappa)I + K^s_{t})}}
    \end{align*}
    and therefore
    \begin{align*}
        M_t = \frac{\exp \left(\frac{1}{2} \norm{\epsilon_{s(t),1:t-(s(t)-1)n}}^2_{((K^{s(t)}_{t-(s(t)-1)n} +\kappa I)^{-1} + I)^{-1}}\right)}{\sqrt{\det((1+\kappa)I + K^{s(t)}_{t-(s(t)-1)n})}}.
    \end{align*}
    Putting things together yields
    \begin{align*}
        \mathbb{P}\left[\bigcup_{s \leq m, t \leq n} B_t^s(\delta) \right] &= \mathbb{P}\left[\tau \leq mn \right]\\
        &= \mathbb{P}\Bigg[\tau \leq mn, \norm{\epsilon_{s(\tau),1:\tau-(s(\tau)-1)n}}^2_{((K^{s(\tau)}_{\tau} +\kappa I)^{-1} + I)^{-1}} >\\
        & \qquad 2 \log\left(\sqrt{\det((1+\kappa)I + K^{s(\tau)}_{\tau-(s(\tau)-1)n}}/\delta\right) \Bigg]\\
        &= \mathbb{P}\left[\tau \leq mn, M_{\tau} > 1/\delta \right]\\
        &\leq \mathbb{P}\left[M_{\tau} > 1/\delta \right]\\
        &\leq \mathbb{E}[M_{\tau}]\delta = \delta.
    \end{align*}
    Now follow the steps of the proof of Theorem 2 of \cite{chowdhury2017kernelized} and the claim follows.
\end{proof}

\begin{lemma}
\label{lem_cvdfg}
    Let $k: \calX \times \calX \to \sR$ be a kernel with $d^{(k)} \in \mathbb{N}$ dimensional feature map and assume $k(x,x') = \bm{\phi}(\bx)^T \bm{\phi}(x') \leq 1, \forall x,x' \in \calX$. Then the maximum information gain of GP-UCB with kernel $k$ and regularization parameter $\lambda_{\mathrm{ucb}}$ satisfies
    \begin{equation*}
        \gamma_n(k) \leq \frac{1}{2} d^{(k)} \log(1 + \frac{\lambda_{\mathrm{ucb}}^{-2}n}{d^{(k)}}).
    \end{equation*}
\end{lemma}

\begin{proof}
    This proofs follows the arguments of \citet{vakili2021information} and \citet{kassraie2022metalearning}. We have that $K_n = \mPhi_n \mPhi_n^T$ and by the Weinstein-Aronszajn identity
    \begin{align*}
        \frac{1}{2} \log \det (I_n + \lambda_{\mathrm{ucb}}^{-2} K_n) &= \frac{1}{2} \log \det (I_{d^{(k)}} + \lambda_{\mathrm{ucb}}^{-2} \mPhi_n^T \mPhi_n) \\
        &\leq \frac{1}{2} d^{(k)} \log\left(\trace(I + \lambda_{\mathrm{ucb}}^{-2} \mPhi_n^T \mPhi_n)/d^{(k)}\right)\\
        &\leq \frac{1}{2} d^{(k)} \log\left(1 + \frac{\lambda_{\mathrm{ucb}}^{-2}}{d^{(k)}}\trace(\mPhi_n^T \mPhi_n)\right).
    \end{align*}
    Now
    \begin{align*}
        \trace( \mPhi_n^T \mPhi_n) &= \sum_{i=1}^{n}  \trace (\bm{\phi}(\bx_i) \bm{\phi}(\bx_i)^T)\\
        &= \sum_{i=1}^{n}  \trace (\bm{\phi}(\bx_i)^T \bm{\phi}(\bx_i))\\
        &= n
    \end{align*}
    and therefore
    \begin{align*}
        \frac{1}{2} \log \det (I_n + \lambda_{\mathrm{ucb}}^{-2} K_n) &\leq \frac{1}{2} d^{(k)} \log\left(1 + \frac{\lambda_{\mathrm{ucb}}^{-2}n}{d^{(k)}}\right).
    \end{align*}
\end{proof}

\begin{corollary}
\label{cor:ll_ucb_const}
    Assume we are in the setting of Corollary \ref{cor:ll_const} with GP-UCB as the base bandit algorithm and $\lambda_{\mathrm{ucb}} = 1 + 2/n$. Then, for all $0<\delta<1$, with probability at least $1-\delta$,
    \begin{align*}
        R(m,n) = \calO \left(  Bm d^\star \sqrt{n}\log\tfrac{n}{d^\star} + m\sqrt{nd^\star \log \tfrac{n}{d^\star}\log\tfrac{ 1}{\delta}} + B\sqrt{n}(m+\log(mp/\delta))\right).
    \end{align*}
\end{corollary}

\begin{proof}
    By Corollary \ref{thm_cvgf} and Lemma \ref{lem_cvdfg} we have with high probability 
    \begin{align*}
        \Roracle(m, n) &= \calO \left( m\sqrt{n} \left((B+ \log(1/\delta))\sqrt{\frac{1}{2} d^\star \log\left(1 + \frac{n^3}{d^\star(n+2)^2}\right)} + \frac{1}{2} d^\star \log\left(1 + \frac{n^3}{d^\star(n+2)^2}\right)\right) \right)\\
        & = \calO \left( Bm d^\star \sqrt{n}\log\tfrac{n}{d^\star} + m\sqrt{nd^\star \log \tfrac{n}{d^\star}\log\tfrac{ 1}{\delta}}\right)
    \end{align*}
    where $d^\star := d^{(\tk)} = \sum_{j \in \tJ} d_j$.
    And using Corollary \ref{cor:ll_const} we have with high probability
    \begin{align*}
        R(m,n) &\leq \calO\left( B\log(mp/\delta) \sqrt{n} + mB \sqrt{n}\right) + \Roracle(m,n)\\
        &= \calO \left(  Bm d^\star \sqrt{n}\log\tfrac{n}{d^\star} + m\sqrt{nd^\star \log \tfrac{n}{d^\star}\log\tfrac{ 1}{\delta}} + Bm\sqrt{n}\right).
    \end{align*}
\end{proof}

\begin{corollary}
\label{cor:ll_ucb_decreasing}
    Assume we are in the setting of Corollary \ref{cor:ll_decreasing} with GP-UCB as the base bandit algorithm and $\lambda_{\mathrm{ucb}} = 1 + 2/n$. Then, for all $0<\delta<1$, with probability at least $1-\delta$,
    \begin{align*}
        R(m,n) = \calO \left( Bm d^\star \sqrt{n}\log\tfrac{n}{d^\star} + m\sqrt{nd^\star \log \tfrac{n}{d^\star}\log\tfrac{ 1}{\delta}} + B n^{1/3} \log^{3/4}(mp/\delta) m^{1/3} + B\sqrt{n} m^{3/4} \right).
    \end{align*}
\end{corollary}

\begin{proof}
    The proof is the same as the proof for Corollary \ref{cor:ll_ucb_const}, except that we use Corollary \ref{cor:ll_decreasing} in place of Corollary \ref{cor:ll_const}.
\end{proof}

\begin{remark}
    Compare the results of Theorem \ref{thm:lifelong} with the default alternative: not learning $k$ and just setting $\hat{k} = \sum_{j=1}^p \frac{1}{p} k_j$. We would then only get a bound of the form
    \begin{equation*}
        R(m,n) \leq \calO\left( m \hat{B} \sqrt{n}\log(n) d  \right),
    \end{equation*}
    where $B = \norm{f}_{\hat{k}} = \frac{p}{s^\star} B$ and $d = \sum_{j=1}^p d_j \geq n$, which is not sublinear in $n$.
\end{remark}

\subsection{Forced Exploration Lower Bound (Proof of \texorpdfstring{Proposition~\ref{prop:forced_exp}}{})} \label{app:kappa_bounds}

\begin{assumption}
\label{ass:der}
    Assume there exists $c_c, c_{od} > 0$
    \begin{align*}
        \frac{m}{N} (\mPhi^T \mPhi)_{i,i} \geq c_d, \qquad \forall i \in \{1,\dots, md\}
    \end{align*}
    and 
    \begin{align*}
        \frac{m}{N} (\mPhi^T \mPhi)_{i,j} < c_{od}, \qquad \forall i \neq j \in \{1,\dots, md\}.
    \end{align*}
\end{assumption}

\begin{lemma}
    Let Assumption \ref{ass:der} be satisfied. Then $\kappa \geq \sqrt{c_{d}/s^\star - 5 c_{od}}$.
\end{lemma}

\begin{proof}
    Let $(b, J) \in S$. We have by definition of $S$
    \begin{align*}
        \left( \frac{\sqrt{m}}{\sqrt{N}} \frac{\|\mPhi \bb\|_2}{ \sum_{j \in J} \|b^{(j)}\|_2}\right)^2
        &=  \frac{m}{N} \frac{b^T (\mPhi^T \mPhi )  b}{\left(\sum_{j \in J} \|b^{(j)}\|_2\right)^2}\\
        & \geq \frac{m}{N} \frac{\sum_{s=1}^m \sum_{i,j \in J} \bb_s^{(i)}(\mPhi^T \mPhi)_{i,j} \bb_s^{(j)}}{\left(\sum_{j \in J} \sqrt{\sum_{s=1}^m (\bb_s^{(j)})^2}\right)^2} \\
        &\quad - 4 \frac{m}{N} \frac{\sum_{s=1}^m \sum_{i \in J, j \in J^c} |\bb_s^{(i)}(\mPhi^T \mPhi)_{i,j} \bb_s^{(j)}|}{\left(\sum_{j \in J} \sqrt{\sum_{s=1}^m (\bb_s^{(j)})^2}\right)\left(\sum_{j \notin J} \sqrt{\sum_{s=1}^m(\bb_s^{(j)})^2}\right)}
    \end{align*}
    By Assumption \ref{ass:der}
    \begin{align*}
        \frac{m}{N} \frac{\sum_{s=1}^m \sum_{i,j \in J} \bb_s^{(i)}(\mPhi^T \mPhi)_{i,j} \bb_s^{(j)}}{\left(\sum_{j \in J} \sqrt{\sum_{s=1}^m (\bb_s^{(j)})^2}\right)^2}
        & \geq \sum_{s=1}^m \sum_{i \in J} \frac{ \bb_s^{(i)} c_2 \bb_s^{(i)}}{\left(\sum_{j \in J} \sqrt{\sum_{s=1}^m (\bb_s^{(j)})^2}\right)^2} \\
        & \quad - \frac{\sum_{s=1}^m \sum_{i \neq j, i,j \in J} |\bb_s^{(i)} c_{od} \bb_s^{(j)}|}{\left(\sum_{j \in J} \sqrt{\sum_{s=1}^m (\bb_s^{(j)})^2}\right)^2}.
    \end{align*}
    Since for $q > 0$ using $\norm{\cdot}_1 \leq \sqrt{s} \norm{\cdot}_2$
    \begin{align*}
        \sum_{s=1}^m \sum_{i \in J} \frac{ \bb_s^{(i)} c_2 \bb_s^{(i)}}{\left(\sum_{j \in J} \sqrt{\sum_{s=1}^m (\bb_s^{(j)})^2}\right)^2} \geq c_2 \sum_{s=1}^m \sum_{i \in J} \frac{ (\bb_s^{(i)})^2}{s^\star\left(\sqrt{\sum_{j \in J} \sum_{s=1}^m (\bb_s^{(j)})^2}\right)^2} = \frac{c_2}{s^\star}
    \end{align*}
    and using Cauchy-Schwarz to prove
    \begin{equation*}
        \sum_{k,l} (x_k y_l)^2 \geq \sum_{k,l} (x_k y_l)(y_k x_l)
    \end{equation*}
    which implies
    \begin{equation*}
        \sqrt{\sum_{k,l} (x_k y_l)^2} \geq \sum_{l} |x_l y_l|
    \end{equation*}
    we get 
    \begin{align*}
        \frac{\sum_{s=1}^m \sum_{i \neq j, i,j \in J} |\bb_s^{(i)} c_{od} \bb_s^{(j)}|}{\left(\sum_{j \in J} \sqrt{\sum_{s=1}^m (\bb_s^{(j)})^2}\right)^2} 
        &= c_{od} \frac{\sum_{s=1}^m \sum_{i \neq j, i,j \in J} |\bb_s^{(i)}| |\bb_s^{(j)}|}{\sum_{i,j \in J} \sqrt{\sum_{k,l} (\bb_l^{(i)}\bb_k^{(j)})^2}} \\
        &\leq c_{od} \frac{\sum_{s=1}^m \sum_{i \neq j, i,j \in J} |\bb_s^{(i)}| |\bb_s^{(j)}|}{\sum_{i,j \in J} \sum_{s=1}^m |\bb_s^{(i)}| |\bb_s^{(j)}|} \\
        & \leq c_{od}.
    \end{align*}
    Also
    \begin{align*}
        &\frac{\sum_{s=1}^m\sum_{i \in J, j \in J^c} |\bb_s^{(i)} c_{od} \bb_s^{(j)}|}{\left(\sum_{j \in J} \sqrt{\sum_{s=1}^m (\bb_s^{(j)})^2}\right)\left(\sum_{j \notin J} \sqrt{\sum_{s=1}^m(\bb_s^{(j)})^2}\right)}\\
        & = c_{od} \frac{\sum_{s=1}^m \sum_{i \in J, j \in J^c} |\bb_s^{(i)} | | \bb_s^{(j)}|}{\sum_{i \in J, j \in J^c} \sqrt{\sum_{k,l} (\bb_l^{(i)}\bb_k^{(j)})^2}}\\
        & = c_{od} \frac{\sum_{s=1}^m \sum_{i \in J, j \in J^c} |\bb_s^{(i)} | | \bb_s^{(j)}|}{\sum_{i \in J, j \in J^c} \sum_{s=1}^m |\bb_s^{(i)}\bb_s^{(j)}|}\\
        & = c_{od}.
    \end{align*}
    Therefore    
    \begin{align*}
        \frac{\sqrt{m}}{\sqrt{N}} \frac{\|\mPhi \bb\|_2}{ \sum_{j \in J} \|b^{(j)}\|_2}
        &\geq \sqrt{c_2/s - 5 c_{od}}.
    \end{align*}
\end{proof}

\begin{proposition}
\label{prop:kappa}
    Let $\mu$ be the Lebesgue measure and $d=p$. Assume that $\bm{\phi}_i \in L_{\mu}^2(\calX)$, $i \in \{1,\dots,p\}$ are orthogonal and satisfy $\|\bm{\phi}_i\|_{L_{\mu}^2(\calX)} / \text{Vol}(\calX) \geq z$, for all $i \in \{1,\dots,p\}$. Assume also that $k_i(x,x) = \bm{\phi}_i(\bx)^2 \leq 1$ for all $\bx \in \calX$. 
    Choose $\bx_1, \dots, \bx_{n}$ i.i.d.~uniformly from $\calX$ and let
    \begin{align*}
        \mPhi_s \coloneqq
        \begin{bmatrix}
            \bm{\phi}(\bx_1) & \dots & \bm{\phi}(\bx_n)
        \end{bmatrix}^T
        \in \sR^{n \times d} \qquad \forall s \leq m.
    \end{align*}
    Then with probability at least $1-\delta$ Assumption \ref{ass:der} is satisfied with
    \begin{align*}
        c_d = \left(z - \sqrt{\frac{1}{2n}\log(4d/\delta)}\right)
    \end{align*}
    and
    \begin{align*}
        c_{od} = \sqrt{\frac{2}{n} \log\left(\frac{4d^2}{\delta}\right)}.
    \end{align*}
\end{proposition}

\begin{proof}
        For the second, let $X$ be a random variable uniformly distributed on $\calX$ and denote by $v_i \coloneqq \bm{\phi}_i(\bx_{1:n})$ the $i$th column of $\mPhi_s$. It holds that
    \begin{align*}
        \mathbb{E}[\bm{\phi}_i(\bx)^2] = \frac{1}{\text{Vol}(\calX)}\int_{\calX} \bm{\phi}_i(\bx)^2 \text{d} \mu(\bx) \geq z, \quad \forall i \leq d.
    \end{align*}
    Therefore
    \begin{equation*}
        \mathbb{E}[\norm{v_i}_2^2] = \mathbb{E}\left[\sum_{i=1}^n \bm{\phi}_i(\bx_i)^2 \right] \geq nz.
    \end{equation*}
    By union bound and Höffding's inequality
    \begin{align*}
        \mathbb{P}\left[\exists i \leq d, \left| \norm{v_i}^2_2 - \mathbb{E}\left[\norm{v_i}^2_2\right] \right| \geq \epsilon \right] \leq 2d\exp(-\frac{2\epsilon^2}{n})
    \end{align*}
    or
    \begin{align*}
        \mathbb{P}\left[\exists i \leq d, \left| \norm{v_i}^2_2 - \mathbb{E}\left[\norm{v_i}^2_2\right] \right| \geq \sqrt{\frac{n}{2} \log(\frac{4d}{\delta})} \right] \leq \delta.
    \end{align*}
    Therefore with probability at least $1-\delta/2$ for all $i \leq d$
    \begin{align*}
        \norm{v_i}_2^2 \geq \mathbb{E}\left[\norm{v_i}^2_2\right] - \sqrt{\frac{n}{2} \log(\frac{4d}{\delta})} \geq nz - \sqrt{\frac{n}{2} \log(\frac{4d}{\delta})}.
    \end{align*}
    Further, for $i \neq j$
    \begin{align*}
        \mathbb{E}[\bm{\phi}_i(\bx)\bm{\phi}_j(\bx)] = \frac{1}{\text{Vol}(\calX)}\int_{\calX} \bm{\phi}_i(\bx)\bm{\phi}_j(\bx) \text{d} \mu(\bx) = 0,
    \end{align*}
    since $\bm{\phi}_i$ and $\bm{\phi}_j$ are orthogonal in $L_{\mu}^2(\calX)$.
    By assumption $\bm{\phi}_i(\bx) \leq 1, \forall i \leq d, \forall \bx \in \calX$ and by Höffding's inequality
    \begin{align*}
        \mathbb{P}[|\langle v_i, v_j \rangle| \geq \epsilon] \leq 2\exp(-\frac{\epsilon^2}{2n}).
    \end{align*}
    and therefore for $0 \leq \delta \leq 1$
    \begin{align*}
        \mathbb{P}\left[\exists i \neq j, |\langle v_i, v_j \rangle| \geq \sqrt{2n \log\left(\frac{4d^2}{\delta}\right)}\right] \leq \delta/2.
    \end{align*}
    We derived that with probability at least $1-\delta$
    \begin{align*}
        (\mPhi^T \mPhi)_{ii} / n\geq c_2 = \left(z - \sqrt{\log(4d/\delta)/2n}\right)
    \end{align*}
    and for $i \neq j$
    \begin{align*}
        (\mPhi^T \mPhi)_{ij} / n < \sqrt{\frac{2}{n} \log\left(\frac{4d^2}{\delta}\right)}.
    \end{align*}
\end{proof}

\begin{corollary}
    Assume the setting of Proposition \ref{prop:kappa}. Then
    \begin{align*}
        \kappa \geq \sqrt{z/s^\star - \sqrt{\frac{\log(4d/\delta) / (2s^\star) + 50 \log(4d^2/\delta)}{n}}} = \calO(1).
    \end{align*}
\end{corollary}
\section{Federated Analysis (Proof of \texorpdfstring{Theorem~\ref{thm:lifelong_federated}}{})}\label{app:federated}

\begin{figure}
    \centering
    \resizebox{0.8\columnwidth}{!}{
        \begin{tikzpicture}[node distance=25mm, roundnode/.style={circle, draw=blue!80, fill=blue!5, very thick, minimum size=7mm},
    squaoraclenode/.style={rectangle, draw=oracle!80, fill=oracle!5, very thick, minimum size=7mm},
    arrow/.style = {thick,-stealth}
    ]
    \draw[ultra thick, rounded corners, dashed, draw=black!80, fill=black!5] (4.5cm, -2.2cm) rectangle (10.5cm, 2.5cm) {};
    \node[rectangle, draw=naive!80, fill=naive!5, very thick, minimum size=7mm, rounded corners=.03cm, align=center] (5) at (6.5cm, -0.6cm) {\bba\\($i > n_s)$};
    \node[squaoraclenode, rounded corners=.03cm] (6) [right of=5] {Environment};
    \node[rectangle, draw=black!80, fill=black!5, very thick, minimum size=7mm, rounded corners=.03cm, align=center] (52) at (6.5cm, 1.7cm) {Forced Exploration\\
    $(i \leq n_s)$};
    
    \draw[rectangle, draw=federated!80, fill=federated!5, very thick, minimum size=7mm, rounded corners=.03cm, align=center] (-0.6cm, 0.5cm) rectangle (2.1cm, 1.5cm) {};    
   \node[align=center] at (0.75cm,1cm) (ml) {$\alpha$-majority Voting\\$\forall 1\leq j\leq p$};
   
   \draw[rectangle, draw=federated!80, fill=federated!5, very thick, minimum size=7mm, rounded corners=.03cm, align=center] (-0.9cm, -1.5cm) rectangle (2.4cm, -0.5cm) {};    
   \node at (0.75cm,-1cm) (gl) {Threshold Group Lasso};
   
    \draw[ thick] (52) -- (5);
    \node (h) at (2.25cm, 2.6cm) {\large$\khat_{s-1}$};
    \draw[ultra thick] (0.75cm, 1.5cm) -- (0.75cm, 2.5cm) -- (1.7cm, 2.5cm);
    \draw[arrow, ultra thick] (2.8cm, 2.5cm) -- ++(1.1cm, 0) -- ++(0, -3.1cm) -- (5);

    \draw[arrow] (5.north)-- ++(0,0.5cm) -- ++(25mm,0) node[midway, above] {$\bx_{s,i}$} -- (6.north);
    \draw[arrow] (6.south)-- ++(0,-0.5cm) -- ++(-25mm,0) node[midway, below] {$f_s(\bx_{s,i}) + \epsilon_{s,i}$} -- (5.south);
    
    \draw[arrow, ultra thick] (7.5cm, -2.2cm)-- ++(0,-0.8cm) -- ++(-4.2cm,0);
    \draw[rounded corners, draw=oracle!80, fill=oracle!5, very thick] (3.25cm, -3.5cm) rectangle (4.25cm, -2.5cm) {};
    \node at (3.75cm, -3cm) {$\Dexp_{s}$};
    \draw[arrow, ultra thick] (3.25cm, -3cm) -- ++(-2.5cm, 0) -- ++(0, 1.5cm);

    \node at (-3.4cm, -1cm) {$\Jhat_{s}$};
    \draw[arrow,  thick] (-3.4cm, -0.5) -- (-3.4cm, 0.5cm);
    \draw[arrow, thick](-0.9, -1) -- (-3, -1);
   
    \node at (-4.75cm, 1cm) {$\Jhat_1$};
    \node at (-4.35, 1cm) {\large ,};
    \node at (-3.95cm, 1cm) {$\Jhat_2$};
    \node at (-3.55, 1cm) {\large ,};
    \node at (-3.1cm, 1cm)  {\large$\dots$};
   \node at (-2.7, 1cm) {\large ,};
    \node at (-2.2cm, 1cm) {$\Jhat_{s-1}$};
    
    \draw [ thick, decorate, decoration={calligraphic brace,amplitude=4pt}] (-5cm, 0.55cm) --(-5cm, 1.45cm) ; %
    \draw [ thick, decorate, decoration={calligraphic brace,amplitude=4pt}] (-1.8cm, 1.45cm) --(-1.8cm, 0.55cm) ;

    \draw[arrow, thick] (-1.6cm, 1cm) -- (ml);
    
    \end{tikzpicture}
    }
    \caption{\falgon visualized. The yellow boxes corresponds to modules of \falgoff.\looseness-1}
    \label{fig:lifelong_algo_fed}
\end{figure}
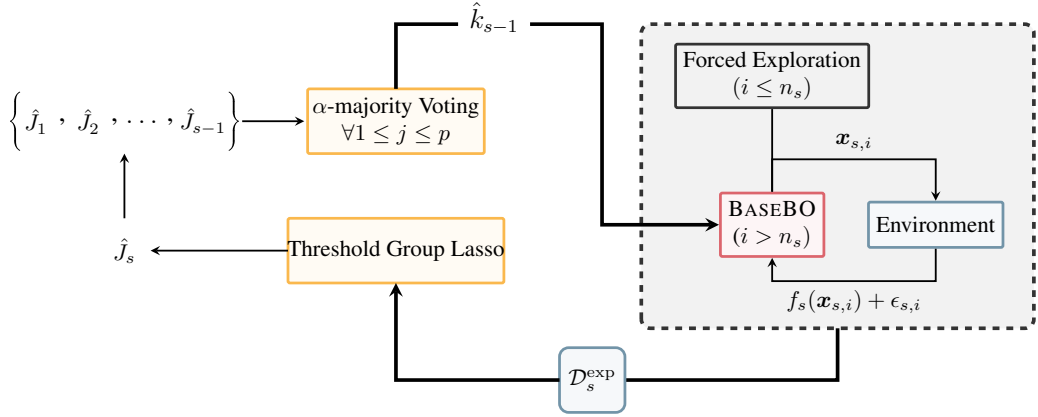

 Recall that in the federated setting, each client minimizes the following loss locally.
 \begin{align}
 \label{eq:meta_loss_federated}
\vbetahat_{s, \mathrm{prvt}} & \coloneqq \argmin_{\vbeta_s \in \sR^{d}} \Ls\left(\vbeta_s; \Dexp_{s} \right) \\
& = \argmin_{\vbeta_s \in \sR^{d}} \frac{1}{n_s} \norm{ \by_s - \mPhi_s \vbeta_s}_2^2  + \lambda \sum_{j=1}^{p} \|\vbeta_s\gj\|_2 .\notag
\end{align}
In this section, for simplicity we refer to the solution as $\vbetahat_s$. We may further omit the subscript $s$, whenever it can be determined from the context.
For our federated analysis, we require a slightly stronger version of the Beta-min assumption.

\begin{assumption}[Beta-min federated]\label{ass:betamin_federated}
    Assume there exists $c_{1,{\mathrm{f}}} > 0$ such that for all $s \leq m$ and $j \in J^\star$
    \begin{align*}
        \norm{\vbeta_s^{\star(j)}}_2 \geq c_{1,{\mathrm{f}}}.
    \end{align*}
\end{assumption}

\begin{remark}
    Note that Assumption \ref{ass:betamin_federated} implies Assumption \ref{ass:betamin}.
\end{remark}

 \subsection{Consistency of the Meta-learned Kernel}
 \label{app:federated_offline}

In this section we prove the equivalent of \cref{thm:offline_main_consitency} in the federated setting. 
 \begin{theorem}[Consistency of \falgoff]
\label{thm:federated_offline}
    Let $\omega \in (0, c_{1, \mathrm{f}})$ and $\bar{\omega} \coloneqq \min \{ \omega, c_{1, \mathrm{f}} - \omega\}$. Let Assumption \ref{ass:betamin_federated} hold.
    Let $n_s = \ubar{n}, \forall s \leq m$ and assume $\ubar{n}$ is large enough to satisfy \smash{$\bar{\omega} > (\sqrt{\log(p/\bar{\alpha})} + 1)32 \sigma/(\sqrt{\ubar{n}} c_{\kappa}^2)$}, where $\bar{\alpha} \coloneqq \max\{\alpha,1-\alpha\}$.
    Assume that $\mPhi_s \in \mathbb{R}^{\ubar{n} \times d}$ satisfy \cref{ass:compatibility} with $c_\kappa$ for $s=1, \dots, m$.
    Let \smash{$\hat{\vbeta}$} be a solution of \Eqref{eq:meta_loss_federated} with regularization parameter \smash{$\lambda = \bar{\omega} c_{\kappa}^2/8$}.
    Then \smash{$\hat{J}_{\mathrm{f}}$} is a consistent estimator in $\ubar{n}$ and $m$, that is
    \begin{align*}
        \lim_{\ubar{n} \to \infty} \mathbb{P}\left[ \hat{J}_{\mathrm{f}} = J^\star \right] = 1 \qquad \text{and} \qquad \lim_{m \to \infty} \mathbb{P}\left[ \hat{J}_{\mathrm{f}} = J^\star \right] = 1.
    \end{align*}
\end{theorem}

We start by proving the necessary lemmas.

\begin{lemma}
\label{lem:federal}
    Let $\hat{\vbeta}$ is a solution of \eqref{eq:meta_loss_federated} and
    \begin{align*}
        \lambda \leq \frac{\bar{\omega} \kappa^2 }{8},
    \end{align*}
    where $\bar{\omega} \coloneqq \min \{ \omega, c_{1,{\mathrm{f}}} - \omega\}$ for $0 < \omega < c_{1,{\mathrm{f}}}$.
    Then we have for $\frac{\lambda n_s}{4 \sigma} > \sqrt{\trace(\Phi_s^{(j)} (\Phi_s^{(j)})^T)}$ with probability at least
    \begin{align*}
        1 - p \max_{j \leq p} \exp\left( -\frac{\left(\frac{\lambda n_s}{4 \sigma} - \sqrt{\trace(\Phi_s^{(j)} (\Phi_s^{(j)})^T)}\right)^2}{2\norm{\Phi_s^{(j)}(\Phi_s^{(j)})^T}_2} \right) 
    \end{align*}
    that
    \begin{equation}
        \sum_{j=1}^p \norm{\hat{\vbeta}_s^{(j)} - \vbeta_s^{\star(j)}}_2 \leq \bar{\omega}.
    \end{equation}
    If additionally Assumption \ref{ass:betamin_federated} holds, then we have with the same probability for
    \begin{align*}
        \hat{J}_{s, \mathrm{f}} = \left\{ j \in \{1,\dots, p\} \ \Big\vert \ \norm{\hat{\vbeta}_s^{(j)}}_2 > \omega \right\}
    \end{align*}
    that
    \begin{align*}
        \hat{J}_{s, \mathrm{f}} = J^\star.
    \end{align*}
\end{lemma}

\begin{proof}
    Follows directly from Corollary \ref{cor:main} with $m=1$.
\end{proof}

\begin{lemma}[Chernoff-Höffding bound]
\label{lem_chbound}
    Let $X_1, \dots, X_n$ be i.i.d Bernoulli random variables with $\mathbb{E}[X_i] = p_i$. Define $p\coloneqq \frac{1}{n} \sum_{i=1}^n p_i$, then for $t < np$,
    \begin{align*}
        \mathbb{P}\left[\sum_{i=1}^n X_i \leq t\right] &\leq \exp\left(-n \left( \frac{t}{n} \log\left(\frac{t}{np}\right) + (1-t/n) \log\left(\frac{1-t/n}{1-p}\right) \right)\right)\\
        &\leq \exp\left(-n  \frac{(p - t/n)^2}{2p(1-p)}\right).
    \end{align*}
\end{lemma}

\begin{lemma}
\label{thm:main_federated}
    Let $ 0 < w < c_{1,{\mathrm{f}}}$ and let $\hat{\vbeta}_s$ be the solution of \eqref{eq:meta_loss_federated} for tasks $s \leq m$ and $\lambda \leq \frac{\bar{\omega} \kappa^2 }{8}$. Define for $j \in \{1,\dots,p\}$
    \begin{align*}
        Q_j = \big\{ s \in \{1,\dots, m\} \ \big\vert \ \norm{\hat{\vbeta}^{(j)}_s}_2 > w\big\}
    \end{align*}
    and for $\alpha > 0$
    \begin{equation}
        \hat{J}_{\mathrm{f}} = \big\{ j \in \{1,\dots,p\}  \ \big\vert \ |Q_j| > m\alpha  \big\}.
    \end{equation}
    Define for $s \in \{1,\dots,m\}$
    \begin{align*}
        v_s &\coloneqq 1 - p \max_{j \leq p} \exp\left( -\frac{\left(\frac{\lambda n_s}{4 \sigma} - \sqrt{\trace(\Phi_s^{(j)} (\Phi_s^{(j)})^T)}\right)^2}{2\norm{\Phi_s^{(j)}(\Phi_s^{(j)})^T}_2} \right)
    \end{align*}
    and
    \begin{align*}
        v \coloneqq \frac{1}{m} \sum_{s=1}^m v_s.
    \end{align*}
    Assume that $\frac{\lambda n_s}{4 \sigma} > \sqrt{\trace(\Phi_s^{(j)} (\Phi_s^{(j)})^T)}, \forall s \leq m$ and $v > \bar{\alpha} \coloneqq \min\{\alpha, 1-\alpha\}$. Then
    \begin{align*}
        \mathbb{P}\left[ J^\star = \hat{J}_{\mathrm{f}} \right] \geq 1 - p\exp\left( -m \frac{(v-\bar{\alpha})^2}{2v(1-v)} \right).
    \end{align*}
\end{lemma}

\begin{proof}
Recall that $\bm{\epsilon}_s = [\epsilon_{s, i}]_{i=1}^n$.
    Since $\bm{\epsilon}_1, \dots, \bm{\epsilon}_m$ are independent,
    \begin{align*}
        \mathbbm{1}_{\left\{ \sum_{j=1}^p \norm{\hat{\vbeta}^{(j)}_1 - \vbeta^{\star(j)}_1}_2 \leq \bar{\omega} \right\}}, \dots, \mathbbm{1}_{\left\{ \sum_{j=1}^p \norm{\hat{\vbeta}^{(j)}_m - \vbeta^{\star(j)}_m}_2 \leq \bar{\omega} \right\}}
    \end{align*}
    are independent and Bernoulli distributed with coefficient 
    \begin{align*}
        \mathbb{P}\left[\sum_{s=1}^p \norm{\hat{\vbeta}^{(j)}_s - \vbeta^{\star(j)}_s}_2 \leq \bar{\omega} \right] \geq v_s, 
    \end{align*}
    where we used Lemma \ref{lem:federal} and set $\bar{\omega} \coloneqq \min \{ \omega, c_{1,{\mathrm{f}}} - \omega\}$.
    If $j \in J^\star$ and $\norm{\hat{\vbeta}^{(j)}_s - \vbeta^{\star(j)}_s}_2 < \bar{\omega}$, then by Assumption \ref{ass:betamin_federated}
    \begin{align*}
        \norm{\hat{\vbeta}_s^{(j)}}_2 > c_{1,{\mathrm{f}}} - \norm{\hat{\vbeta}_s^{(j)} - \vbeta^{\star(j)}}_2  \geq c_{1,{\mathrm{f}}} - \bar{\omega} \geq \omega,
    \end{align*}
    which implies $J^\star \subset \hat{J}$.
    If $j \notin J^\star$ and $\norm{\hat{\vbeta}^{(j)}_s - \vbeta^{\star(j)}_s}_2 < \bar{\omega}$, then
    \begin{align*}
        \norm{\hat{\vbeta}_s^{(j)}}_2 \leq \norm{\hat{\vbeta}_s^{(j)} - \vbeta_s^{\star(j)}}_2 +  \norm{\vbeta_s^{\star(j)}}_2 \leq \bar{\omega} \leq \omega.
    \end{align*}   
    We have by Lemma \ref{lem_chbound} and for $v > \bar{\alpha}$,
    \begin{equation}
    \label{eq:main_federated}
    \begin{split}
        \mathbb{P}\left[  \hat{J}_{\mathrm{f}} = J^\star \right] &\geq
        \mathbb{P}\left[ \forall j \in J^\star, |Q_j| \geq m/x; \forall j \notin J^\star, |Q_j| < m/x \right]\\
        &\geq \mathbb{P}\Bigg[ \forall j \notin J^\star, \sum_{s=1}^m \mathbbm{1}_{\left\{ \norm{\hat{\vbeta}^{(j)}_s}_2 \leq w \right\}} \geq m/x; \forall j \in J^\star, \sum_{s=1}^m \mathbbm{1}_{\left\{ \norm{\hat{\vbeta}^{(j)}_s}_2 > w \right\}} \geq m/x \Bigg]\\ 
        &\geq \mathbb{P}\left[ \forall j \in \{1, \dots,p\}, \sum_{s=1}^m  \mathbbm{1}_{\left\{ \norm{\hat{\vbeta}^{(j)}_s - \vbeta^{\star(j)}_s}_2 \leq \bar{\omega} \right\}} \geq m\min\{\alpha, 1 - \alpha\} \right]\\ 
        &\geq \mathbb{P}\left[\sum_{s=1}^m \mathbbm{1}_{\left\{ \sum_{j=1}^p \norm{\hat{\vbeta}^{(j)}_s - \vbeta^{\star(j)}_s}_2 \leq \bar{\omega} \right\}}  \geq m \bar{\alpha} \right]\\
        & \geq 1- \exp\left(-m \left( \bar{\alpha} \log\left(\frac{\bar{\alpha}}{v}\right) + (1-\bar{\alpha}) \log\left(\frac{1-\bar{\alpha}}{1-v}\right) \right)\right)\\
        & \geq 1 - \exp\left( -m \frac{(v-\bar{\alpha})^2}{2v(1-v)} \right).
        \end{split}
    \end{equation}
\end{proof}

\begin{proof}[\textbf{Proof of \cref{thm:federated_offline}}]
Assume the setting of \cref{thm:main_federated} and that there exists $c_{\kappa}> 0$ such that $\kappa \geq c_{\kappa}$. Set
    $\lambda = \frac{\bar{\omega} c_{\kappa}^2 }{8}$, $n_s = \ubar{n}, \forall s\leq m$ and assume
    $\frac{\lambda \sqrt{\ubar{n}}}{4 \sigma} > 1$ and $v = 1 - p\exp(-(\lambda \sqrt{\ubar{n}} / 4 \sigma - 1)^2/2)  > \bar{\alpha}$. 
    
    Note that $\Phi_s^{(j)} \in \sR^{N \times md_j}$ is block-diagonal. Since by assumption $k_j(x,x') \leq 1, \forall j \leq p$, we have
    \begin{align*}
        \norm{(\Phi_s^{(j)})^T \Phi_s^{(j)}}_2 \leq \trace((\Phi_s^{(j)})^T \Phi_s^{(j)}) = \trace(\Phi_s^{(j)} (\Phi_s^{(j)})^T) = \sum_{i=1}^{\ubar{n}} k_j\left(x^{(s)}_i, x^{(s)}_i\right) \leq \ubar{n}.
    \end{align*}
     \cref{thm:main_federated} yields the result.
\end{proof}

 \subsection{Lifelong Regret of \falgon (Proof of \texorpdfstring{Theorem~\ref{thm:lifelong_federated}}{})} \label{app:federated_lifelong}
 
 We start by stating \cref{thm:lifelong_federated} more rigorously. 
 
\begin{theorem}
\label{thm:lifelong_main_federated}
    Assume that the true reward functions $f_1, \dots, f_m$ satisfy $\norm{f_i}_{\tH} \leq B$ for some constant $B > 0$. Let $\bar{n}$ be the number of times forced exploration is used in each task. Let $\nu$ be a distribution on $\calX^{\bar{n}m}$ independent of $\bm{\epsilon}_1, \dots, \bm{\epsilon}_m$. Let $V \sim \nu$ be the random vector used for forced exploration. Let $\tilde{\Phi}_s \in \sR^{\bar{n} \times md}$ be the data matrix obtained by forced exploration in task $s$. Set $\lambda = \bar{\omega} c_{\kappa}^2/8$. Assume the forced exploration distribution $\nu$ and $\{k_j\}_{j\leq p}$ are such that, with probability at least $1-\delta/4$, there exists $c_{\kappa} > 0$ such that $\kappa(\tilde{\Phi}_s) \geq c_{\kappa}, \forall s \leq m$.
    Assume further that the base bandit algorithm using the true kernel function achieves on $m$ tasks with independent noise with probability at least $1-\delta/2$ cumulative regret lower than $\Roracle(n, m)$. Define
    \begin{align*}
        v &\coloneqq 1 - p\exp\left( -\frac{1}{2} \left(\frac{\bar{\omega} c_{\kappa}^2\sqrt{\bar{n}}}{32 \sigma} -1 \right)^2 \right).
    \end{align*}
    and assume for all $s \leq m$
    \begin{equation}
        v \geq \bar{\alpha}, \qquad \qquad \qquad \frac{\bar{\omega} c_{\kappa}^2\sqrt{\bar{n}}}{32 \sigma} > 1.
    \end{equation}
    Then with probability at least $1-\delta$, \algon (using \falgoff to predict the kernel) achieves
    \begin{align*}
        R(m,n) \leq \calO\left( B n \log(mp/\delta) / \bar{n} + B m \bar{n}  \right)+ \Roracle(n, m).
    \end{align*}
\end{theorem}

\begin{proof}
    Similar to the proof of Theorem \ref{thm:lifelong_main} we have by  \Eqref{eq:main_federated} for all $s$ and $v' \in C$
    \begin{align*}
        \mathbb{P}\left[  \hat{J}_{s} = J^\star \mid V = v' \right] \geq 1 - p\exp\left(-s \left( \bar{\alpha} \log\left(\frac{\bar{\alpha}}{v}\right) + (1-\bar{\alpha}) \log\left(\frac{1-\bar{\alpha}}{1-v}\right) \right)\right).
    \end{align*}
    By union bound we have for $m_0 \leq m$
    \begin{align*}
        \mathbb{P}\Big[ \forall m \geq s \geq m_0, \hat{J}_s = J^\star &\mid V = v'  \Big] \\
        &\geq 1 - \sum_{s=m_0}^m p\exp\left(-s \left( \bar{\alpha} \log\left(\frac{\bar{\alpha}}{v}\right) + (1-\bar{\alpha}) \log\left(\frac{1-\bar{\alpha}}{1-v}\right) \right)\right)\\
        & \geq 1 - m p\exp\left(-m_0 \left( q - (1-\bar{\alpha}) \log\left(1-v \right) \right)\right),
    \end{align*}
    where $q \coloneqq \bar{\alpha} \log\left(\bar{\alpha}\right) + (1-\bar{\alpha}) \log\left(1-\bar{\alpha} \right)$.
    Set
    \begin{align*}
        m_0 = \left\lceil \frac{\log(4mp/\delta)}{\bar{q} + (1-\bar{\alpha})(\bar{w} c_{\kappa}^2 \sqrt{\bar{n}}/32 \sigma-1)^2/2 } \right\rceil,
    \end{align*}
    where $\bar{q} \coloneqq q  - (1-\bar{\alpha}) \log(p)$.
    Following the same steps as in the proof of Theorem \ref{thm:lifelong_main} we get
    \begin{align*}
        R(m,n) &\leq \calO\left(m_0 n L +  L m \bar{n} \right)+ \Roracle(n, m - m_0)\\
        &\leq \calO\left(2B m_0 n + 2B m \bar{n} \right)+ \Roracle(n, m) \\
        & \leq \calO\left( B n \log(mp/\delta) / \bar{n} + B m \bar{n}\right) + \Roracle(n, m) .
    \end{align*}
\end{proof}

\begin{corollary}
\label{cor:main_ll_federated}
    Assume the setting of Theorem \ref{thm:lifelong_main_federated} and set $\bar{n} =  \sqrt{n} $. Then with probability at least $1-\delta$ we have
    \begin{align*}
        R(m,n) \leq \calO\left( B \sqrt{n} (\log(mp/\delta) +m) \right) + \Roracle(n, m) .
    \end{align*}
\end{corollary}

 \subsection{Performance of \gpucb paired with \falgon} \label{app:federated_ucb}
 
 \begin{corollary}
\label{cor:ll_ucb_federated}
    Assume we are in the setting of Corollary \ref{cor:main_ll_federated} with GP-UCB as the base bandit algorithm and $\lambda_{\mathrm{ucb}} = 1 + 2/n$. Then, for all $0<\delta<1$, with probability at least $1-\delta$,
    \begin{align*}
  R(m,n) = \calO \left(  Bm d^\star \sqrt{n}\log\tfrac{n}{d^\star} + m\sqrt{nd^\star \log \tfrac{n}{d^\star}\log\tfrac{ 1}{\delta}} + B\sqrt{n}(m+\log(mp/\delta))\right).
    \end{align*}
\end{corollary}

\begin{proof}
    The proof is the same as the proof for Corollary \ref{cor:ll_ucb_const}, except that we use Corollary \ref{cor:main_ll_federated} in place of Corollary \ref{cor:ll_const}.
\end{proof}
\section{Experiment Details}
\label{app:experiment_details}

For the synthetic experiments, we initiate the algorithms with $\omega = c_1/2$.
For all experiments, the exploration coefficient of the \gpucb algorithm is set to $\nu_i = 10$ and $\lambda_{\mathrm{ucb}} = 0.1$. 
Experiment are all repeated $20$ times for difference random seeds, and the plots show the corresponding standard error. 
The remaining experiment settings are detailed in the following subsections.

\subsection{Offline Data experiments}\label{app:exp_offline_detail}
We generate the reward functions $f_1, ..., f_{30}$ from the synthetic environment. Corresponding to each $f_s$, we generate a data set $\mathcal{D}_s$ of size $n=10$ by sampling points $\bx_{s,1}, \dots, \bx_{s,n}$ i.i.d from a uniform distribution $\calU(\calX)$ over the domain $\calX = [0,1]$ and collecting the corresponding noisy function values $y_{s,i} = f_s(\bx_{s,i}) + \epsilon$, where the noise is samples from $\calN(0, \sigma^2=0.01)$.
We initiate \algoff with the lasso regularization parameter of $\lambda = 0.25$ and $\falgoff$ with $\lambda = 0.015$. For \falgoff, we set the majority vote threshold to $\alpha = 0.25$.

\subsection{Lifelong Data Experiments}\label{app:exp_lifelong_detail}
For experiments using synthetic data, we set $n=100$, and for the experiments on \textsc{GLMNET} data, there are $n=144$ BO steps in each task.
To run \algon on the synthetic environment we set $\lambda = 0.5$ and for \falgon we set $\lambda = 0.2$.
On the \textsc{GMLNET} environment, we instantiate \algon with $\omega=0.25$ and $\lambda=0.015$, and \falgoff with $\alpha = 0.25$, $\omega=10^{-6}$, $\lambda= 2.6 \times 10^{-6}$.

\subsection{Further Experiments with Synthetic Data}
\label{app:experiment_results}

\begin{figure}[ht]
    \centering
    \begin{subfigure}{0.45\textwidth}
        \centering
        \includegraphics[width=\textwidth]{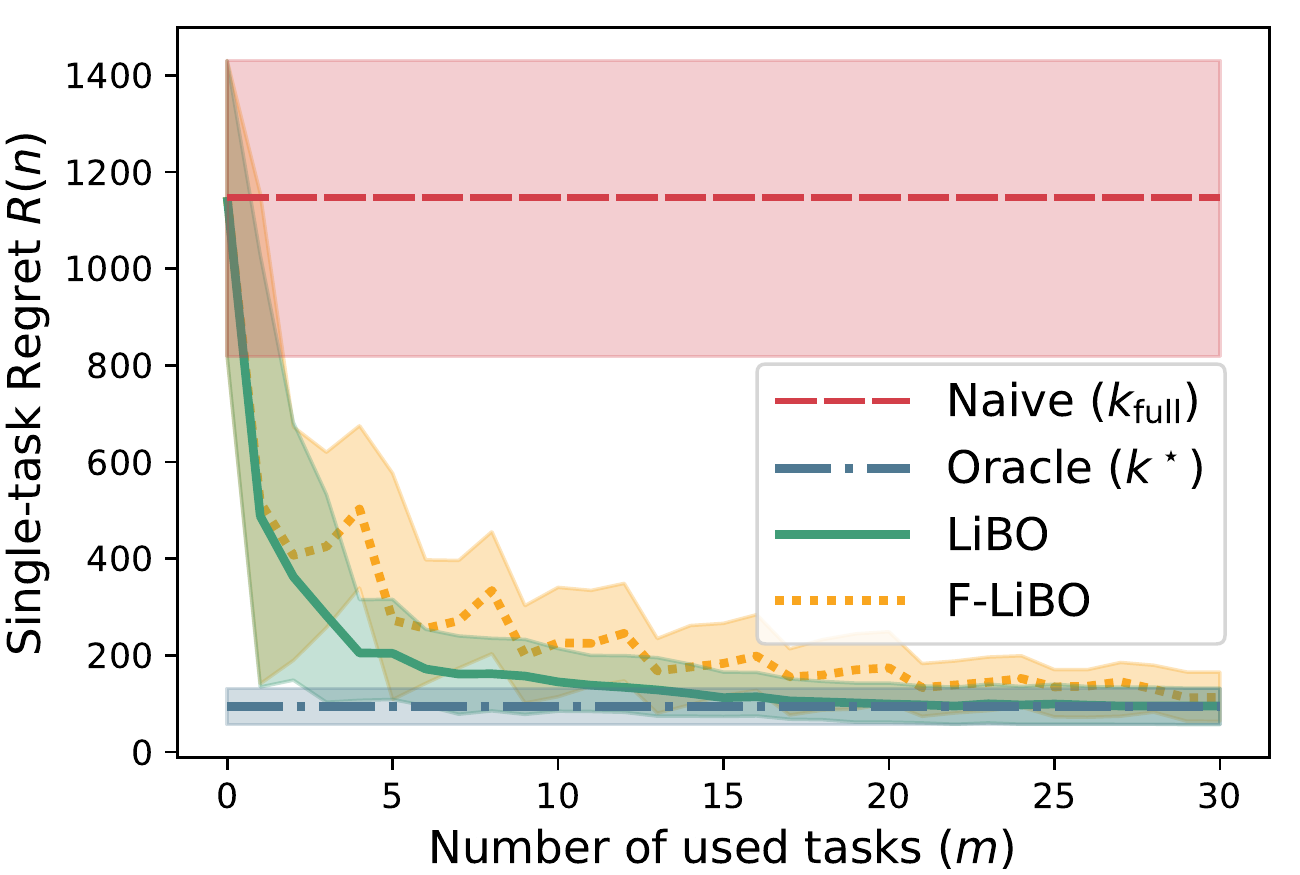}
    \end{subfigure}
    \hfill
    \begin{subfigure}{0.45\textwidth}
        \centering
        \includegraphics[width=\textwidth]{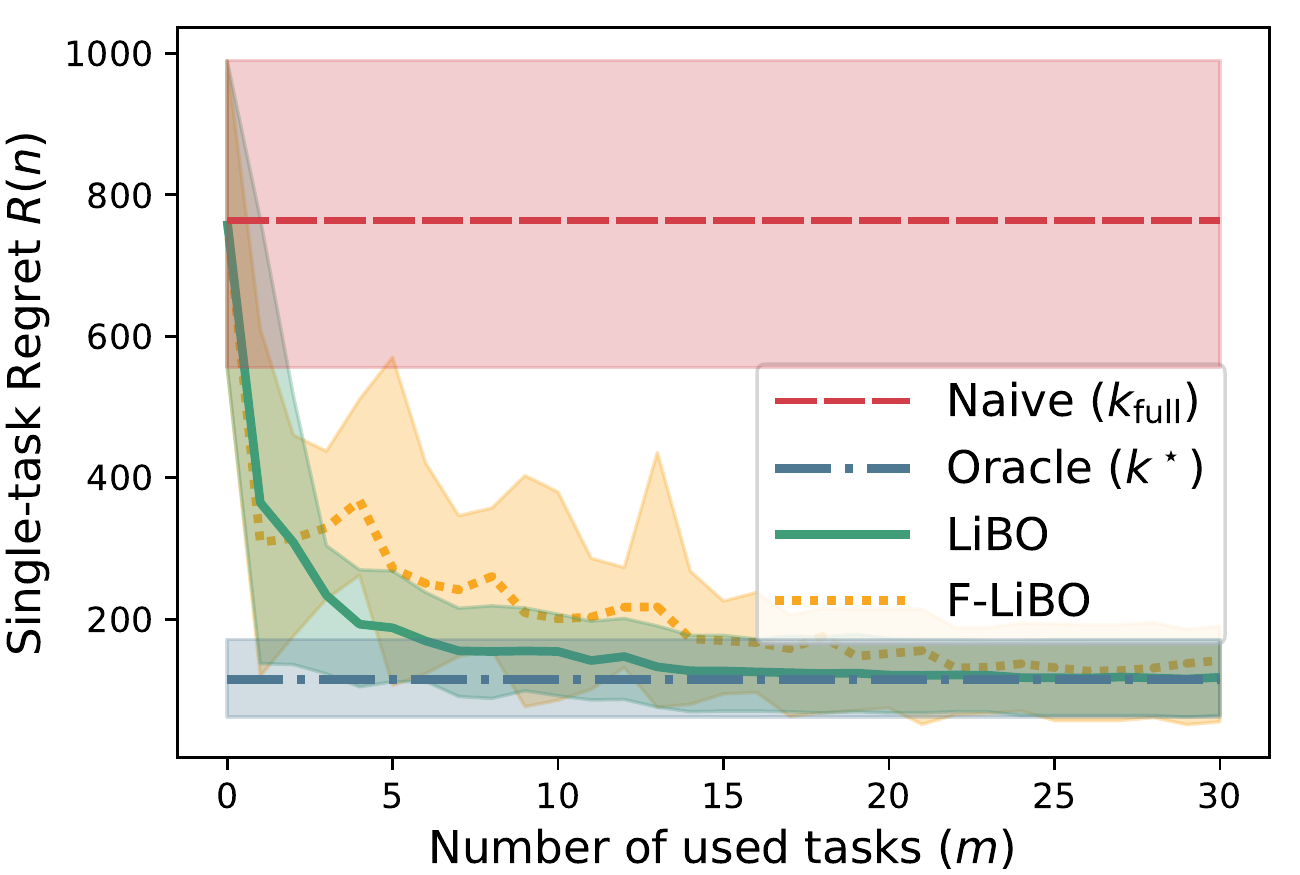}
    \end{subfigure}
    \caption{Single task cumulative regret of GP-UCB with meta-learned kernel $\hat{k}$ on an increasing number of meta-training tasks. Left: base kernels constructed with 2-dimensional cosine basis. Right: base kernels constructed with 1-dimensional Legendre polynomials. The BO performance with meta-leaned kernels quickly approaches oracle performance as the number of meta-training task increases.\label{fig:offline_more}}
\end{figure}

\paragraph{Offline Data}  Analogous to the offline data experiments in Section \ref{sec:offline_exps}, we provide additional results for a two-dimensional domain and Legendre polynomials instead of cosine bases in Figure \ref{fig:offline_more}. In particular, the left plot corresponds to $\mathcal{X} = [0, 1]^2$ as the domain and the first $50$ 2-dimensional cosine basis functions, i.e., $\phi_{i,j}(x) = \cos(i \pi x_1)\cos( \pi x_2), \forall x \in \mathcal{X}$, as the feature maps. For the right plot we choose $\mathcal{X} = [-1, 1]$ as the domain and use the first $50$ Legendre Polynomials as the feature maps. 

Figure \ref{fig:offline_more} shows that both meta-learners converge with increasing number of tasks to the oracle kernel. This holds for different sets of base kernels and kernels with more than $1$ input dimension. This empirically validates the theoretical findings of \cref{thm:offline_main_consitency} and \cref{thm:federated_offline}. Somewhat peculiar is that we can observe oscillating behavior for the federated algorithm (yellow). This is a result of discrete nature of the voting system. The the total of number of tasks is a multiple of $\alpha$ the value $\vert \Jhat_s \vert$ is large, while for points directly after that $\vert \Jhat_s \vert$ are small. With increasing number of tasks the discretization has a lesser impact on the kernel estimation and the amplitude of the oscillations decreases. 

\begin{figure}[ht]
    \centering
    \begin{subfigure}{0.45\textwidth}
        \centering
        \includegraphics[width=\textwidth]{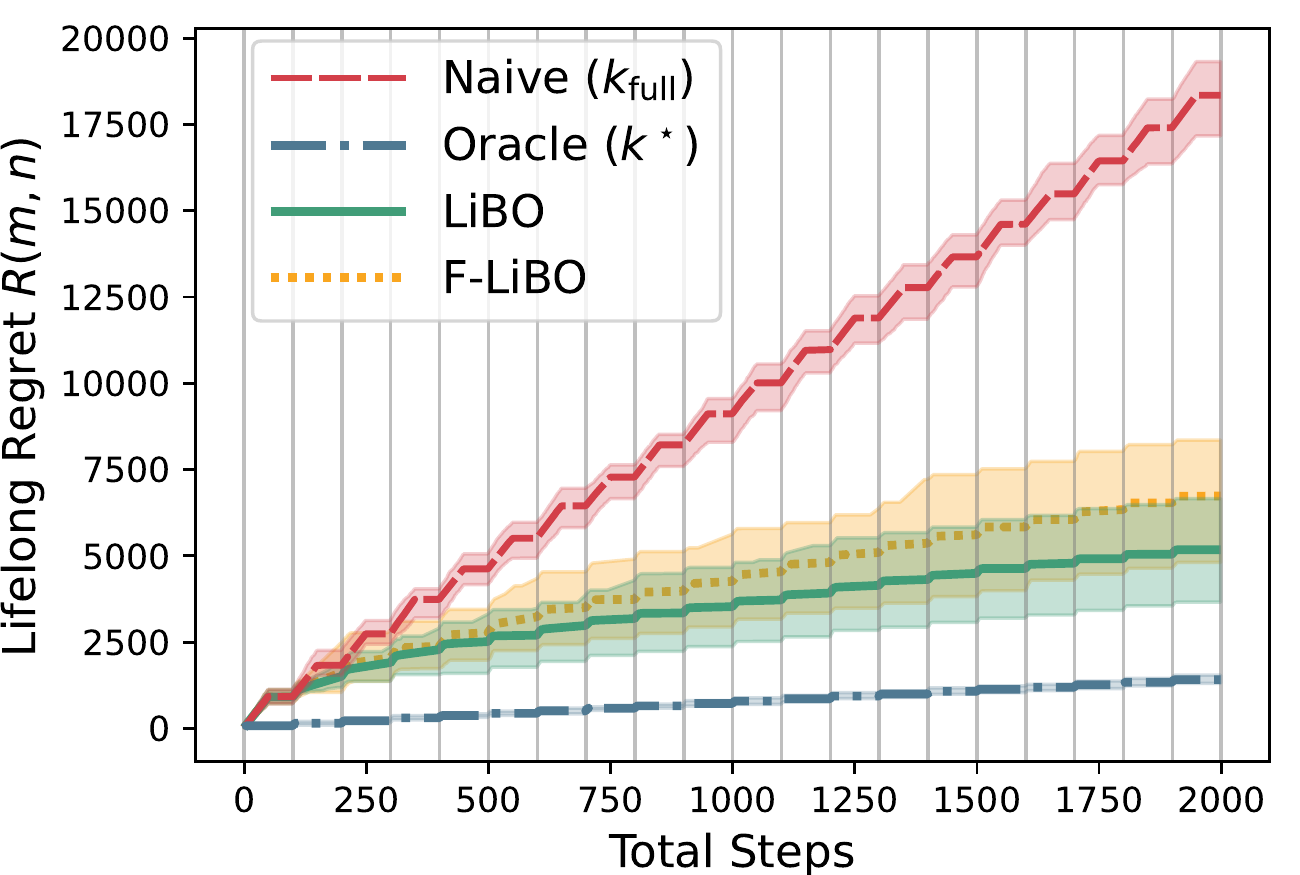}
        \caption{}
    \end{subfigure}
    \hfill
    \begin{subfigure}{0.45\textwidth}
        \centering
        \includegraphics[width=\textwidth]{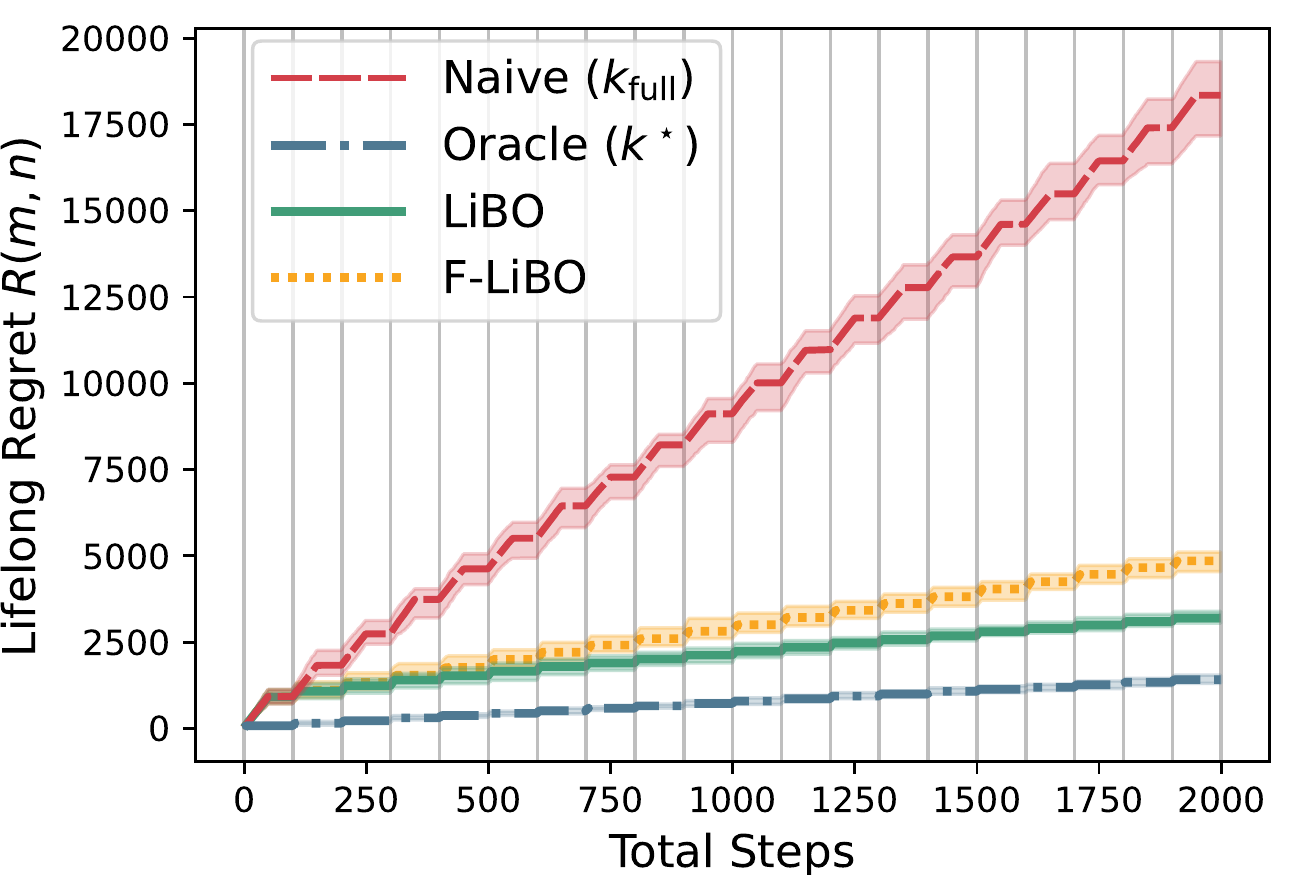}
        \caption{}
    \end{subfigure}
        \begin{subfigure}{0.45\textwidth}
        \centering
        \includegraphics[width=\textwidth]{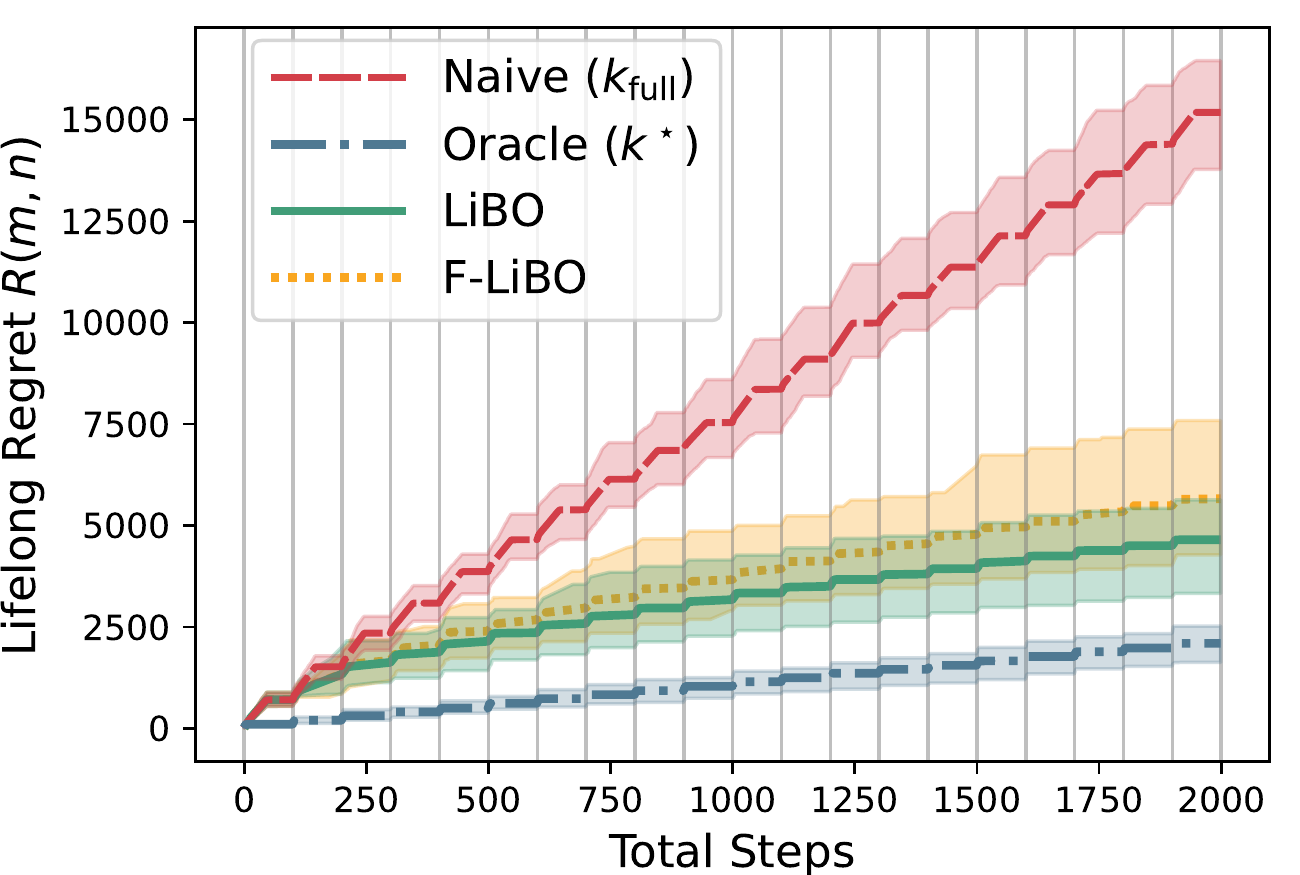}
        \caption{}
    \end{subfigure}
    \hfill
    \begin{subfigure}{0.45\textwidth}
        \centering
        \includegraphics[width=\textwidth]{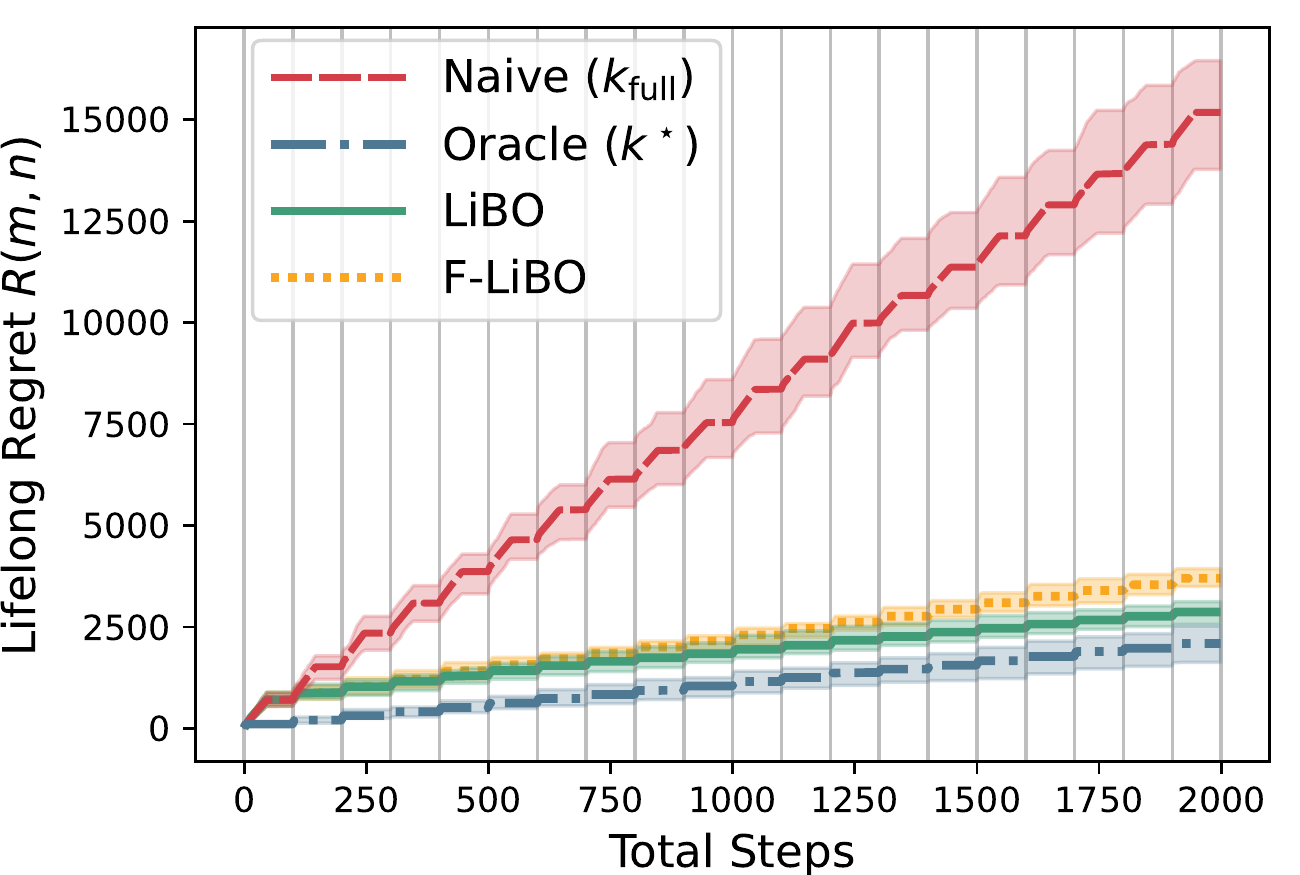}
        \caption{}
    \end{subfigure}
    \caption{Lifelong regret with cosine basis (\textbf{a} \& \textbf{b}) and Legendre polynomials as feature maps (\textbf{c} \& \textbf{d}).
    In the plots \textbf{(a \& c)} on the left, we use only the forced exploration data $\Dexp_{1:s}$ for meta-learning the kernel. For the plots \textbf{(b \& d)} on the right, $D_{1:s}$, the data from all previous bandit interactions, is used. We observe that convergence is much faster when all interaction data is used in \algon and \falgon. \label{fig:online_more}}
\end{figure}

\paragraph{Lifelong Data} We now present modifications of the lifelong BO experiments in Section \ref{exp:lifelong}. In particular, we consider other base kernels as well as a modification of \algon where we use all collected data for meta-learning $\hat{k}$ instead of only the forced exploration data $\Dexp_{1:s}$. The results are depicted in Figure \ref{fig:online_more}. Figure \textbf{(a)} and \textbf{(b)} correspond to $50$ cosine basis functions as feature maps for the base kernels. For Figure \textbf{(c)} and \textbf{(d)}, we use the first $50$ Legendre polynomials as feature maps. The plots on the left (i.e. Fig. a, c) are generated with \algon and \falgon, as presented in Algorithm \ref{alg:lifelong} and \ref{alg:lifelong_fed}, where only the forced exploration data is used for meta-learning. The plots on the right (i.e. Fig. b, d) correspond to a modified version of \algon and \falgon where we use $D_{1:s}$, i.e., all previous bandit interactions, to meta-learn the kernel.

Generally, we observe that \algon and \falgon substantially outperform the naive method which uses all base kernels.
The gray vertical lines in Figure \ref{fig:online_more} indicate the beginning of a new task. We see that for every new task all algorithms initially experiences high regret, but, over time, as reward estimation improves, the cumulative regret flattens. As the rate of single-task convergence is dependent on the kernel, we see that differences in the performance between the algorithms emerge.
When running \algon, over time, forced exploration decreases and the estimated kernel converges to the true kernel.
This means that, over time, the behavior of the agent using the \algon estimator becomes indistinguishable form the agent using the oracle kernel.
This is evident from \ref{fig:online_more} \textbf{(a)} as the slope of the single-task cumulative regret of the meta-agent (green) becomes the same as for the oracle agent (blue).
In the federated case (yellow), while the estimated kernel also converges to the true kernel, the more restrictive setting forces us to use a constant exploration rate (see \cref{alg:lifelong_fed} and \cref{sec:federated}) which means that the behavior of the federated meta-learner is always slightly sub-optimal.
This can be observed by noting that the slope of the single-task cumulative regret of the federated meta-learner (yellow) is higher compared to the oracle agent even after the estimated kernel converges to the true kernel.

When we adjust \algon and \falgon to use all available data to predict the kernel instead of only using $\Dexp_{1:s}$, the lifelong regret decreases.
As we would expect, using more data for meta-learning the kernel speeds up the convergence of $\hat{k}$ to $k^*$ which, in turn, makes the BO runs more efficient. In practice, using the data from all interactions, not just the ones obtained by forced exploration, seems to be the best choice. From a theoretical perspective, this comes with additional technical challenges, as we point out in \cref{sec:forced_exp}.

\end{document}